 \documentclass[10pt]{IEEEtran}

\usepackage{lineno,hyperref}

\usepackage{amsmath,amssymb,color,accents,ushort}
\usepackage{graphicx}

\usepackage{subfigure}
\usepackage{epsfig}
\usepackage{epstopdf}

\usepackage{microtype}

\newcommand{\cred}{}
\newcommand{\cblue}{}

\newcommand{\ua}{\uparrow}

\newcommand{\vs}{\mathbf{s}}
\newcommand{\vp}{\mathbf{\textstyle{p}}}

\newcommand{\ve}{\mathbf{e}}

\newcommand{\vx}{\mathbf{x}}
\newcommand{\vxh}{\hat{\mathbf{x}}}

\newcommand{\vy}{\mathbf{y}}
\newcommand{\mG}{\mathbf{G}}

\newcommand{\mD}{\mathbf{D}}

\newcommand{\mS}{\mathbf{S}}

\newcommand{\mH}{\mathbf{H}}

\newcommand{\mM}{\mathbf{M}}

\newcommand{\nc}{\newcommand}
\nc{\da}{\downarrow} \nc{\hc}{\hat{c}} \nc{\hS}{\hat{S}}
\nc{\bra}{\langle} \nc{\ket}{\rangle} \nc{\eq}{equation (\ref}
\nc{\h}{\hat} \nc{\hT}{\h{T}}\nc{\be}{\begin{eqnarray}}
\nc{\ee}{\end{eqnarray}}\nc{\rd}{\textrm{d}}\nc{\e}{eqnarray}\nc{\hR}{\hat{R}}\nc{\Tr}{\mathrm{Tr}}
\nc{\tS}{\tilde{S}}\nc{\tr}{\mathrm{tr}}\nc{\8}{\infty}\nc{\lgs}{\bra\ua,\phi|}\nc{\rgs}{|\ua,\phi\ket}
\nc{\hU}{\hat{U}}\nc{\lfs}{\bra\phi|}\nc{\rfs}{|\phi\ket}\nc{\hZ}{\hat{Z}}\nc{\hd}{\hat{d}}
\nc{\bd}{\bar{d}}\nc{\bc}{\bar{c}}\nc{\mc}{\mathcal}\nc{\ea}{eqnarray}\nc{\bce}{\begin{center}}
\nc{\ece}{\end{center}}

\newcommand{\figurespath}{figs/}

\usepackage{setspace}

\begin{document}


\title{Super-Resolution Reconstruction of Electrical Impedance Tomography Images}


\author{\thanks{This work has been supported by the the National Council for Scientific and Technological Development (CNPq).}
Ricardo A. Borsoi,
\thanks{R. A. Borsoi and J. C. M. Bermudez are with the Department of Electrical Engineering, Federal University of Santa Catarina, Florian\'opolis, Brazil}
\and Julio C. C. Aya,
\thanks{J. C. C. Aya and G. H. Costa are with University of Caxias do Sul, Caxias do Sul, Brazil}
\and Guilherme H. Costa,
\and Jos\'e C. M. Bermudez}









\maketitle


\begin{abstract}
Electrical Impedance Tomography (EIT) systems are becoming popular because they present several advantages over competing systems. However, EIT leads to images with very low resolution. Moreover, the nonuniform sampling characteristic of EIT precludes the straightforward application of traditional image super-resolution techniques. In this work, we propose a resampling based Super-Resolution method for EIT image quality improvement. Results with both synthetic and \textit{in vivo} data indicate that the proposed technique can lead to substantial improvements in EIT image resolution, making it more competitive with other technologies.
\end{abstract}


\section{Introduction}


Real-time imaging systems are particularly important in medicine as they provide useful means of monitoring the patient in intensive care or during medical procedures like intraoperative image guidance.
However, most intraoperative imaging systems are expensive, offer radiation risks and do not operate in real-time~\cite{Helm2015}.
Likewise, real-time monitoring can avoid lung overdistension and collapse in mechanical lung ventilation, reducing ventilator-associated pneumonia rates \cite{Holder04}.

Electrical Impedance Tomography (EIT) is an imaging method that leads to low cost, highly portable, real-time and radiation-free imaging devices. These characteristics make it suitable for bedside diagnosis and intraoperative imaging. However, the lower image quality when compared to other tomography methods still keeps EIT from being widely applied in medical diagnosis and guidance.

The image resolution in tomography methods such as Computed Tomography (CT) and Positron Emission Tomography (PET) has significantly improved with the direct application of Super-Resolution Reconstruction (SRR) techniques. SRR consists basically in the reconstruction of a high resolution (HR) image by extracting non-redundant information from several low resolution (LR) images (usually acquired in the presence of motion) of the same object. SRR was originally a technique intended for overcoming physical limitations of optical imaging sensors. More recently, it has been successfully adapted for other imaging systems \cred{such as CT~\cite{Kennedy06}, PET~\cite{wallach2012PET_SRR}, magnetic resonance~\cite{shi2015MR_SRR} and optoacoustic tomography~\cite{he2016optoacoustic_SRR}, leading to the improvement of the resulting image quality.}

Although SRR algorithms developed for optical systems require some adaptations for super-resolving tomographic images originating from PET or CT, the principle of operation of these systems is quite similar to that of optical image acquisition systems. An EIT system, on the other hand, is based on different operation principles. It employs diffuse electrical currents instead of a coherent X-ray beam to generate images, which makes the image acquisition process inherently nonlinear. Besides, the use of traditional finite element (FE) techniques for solving the EIT inverse problem leads to an image represented by a nonuniform grid of conductivity values, as opposed to the uniformly sampled images acquired by sensor grids used in optical systems. These characteristics preclude the direct application of traditional SRR methods to EIT.

Recently, a new SRR method has been developed for the reconstruction of optical images generated by hypothetical nonuniform pixel arrays called Penrose Pixels~\cite{Ben11}. Penrose tiling is a method of aperiodic tiling that uses different geometric forms at different orientations to fill a plane~\cite{Penrose74}. Rhombus Penrose tiling is a specific type of aperiodic tiling that employs only two types of rhombus of equal side length. To allow the representation of the nonuniform pixel grid using matrix-vector notation, it is proposed in~\cite{Ben11} that the LR images be upsampled to a regular HR grid as soon as they are acquired. The upsampling operation is dependent on the sensor layout map.  The resulting image is called Intermediate HR (IHR) image. After upsampling, image registration and SRR algorithms can be applied on the IHR grid, where the processes of downsampling to and upsampling from the nonuniform grid are incorporated to the reconstruction algorithm by means of an acquisition model. This method has been shown to be capable of achieving larger magnification factors than those obtained using regular pixel arrays, as there are no constraints concerning (sub)pixel displacements between the LR observations because the grid has no translational symmetry. Unfortunately, however, such sensor arrays are currently not being manufactured, and this promising technique remains largely without practical application.


In this paper, a new super-resolution strategy for EIT images is proposed.
By employing the resampling strategy used in~\cite{Ben11} to deal with the aperiodic tiling characteristic of Penrose pixels, we developed a method for super-resolving images generated with arbitrarily irregular tiling such as EIT images, which are composed of non-uniformly spaced pixels forming a finite element mesh (FEM).
The EIT data acquisition system is modeled as in an optical SRR problem with nonuniform sampling, and the image reconstruction is performed independently from the EIT inverse problem. This allows the proposed methodology to be used with arbitrary EIT algorithms and FEMs, which includes commercial EIT systems already in operation.
Quantitative and perceptual evaluations show substantial improvements in the resolution of the reconstructed images when compared to the original ones in LR. This observation remains consistent for different choices of algorithms used in the EIT inverse problem.



This paper is organized as follows. In Section~\ref{sec:prelim_eit_srr} the basic principles of EIT and SRR are explained. In Sections~\ref{sec:proposed_principle} and~\ref{sec:model}, the EIT imaging system is modelled as a nonuniform optical imaging system. Results illustrating the super-resolution of EIT \cblue{images are presented in Section~\ref{sec:results}, along with a discussion about its complexity and robustness.} Finally, Section~\ref{sec:conclusions} concludes this paper.

\section{Operation Principle of EIT and SRR}
\label{sec:prelim_eit_srr}

\subsection{The EIT Working Principle}
\label{sec:prelim_eit}

In electrical impedance tomography, the electrical conductivity $\sigma(\vp)$ inside a domain $\Omega$ is estimated from voltage and current measurements $\{V(\vp),I(\vp)\}$ at its boundary~$\partial\Omega$, where~$\vp$ denotes the spatial position vector.
The physical process relating the electrical conductivity of the domain with the measurements is described by the following partial differential equations~\cite{Holder04}:
%
\begin{align} \label{eq:eit_model_1}
    \begin{array}{ll}
        \emph{div} \,\big(\sigma(\vp)\nabla V(\vp)\big) = 0 \,\text{,}
        & \vp\in\Omega \\[0.2cm]
        \displaystyle{\sigma(\vp) \frac{\partial V(\vp)}{\partial\boldsymbol{\nu}} = j(\vp)} \,\text{,}
        & \vp\in\partial\Omega
    \end{array}
\end{align}
where $\emph{div}(\cdot)$ is the divergence operator, $\boldsymbol{\nu}$ is an outwardly-pointing unitary vector normal to $\partial\Omega$ and $j(\vp)$ is the current density on $\partial\Omega$, satisfying~$\int_{\partial\Omega}j(\vp)d\vp=0$.

Since the voltage and current values are measured through a finite number of electrodes attached to the surface of the body, this discretization effect must be represented in some form~\cite{Cheng1989electrode}. One simple model for the influence of the electrodes is the \textit{gap model}, where the boundary condition of~\eqref{eq:eit_model_1} is described as:
%
\begin{align} \label{EQ:eit_model_ii}
    \sigma(\vp) \frac{\partial V(\vp)}{\partial\boldsymbol{\nu}} =
    \left\{\begin{array}{cl}
        I_l/A_l\,,  &  \vp\in\partial\Omega_{E_l}
        \,\text{,}\quad 
        {l=1,\ldots,L}\\
        0\,, & \vp\notin \cup_{l=1}^L \partial\Omega_{E_l}
    \end{array}\right.
\end{align}
where $L$ is the number of electrodes, and $I_l$ and $A_l$ are, respectively, the current and the area of the $l-$th electrode, which is denoted by $E_l$. The region of the domain boundary $\partial\Omega$ occupied by the $l-$th electrode is denoted by $\partial\Omega_{E_l}$, and the voltage $V(\vp)$ at the center of $\partial\Omega_{E_l}$ is denoted by $V_l$.
The condition~$\int_{\partial\Omega}j(\vp)d\vp=0$ is replaced with $\sum_{l=1}^LI_l=0$, and the integral of $j(\vp)$ over $\partial\Omega_{E_l}$ equals $I_l$.

Besides~\eqref{eq:eit_model_1} and~\eqref{EQ:eit_model_ii}, one must also consider the effect of the high conductivity of the electrodes in the measured voltages.
This can be performed either by simple models like the \textit{shunt model}, which considers $V_l$ to be constant on~$\partial\Omega_{E_l}$, or using more elaborate models which take the contact impedance between the electrode and the body into account~\cite{cheney1999eitreview}.
Along with the choice of a ground level such as $\sum_{l=1}^LV_l=0$, this set of considerations constitute the so-called \textit{complete electrode model}~\cite{cheney1999eitreview}.


With this model, one can fully describe the behavior of the voltages ${\mathbf{V}_{\!F}}=[V_{1},\ldots,V_{L}]$ in the electrodes at the object/domain boundary $\partial\Omega_{E_l}$ given the conductivity of the medium $\sigma(\vp)$ and a set of injected currents~$\mathbf{I}_F=[I_1,\ldots,I_L]$.
%
This process is described in compact form by a nonlinear forward operator defined by $\mathbf{F}({\sigma}(\vp),\mathbf{I}_F)={\mathbf{V}_{\!F}}$, and constitutes the \textit{EIT Forward Problem}~\cite{Holder04}.

Given a set of $L-1$ linearly independent voltage and current measurements $\mathbf{I}_F^k=[I_1^k,\ldots,I_L^k]$
and $\mathbf{V}_{\!F}^k=[V_1^k,\ldots,V_L^k]$, $k=1,\ldots,L-1$, the goal of the \textit{EIT inverse problem} is to find the approximate conductivity $\vy_{\Delta}=\hat{\sigma}(\vp)$ such that the voltages predicted through the forward problem $\widehat{\mathbf{{V}}}=\mathbf{F}(\hat{\sigma}(\vp),\mathbf{I}_F^1,\ldots,\mathbf{I}_F^{L-1})$
\footnote{Note that the forward operator $\mathbf{F}(\sigma(\vp),\mathbf{I}_F^1,\ldots,\mathbf{I}_F^k)$ describes an arbitrary number of sets of voltage measurements $\mathbf{V}_{\!F}^1,\ldots,\mathbf{V}_{\!F}^k$, depending on the number $k$ of input current measurements $\mathbf{I}_F^1,\ldots,\mathbf{I}_F^k$.}
are as close to the actual measurements $\mathbf{V}_M=[\mathbf{V}_{\!F}^1,\ldots,\mathbf{V}_{\!F}^{L-1}]$ as possible~\cite{Holder04}.

Many methods exist for solving the EIT inverse problem (including for example neural networks~\cite{adler1994neural} and the D-Bar method~\cite{isaacson2004Dbar}).
One of the most prominent choices consist of employing Finite Element (FE) techniques for the numerical computation of the EIT forward model based on a discrete approximation of the domain~$\Omega$.
The EIT inverse problem can then be described 
by the following optimization problem~\cite{Holder04}:
\begin{align} \label{eq:eit_opt_prob_i}
\begin{split}
    %
    \vy_{\Delta} {}={} 
    \arg\min_{{\sigma}(\vp)} \,\,
    & \big\|\mathbf{V}_M-\mathbf{F}({\sigma}(\vp),\mathbf{I}_F^1,\ldots,\mathbf{I}_F^{L-1})\big\|^2
    \\ & 
    + \alpha_{\text{EIT}} \, R({\sigma}(\vp))
    \,\text{,}
\end{split}
\end{align}
where $R(\hat{\sigma}(\vp))$ is a regularization functional that includes additional information about the solution. Common choices for the regularization term include the Thikonov and Total Variation regularizations~\cite{borsic2010invivoTV,Holder04}. The scalar $\alpha_{\text{EIT}}$ is the regularization parameter.

In many applications of EIT such as lung monitoring, non-idealities like breathing motion and electrode misplacement can introduce substantial artifacts in images obtained as solutions to optimization problem~\eqref{eq:eit_opt_prob_i}.
Therefore, it is a common practice to employ \textit{linearized difference imaging} in medical applications to reduce the occurrence of these image artifacts~\cite{Holder04}.
In this case, first a set of voltages $\mathbf{V}_{\text{ref}}$ corresponding to a reference conductivity distribution $\sigma_{\text{ref}}(\vp)$ is acquired beforehand. Afterwards, for the measured voltages $\mathbf{V}_M$ we try to reconstruct the conductivity change $\delta{\sigma}(\vp)=\hat{\sigma}(\vp)-\sigma_{\text{ref}}(\vp)$ in the body for a linearized version of the forward operator $F(\cdot)$, corresponding to the optimization problem
\begin{align} \label{eq:eit_opt_prob_ii}
    \vy_{\Delta} {}={} &
    \arg\min_{\delta{\sigma}(\vp)}
    \|\delta\mathbf{V}-\mathbf{J} \delta{\sigma}(\vp)\|^2 
    + \alpha_{\text{EIT}} \, R(\delta {\sigma}(\vp))
    \,\text{,}
\end{align}
where $\delta\mathbf{V}=\mathbf{V}_M-\mathbf{V}_{\text{ref}}$ is the measured voltage difference and $\mathbf{J}$ is the Jacobian of $\mathbf{F}(\cdot)$.
The optimization problems~\eqref{eq:eit_opt_prob_i} and \eqref{eq:eit_opt_prob_ii} can be solved in various ways like, for instance, either by using Newton-Rhaphson or Gauss-Newton techniques~\cite{Holder04}. More advanced techniques are employed in the case of non-smooth regularization terms~\cite{borsic2010invivoTV}.

It is important to note that more advanced methods in EIT image reconstruction consider optimization problems that might differ from~\eqref{eq:eit_opt_prob_i}.
These works mainly search for measures of similarity for predicted and measured voltages, which leads to an increased robustness to modeling errors, and use regularization terms that promote sparsity and steep variations in the solution~\cite{borsic2012primaldualL1L1,borsic2010invivoTV}.



\subsection{Traditional Super-Resolution Reconstruction}
\label{sec:prelim_srr}

The objective of SRR techniques is to obtain one or more images of high resolution (HR) from a set of observed low resolution (LR) images.
This is usually done by solving an inverse problem, which is formulated
based on a mathematical model for the relationship between the different HR and LR images.
The super-resolution of images deriving from conventional imaging sensors has received a lot of interest, and is already a consolidated methodology~\cite{Nasrollahi14}.
In this case, the acquisition process of an image at the $t-$th discrete time instant can be modeled as~\cite{Nasrollahi14}
\begin{equation} \label{eq:aquis}
    \vy(t) = \mD\mH\vx(t) + \ve(t) \,
    \text{,}
\end{equation}
where $\vy(t)$ ($N^2 \times 1$) and $\vx(t)$ ($M^2 \times 1$) are the lexicographic (i.e. vectorized) representations of the degraded (LR) and original (HR) images, respectively.
Matrix $\mD$ ($N^2 \times M^2$) models the subsampling taking place at the sensor, and $\mH$ ($M^2 \times M^2$) models the blurring in the acquisition process. These matrices are assumed to be known in advance.
Vector $\ve(t)$ ($N^2 \times 1$) models the observation noise in the sensor.

In the case of video sequences, the relationship between each pair of HR images can be modelled by~\cite{Elad99}
\begin{equation}\label{eq:dinam}
    \vx(t) = \mG(t) \vx(t-1) + \vs(t)
    \,\text{,}
\end{equation}
where $\mG(t)$ is a warp matrix that represents the relative displacement between the HR images $\vx(t-1)$ and $\vx(t)$ at adjacent time instants.
The motion between the images is estimated from the LR observations using image registration techniques, and can be described either by simple parametric models (such as global rotations and translations) or by a dense motion field, where each pixel of the image can move independently of the others~\cite{Sun10}.
Vector $\vs(t)$ models the innovations in $\vx(t)$, which consists of unpredictable changes happening in the video sequence between time $t-1$ and $t$, i.e. the information in $\vx(t)$ which cannot be represented as a linear combination of the pixels in $\vx(t-1)$, usually due to scene changes such as occlusions.

The reconstruction of the $t-$th image based on $L$ previously observed frames is performed by minimizing some metric of the difference between the degraded HR (estimated) image and the LR observations.
For the case of the squared L$_2$ norm, the minimization problem is formulated as~\cite{Nasrollahi14}:
\begin{align}
    \vxh(t) {}={} & \arg\min_{\vx(t)} \,\, \Big\{ \sum_{l=0}^{L-1} \|\vy(t-l)-\mD\mH\,\mM_{t-l,t}\,\vx(t)\|^2
    \nonumber\\ & \hspace{2.5cm}
    + \alpha_{\text{SRR}}\,\|\mS\vx(t)\|^2 \Big\}
\end{align}
where $\mM_{t-l,t}$ represents the total motion from the $t-$th to the $(t-l)-$th image (which is obtained from the dynamical model~\eqref{eq:dinam}).
The last term is a regularization that incorporates additional \textit{a priori} information about the desired solution.
This is usually performed by assuming that the images are smooth, which results in the matrix $\mS$ being a high pass filter (e.g. a Laplacian). The scalar $\alpha_{\text{SRR}}$ is the regularization parameter.

Note that the SRR process described considers imaging sensors consisting of regular arrays of square pixels.
EIT images from a finite element mesh, on the other hand, consist of a nonuniform array of irregular elements.
This renders the direct application of the SRR methodology described above unfeasible.
In the following, we shall represent the EIT image formation process as an optical imaging system with nonuniform pixels.
This will allow for the application of a SRR methodology previously designed for the reconstruction of images originating from hypothetical sensors composed by Rhombus Penrose tiling~\cite{Ben11,Penrose74}.

\begin{figure*}
    \centering 
    \includegraphics[width=13cm]{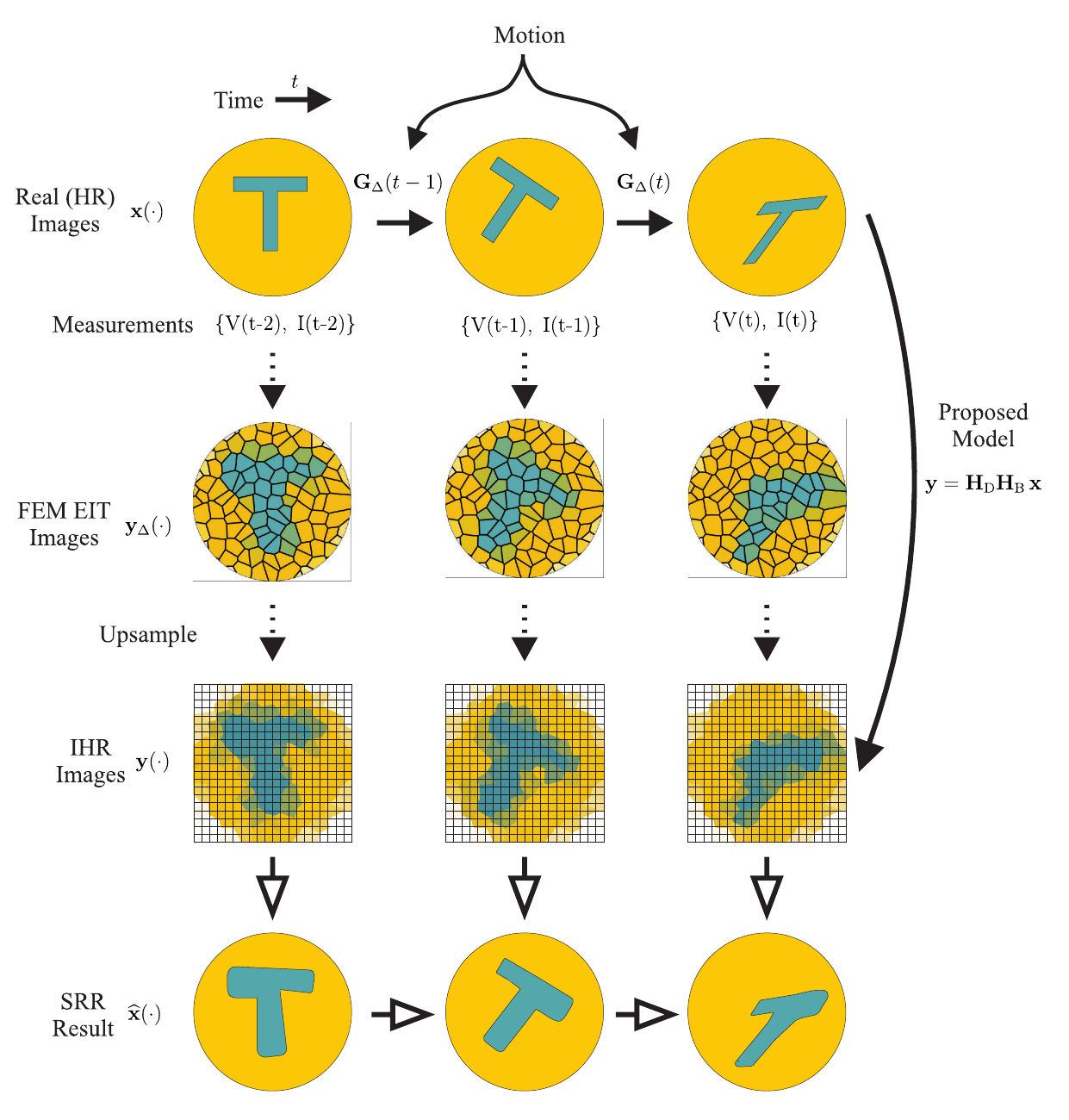}
    \caption{Illustration of the proposed SRR methodology for EIT images.
    First row: real (HR) images of a physical body subject to motion, from which the voltage and current measurements are derived.
    Second row: The EIT (LR) images are reconstructed from the measurements using FEM techniques in the EIT inverse problem.
    Third row: The LR FEM images are upsampled to an uniform grid, resulting in the IHR images.
    Fourth row: The proposed SRR technique is applied to estimate the HR images from the IHR observations.}
    \label{fig:proposed_method_general}
\end{figure*}

\section{Resampling-based Super-Resolution for EIT}
\label{sec:proposed_principle}

The EIT image formation process, which is derived from the EIT inverse problem in~\eqref{eq:eit_opt_prob_i} or~\eqref{eq:eit_opt_prob_ii}, does not resembles an optical image acquisition system as described in~\eqref{eq:aquis}.
In this case, each acquired image consists of a nonuniform mesh (i.e. a finite element mesh - FEM) composed of elements of arbitrary sizes and shapes, each one presenting a constant conductivity value. Moreover, the diffuse nature of the electrical currents injected in the boundary of the domain and the ill-posedness of the inverse problem inflict severe distortions on the resulting image. We will assume these distortions to be appropriately represented by a combination of a space invariant blurring and a nonlinear, spatially variant distortion. Hence, the typical mathematical modeling of the EIT image formation is quite different from the model used in optical SRR algorithms.

In this work we propose to model the EIT image acquisition process using a formulation similar to that used for traditional image systems. An overall illustration of the proposed methodology can be seen in Figure~\ref{fig:proposed_method_general}. By considering each element of the finite element mesh as a pixel of a nonuniform sensor, we propose to employ the Penrose SRR strategy to super resolve the low resolution tomography images. The EIT images are upsampled from the finite element mesh into a uniform grid (IHR) before being processed by a super resolution algorithm.
For the $t-$th time instant, this process is represented as $\vy^{\uparrow}(t)=\uparrow\vy_{\Delta}(t)$, where $\vy^{\uparrow}(t)$ denotes the upsampled image and $\uparrow(\cdot)$ the usampling operator.
This operator can be mathematically represented as
\begin{align}
    y_i^{\uparrow}(t) {}={} \vp_i^\top \vy_{\Delta}(t) \,, \quad i=1,\ldots,M
\end{align}
where $\vp_i$ is a binary vector with zeros in all positions except the one corresponding to the nonuniform pixel (finite element) that contains the $i$-th pixel in the IHR grid.

\begin{figure}
\centering{\scalebox{.375}{\includegraphics{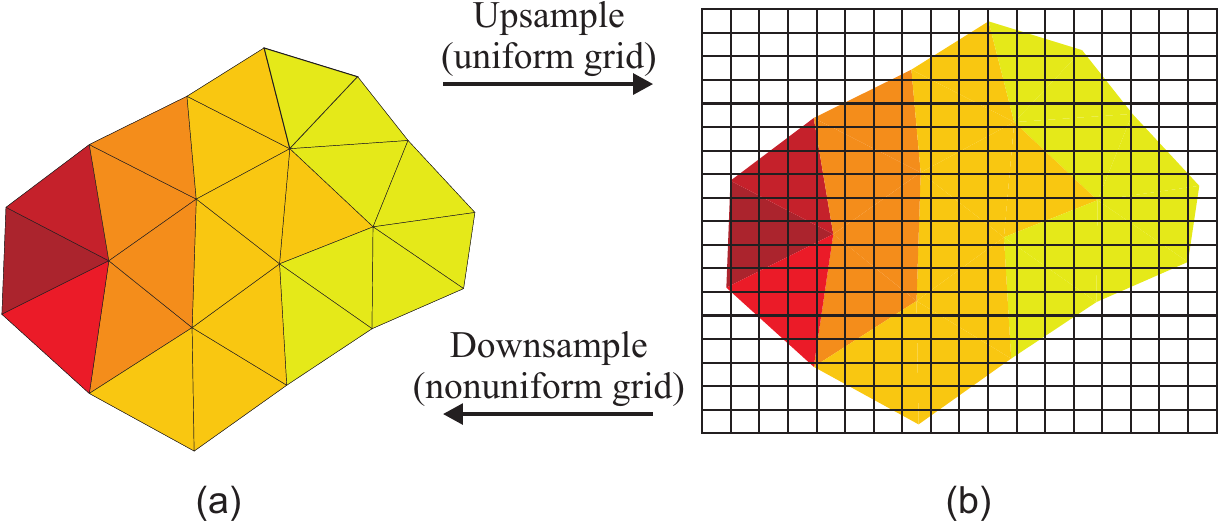}}}
\caption{Illustration of the resampling process. (a) EIT image given by a nonuniform mesh, with constant conductivity value per element. (b) IHR image on the uniform grid. The upsampling process consists of sifting the finite element mesh in order to generate an uniform IHR. The downsample process averages every pixel on the uniform grid that falls within the area of a single element of the mesh, assigning this value for the corresponding element on the nonuniform grid.}
\label{fig:resampling}
\end{figure}

As will be discussed later, the reconstruction method also requires the conjugate downsampling process to be performed. It consists on the transition from the IHR to the LR grid, which is done by assigning to each element of the nonuniform mesh the average value of all the pixels within the corresponding area of the IHR grid.
\cred{Mathematically, this is given by
\begin{align}
    y_{\Delta_p}(t) {}={} \frac{1}{N_{\Delta_p}} \sum_{i\in\Delta_p} y_i^{\uparrow}(t) \,,
    \quad p=1,\ldots,P
\end{align}
where $N_{\Delta_p}$ is the number of IHR pixels inside the $p$-th nonuniform pixel $\Delta_p$.} \cred{The upsampling and downsampling} process is illustrated in Figure~\ref{fig:resampling}.


It should be noted that by assigning a position to the center of each pixel in the regular IHR grid (which encloses the whole FEM domain), the value of each IHR pixel can be computed straightforwardly during the upsampling process by determining which element of the FEM circumscribes its position. This is performed using the geometric information about the FEM, which is available with the LR EIT images~\cite{Holder04}.
Since the position of the center of each IHR pixel $y_{i}^{\uparrow}(t)$ lies inside an unique LR image element $y_{\Delta_p}(t)$ for some $p$, the upsampling process is identical to the uniform impulsive sampling of the nonuniform LR grid, where we assign to $y_{i}^{\uparrow}(t)$ the value of~$y_{\Delta_p}(t)$.



We also propose to model the non-linear distortions caused by the inverse problem using a linear filtering approximation (such as the typical optical blurring model), thereby only addressing the space invariant and more well-behaved part of the blurring process. Although the nonlinear and spatially varying portion of the distortion is not accounted for, it shall nevertheless be shown that the resulting SRR performance employing this model is good enough for practical purposes.
%
%
This new formulation allows the direct application of image SRR techniques to EIT images.



\section{The Proposed EIT Imaging Model} \label{sec:model}

The proposed model for the EIT image formation can be mathematically described as
\begin{equation} \label{eq:image}
    \vy^{\uparrow}(t) = \mH_{\tiny{\text{D}}}(t) \mH_{\tiny{\text{B}}}(t) \vx(t)
    \,\text{,}
\end{equation}
where $t$ is the temporal index, $\vx(t)$ is the lexicographically ordered desired uniform HR image, and the vector $\vy^{\uparrow}(t)$, with the same dimension of $\vx(t)$, is the lexicographically ordered uniform IHR image. Notice that the EIT LR image is the downsampled IHR image $\vy_{\triangle}(t) = \downarrow \vy^{\uparrow}(t)$, where $\downarrow(\cdot)$ represents the downsampling operation illustrated in Figure~\ref{fig:resampling}. The resulting LR image, $\vy_{\triangle}(t)$, is a mesh composed of nonuniformly distributed elements (pixels), each with constant conductivity.

The matrix $\mH_{\tiny{\text{B}}}(t)$ models the EIT distortion with the linear approximation previously discussed, which can be computed in the form of a convolution operator.
Matrix $\mH_{\tiny{\text{D}}}(t)$ is a spatially variant kernel convolution matrix performing the transformation from the HR to the IHR grid. Differently from the decimation performed in traditional SRR, $\mH_{\tiny{\text{D}}}(t)$ in~\eqref{eq:image} is a square matrix which is constructed based on the finite element mesh. When processed by this operator, each pixel $y_{i}^{\uparrow}(t)$ of the resulting IHR image lying within the area of the $p$-th nonuniform LR image element $y_{\triangle_{p}}(t)$ gets assigned the value corresponding to the average of all pixels in the HR image $x_j(t)$ that falls within the area of that specific element of the nonuniform LR image.
%
%
%
This process is illustrated in Figure~\ref{fig:resampling}, where matrix $\mH_{\tiny{\text{D}}}(t)$ corresponds to downsampling the HR image to the LR grid and upsampling the resulting LR image back to the IHR grid, which basically amounts to averaging the pixel values in each LR region.

The relationship between the HR images is described in the uniform HR grid, and does not depend on the finite element mesh or reconstruction algorithm employed in the EIT inverse problem. Therefore, it can be described in the same way as in the case of conventional SRR, by using \cred{a dynamical model of the form
\begin{equation} \label{eq:dinam2}
    \vx(t) = \mG(t) \vx(t-1) + \vs(t)
    \,\text{,}
\end{equation}
where $\mG(t)$ and $\vs(t)$ are the registration matrix \cblue{and the innovations vector} in the same way as described in~\eqref{eq:dinam}.}
In this case, the image registration (motion estimation) is performed performed based on the IHR images~\cite{Sun10}.

Since the imaging model proposed for the EIT system given by~\eqref{eq:image} and~\eqref{eq:dinam2} is now in the same form as the models encountered for traditional optical systems, traditional SRR algorithms can be applied to super-resolve the EIT images using the proposed observation model and the upsampled EIT images as input, resulting in a large flexibility for the choice of the image reconstruction method. 

The super resolution process can be performed both in an offline setting as reviewed in Section~\ref{sec:prelim_srr}, where the cost function becomes of the form of~\cite{Nasrollahi14,elad1997srr_simple}
\begin{align} 
    & \vxh(t) {}={}
    \arg\min_{\vx(t)} \,\, \Big\{ \alpha_{\text{SRR}}\,\|\mS\vx(t)\|^2
    \\ & \hspace{0.5cm}
    + \sum_{l=0}^{L-1} \|\vy(t-l)
    -\mH_{\tiny{\text{D}}}(t-l) \mH_{\tiny{\text{B}}}(t-l)\,\mM_{t-l,t}\,\vx(t)\|^2
    \Big\}
    \nonumber
\end{align}
or as an iterative image recontruction problem, which is based on a cost function of the form~\cite{Nasrollahi14,Elad99,borsoi2017srr_conf}:
\begin{align} \label{eq:cost_function_illustrative_vid}
    \vxh(t) & {}={}  \arg\min_{\vx(t)} \,\, \Big\{\|\vy(t)-\mH_{\tiny{\text{D}}}(t) \mH_{\tiny{\text{B}}}(t)\,\mM_{t-l,t}\,\vx(t)\|^2 
\nonumber\\ & \hspace{0.0cm}
    + \alpha_{\text{T}} \|\vx(t) - \mG(t) \vx(t-1)\|^2
    + \alpha_{\text{SRR}}\,\|\mS\vx(t)\|^2 \Big\}
\end{align}
which is solved iteratively for $t=1,\ldots,T$, and is more suitable for online processing of EIT video sequences. The second term in~\eqref{eq:cost_function_illustrative_vid} consists of a temporal regularization, and is responsible for including information from previous images in the reconstruction of~$\vxh(t)$~\cite{borsoi2017srr_conf}.

\newcommand{\sizeA}{2.3}
\newcommand{\sizeAhlf}{1.15}
\begin{figure*}[!htb] 
\begin{center}
\begin{tabular}{ccc}
  \includegraphics[width=\sizeA cm]{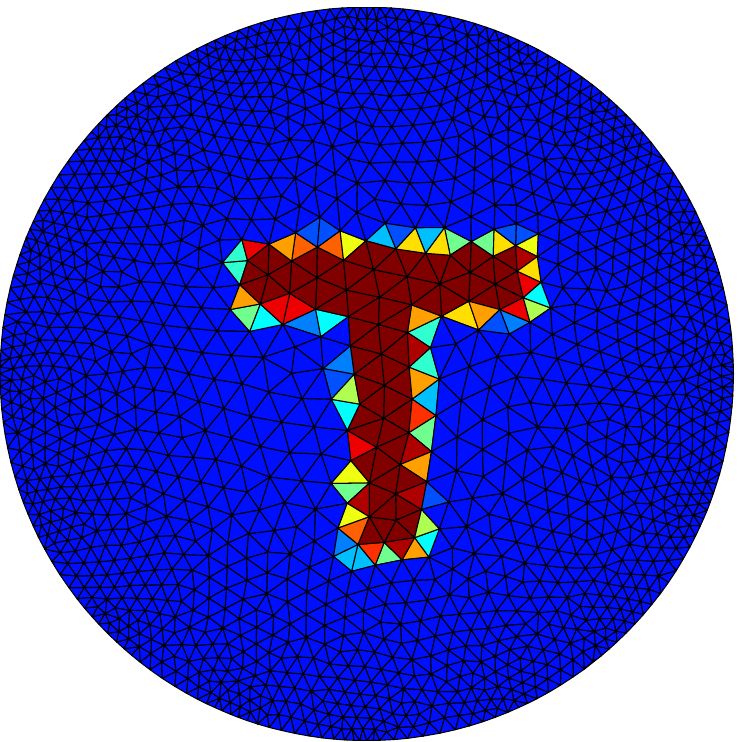}
  \hspace{\sizeAhlf cm}
  &
  \hspace{\sizeAhlf cm}
  \includegraphics[width=\sizeA cm]{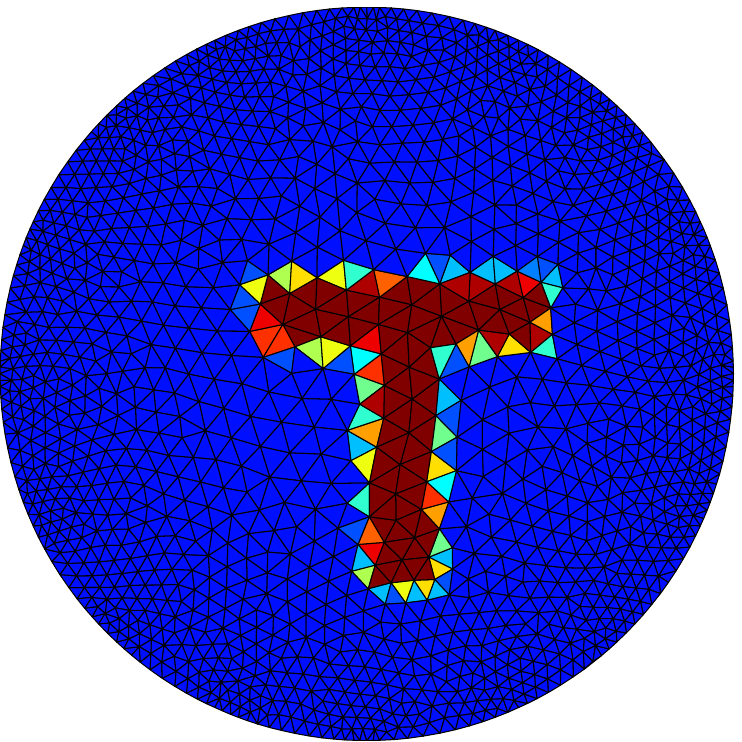}
  \hspace{\sizeAhlf cm}
  &
  \hspace{\sizeAhlf cm}
  \includegraphics[width=\sizeA cm]{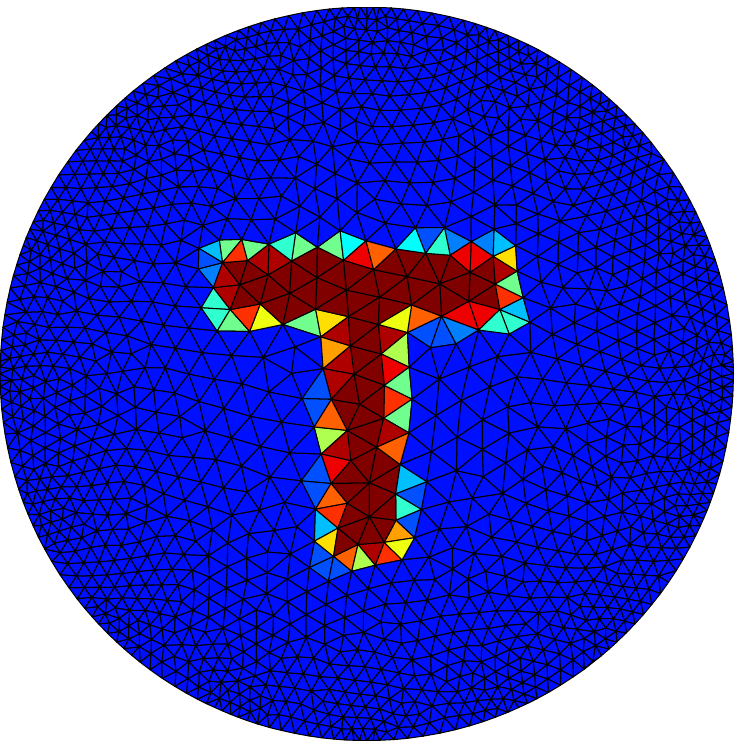}
  \\
  t=10 \hspace{\sizeAhlf cm} & \hspace{\sizeAhlf cm} t=15 \hspace{\sizeAhlf cm} & \hspace{\sizeAhlf cm} t=20
\end{tabular}
\begin{tabular}{cc||cc||cc}
  \includegraphics[width=\sizeA cm]{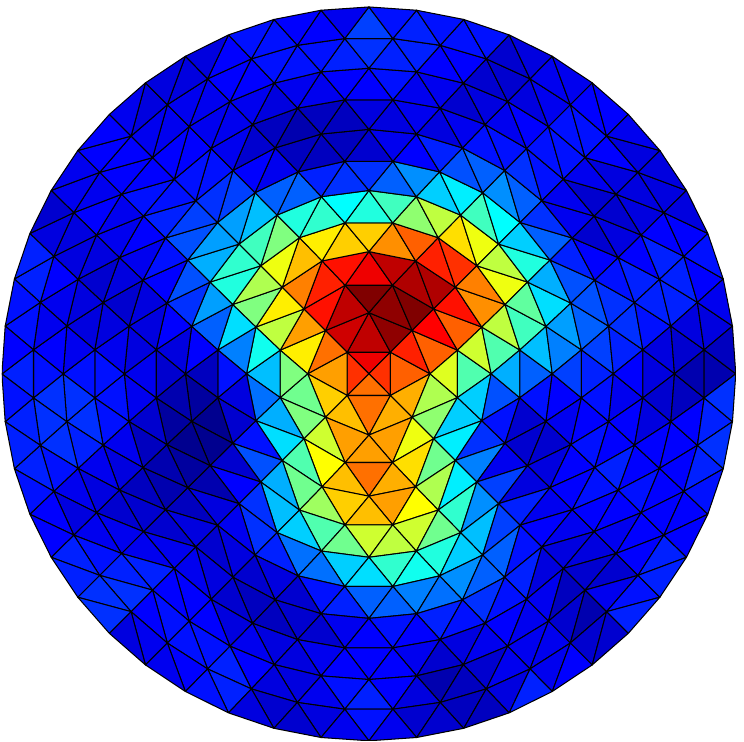}  &
  \includegraphics[width=\sizeA cm]{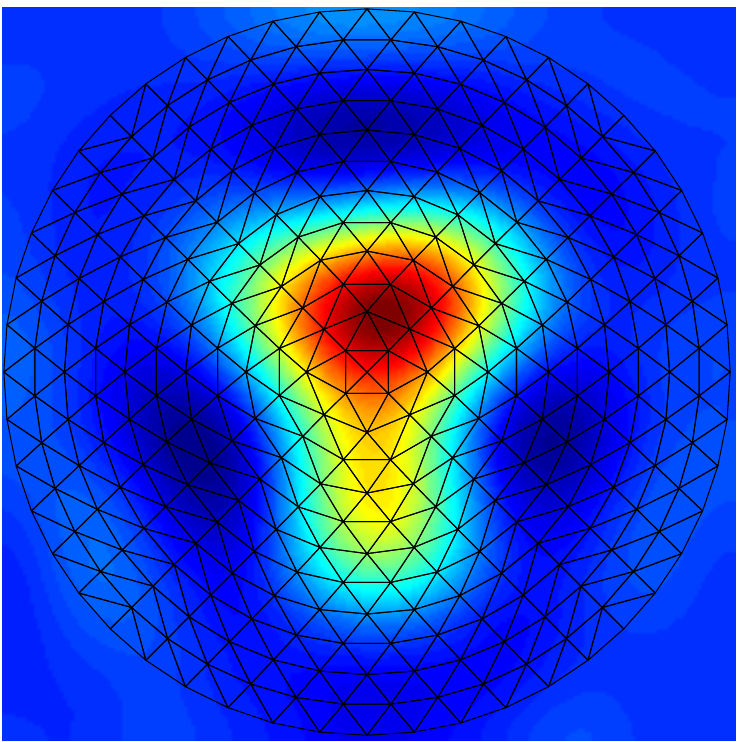}
  &
  \includegraphics[width=\sizeA cm]{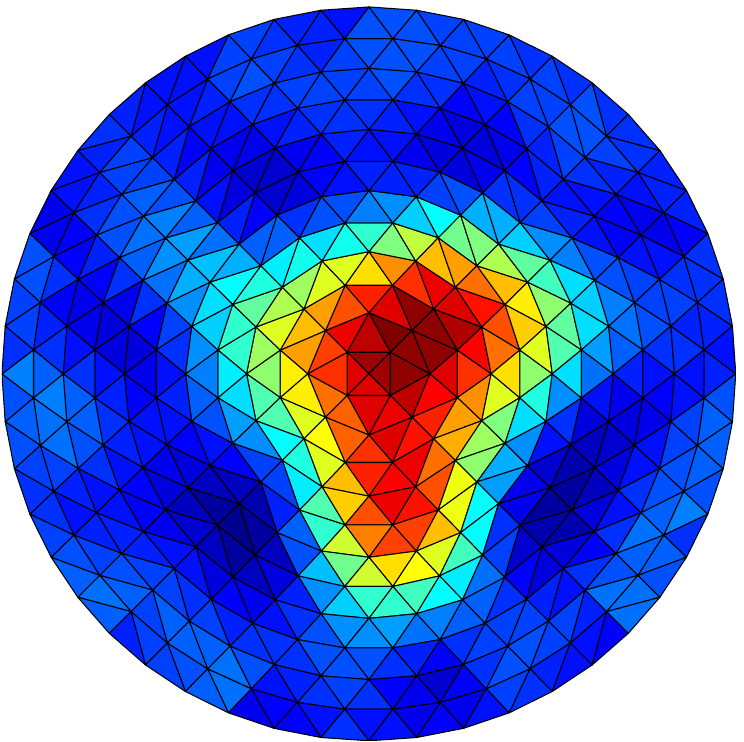}  &
  \includegraphics[width=\sizeA cm]{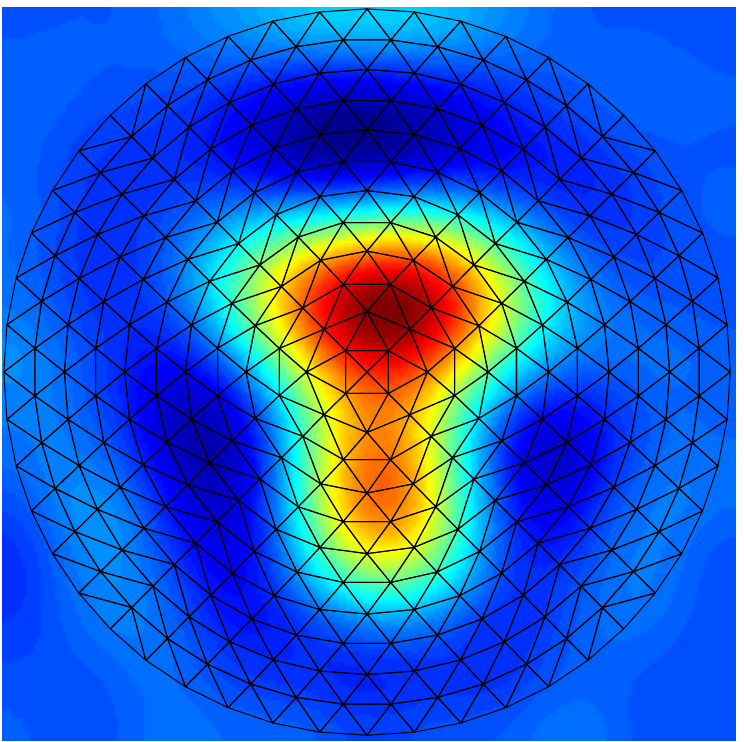}
  &
  \includegraphics[width=\sizeA cm]{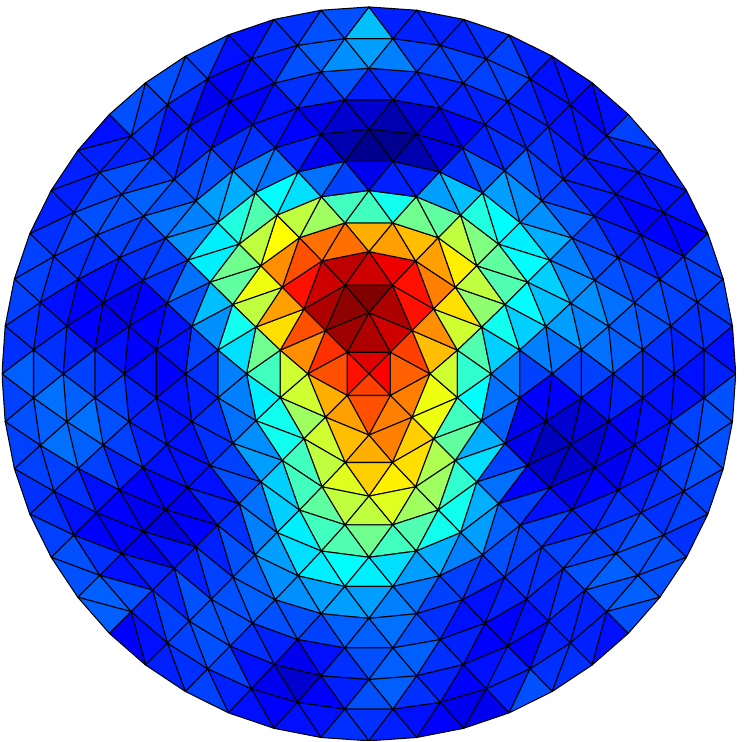}  &
  \includegraphics[width=\sizeA cm]{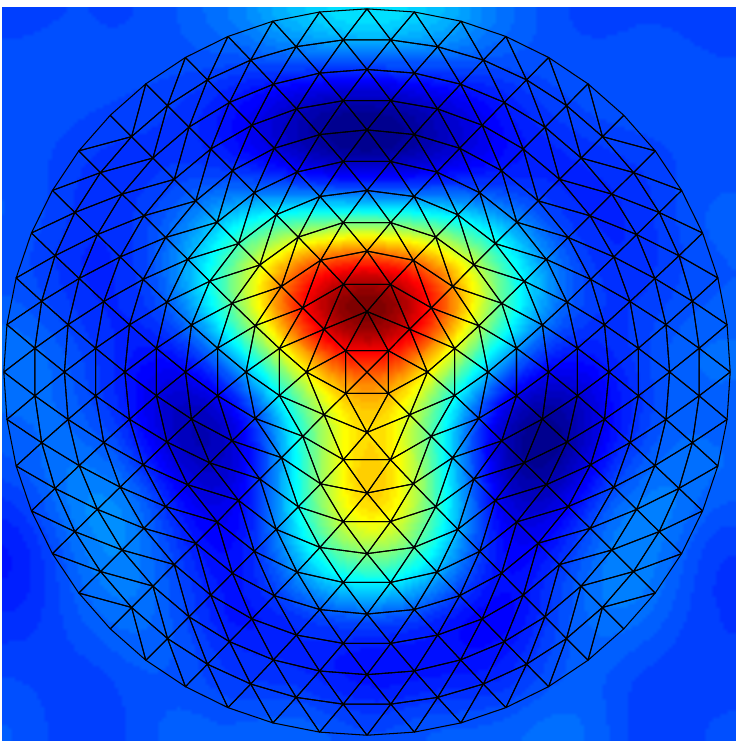} \\
  \includegraphics[width=\sizeA cm]{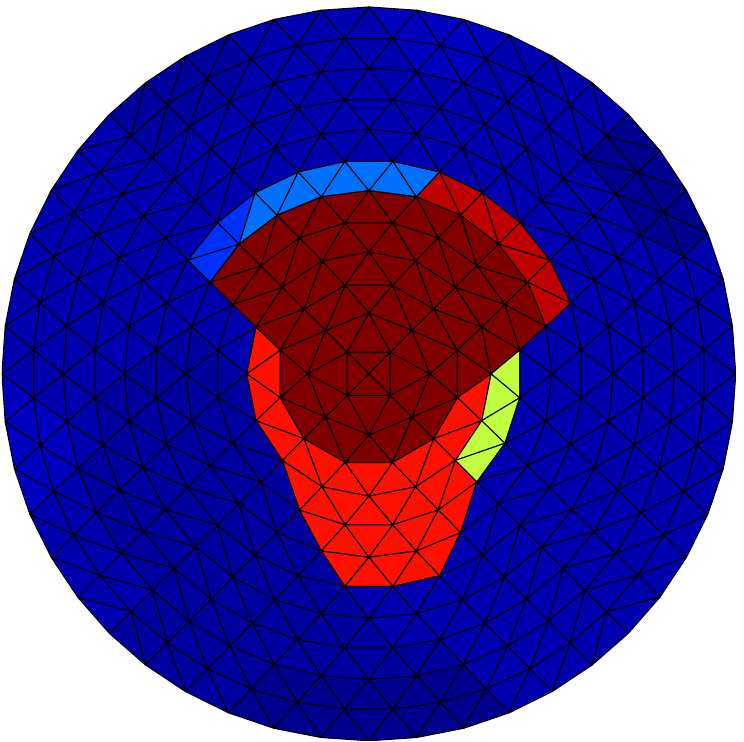}  &
  \includegraphics[width=\sizeA cm]{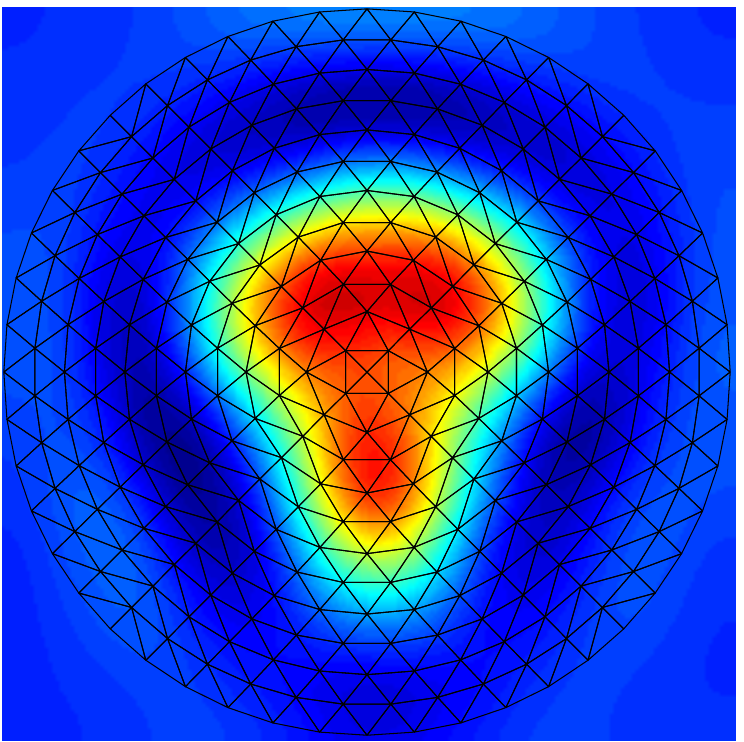}
  &
  \includegraphics[width=\sizeA cm]{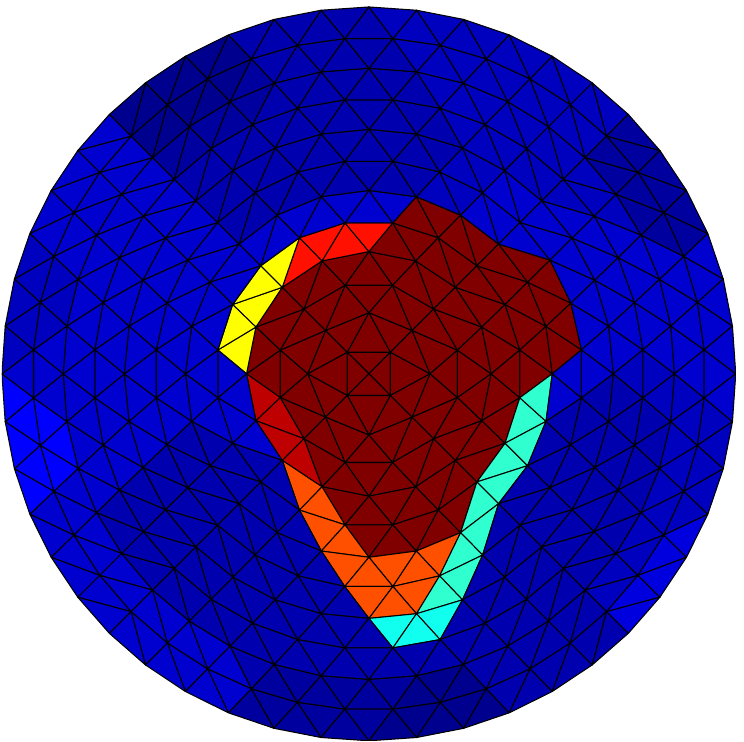}  &
  \includegraphics[width=\sizeA cm]{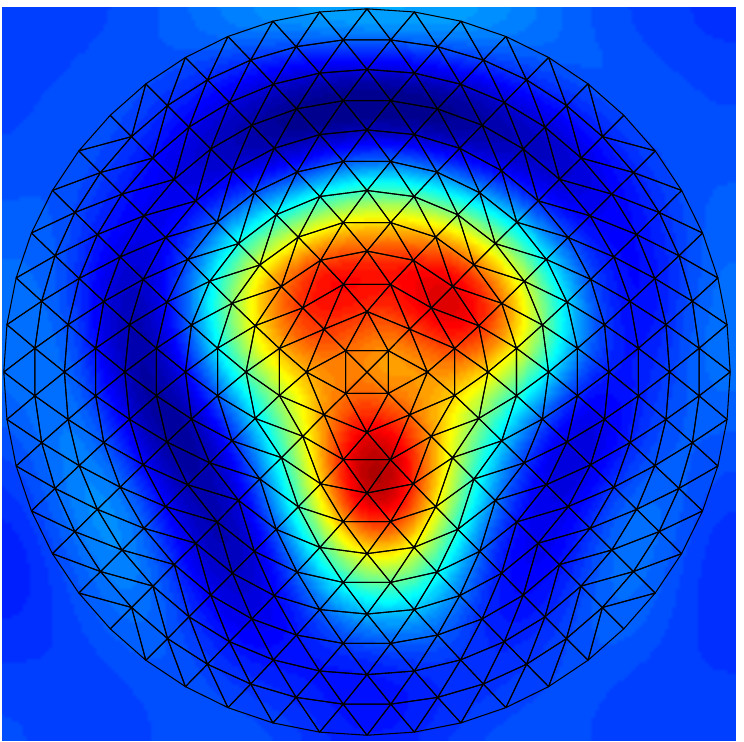}
  &
  \includegraphics[width=\sizeA cm]{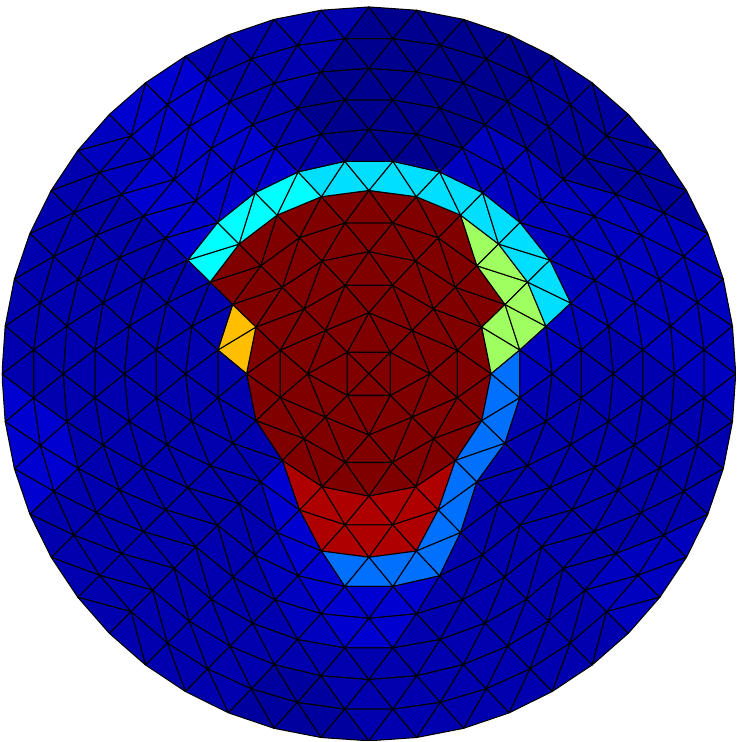}  &
  \includegraphics[width=\sizeA cm]{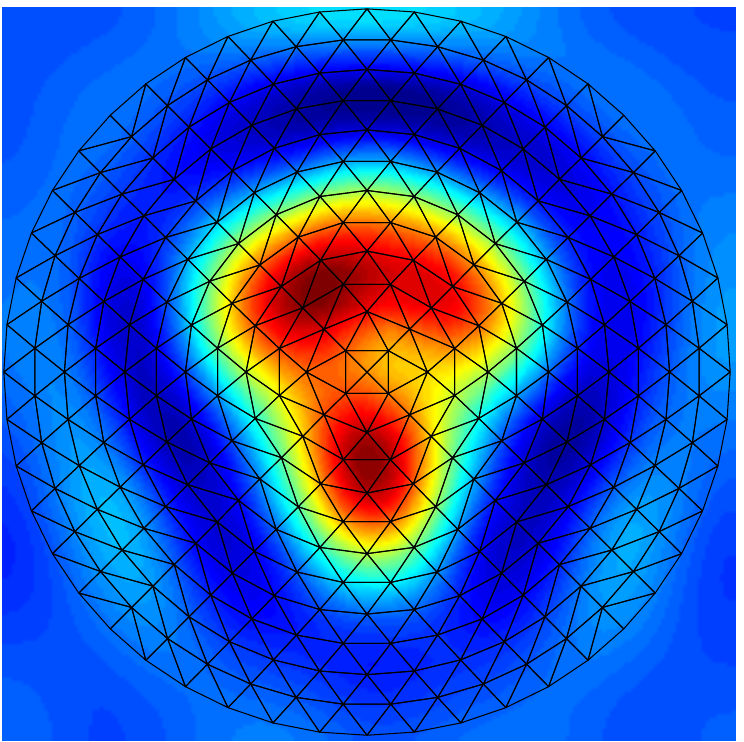} \\
  \includegraphics[width=\sizeA cm]{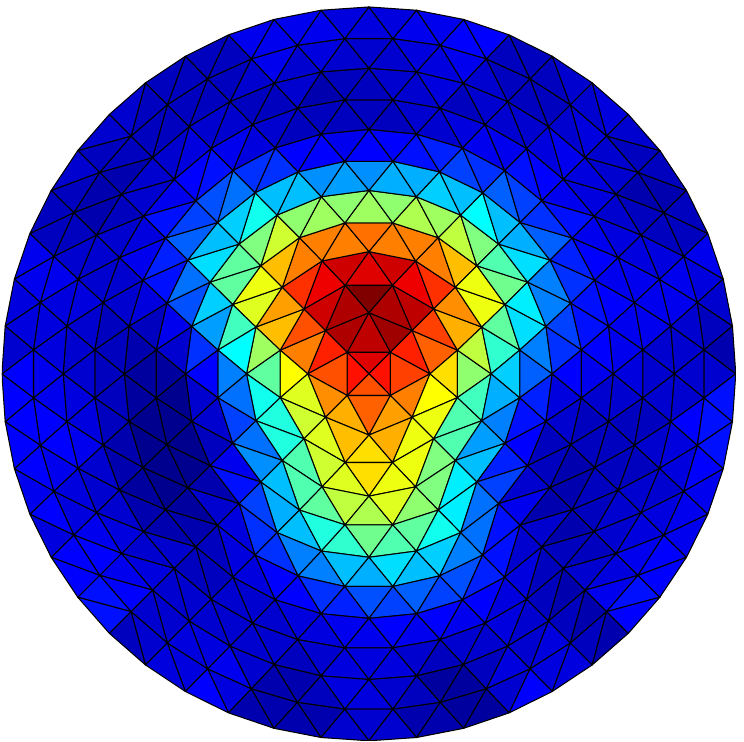}  &
  \includegraphics[width=\sizeA cm]{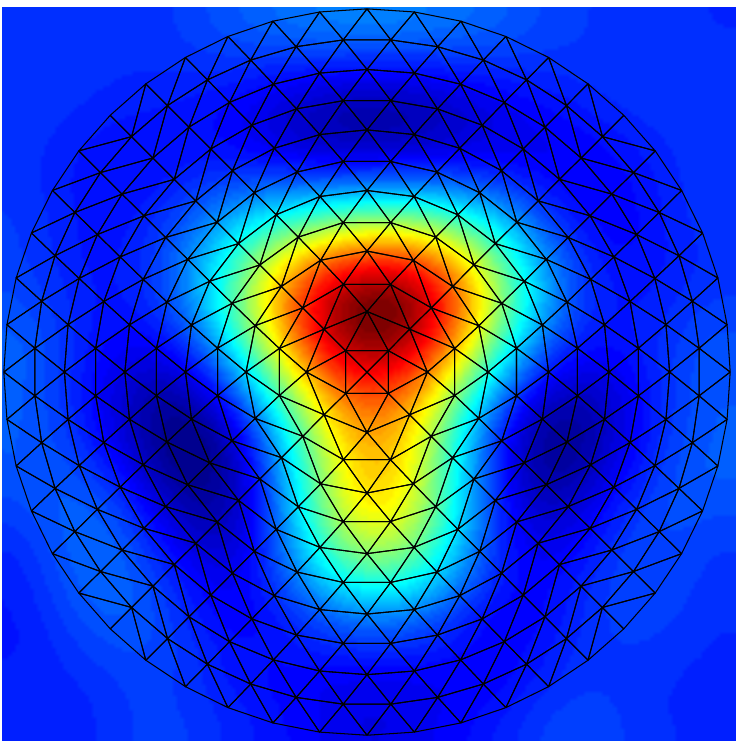}
  &
  \includegraphics[width=\sizeA cm]{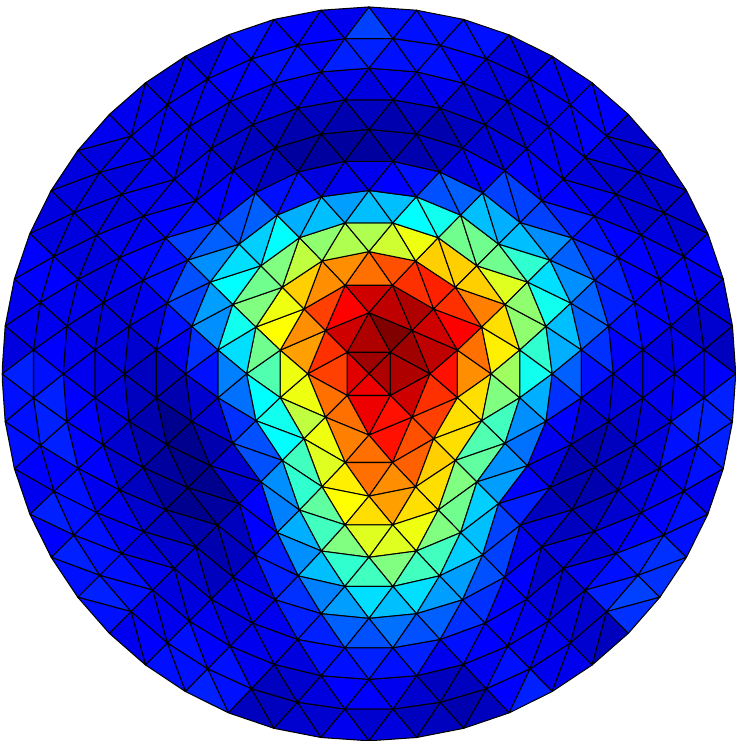}  &
  \includegraphics[width=\sizeA cm]{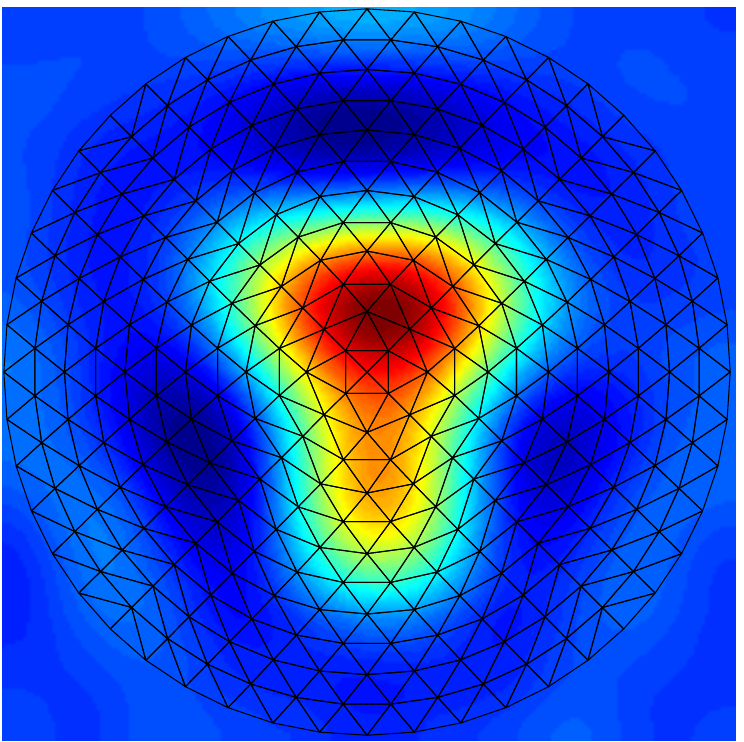}
  &
  \includegraphics[width=\sizeA cm]{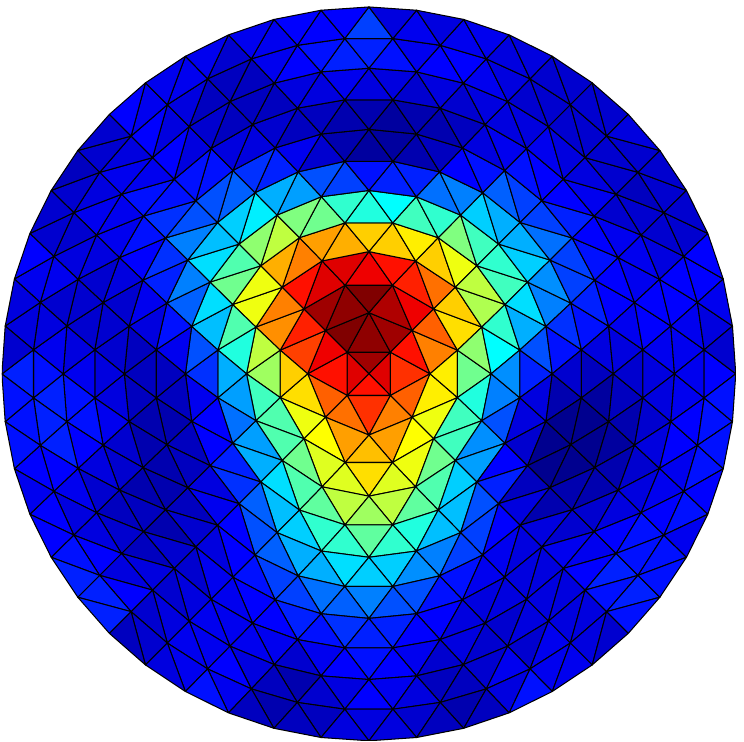}  &
  \includegraphics[width=\sizeA cm]{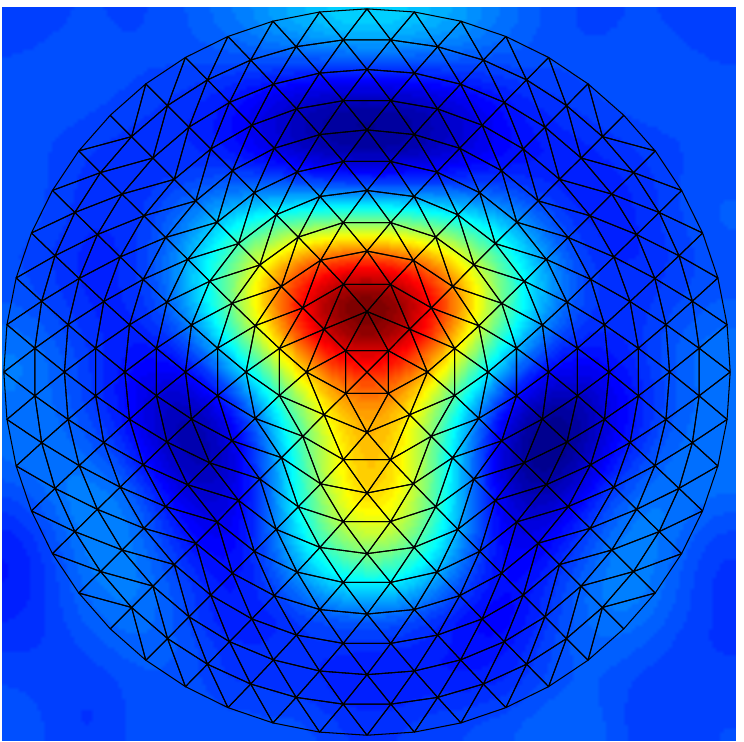} \\
  LR   &   SRR   &   LR   &   SRR   &   LR   &   SRR
\end{tabular}
\end{center}
\caption{Example 1-a: Translating T-shaped object with high measurement SNR and finer FEM. Column groups indicate time instants (t=10, t=15, t=20). First row: Synthetic HR (desired) images (used in the EIT direct problem). Second, third and fourth row: LR EIT image and super resolved results (side by side) for the NOSER, TV and TS algorithms, respectively. Due to space limitations, only the super-resolved images considering motion estimated from the LR observations is displayed.}
\label{ex1a_results}
\end{figure*}

\renewcommand{\sizeA}{2.3}
\renewcommand{\sizeAhlf}{1.15}
\begin{figure*}[!htb] 
\begin{center}
\begin{tabular}{ccc}
  \includegraphics[width=\sizeA cm]{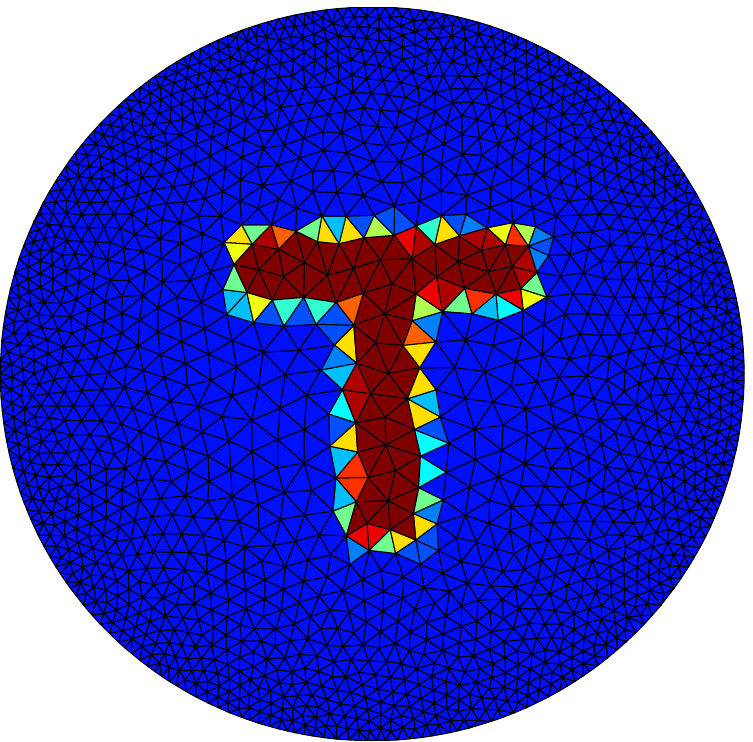}
  \hspace{\sizeAhlf cm}
  &
  \hspace{\sizeAhlf cm}
  \includegraphics[width=\sizeA cm]{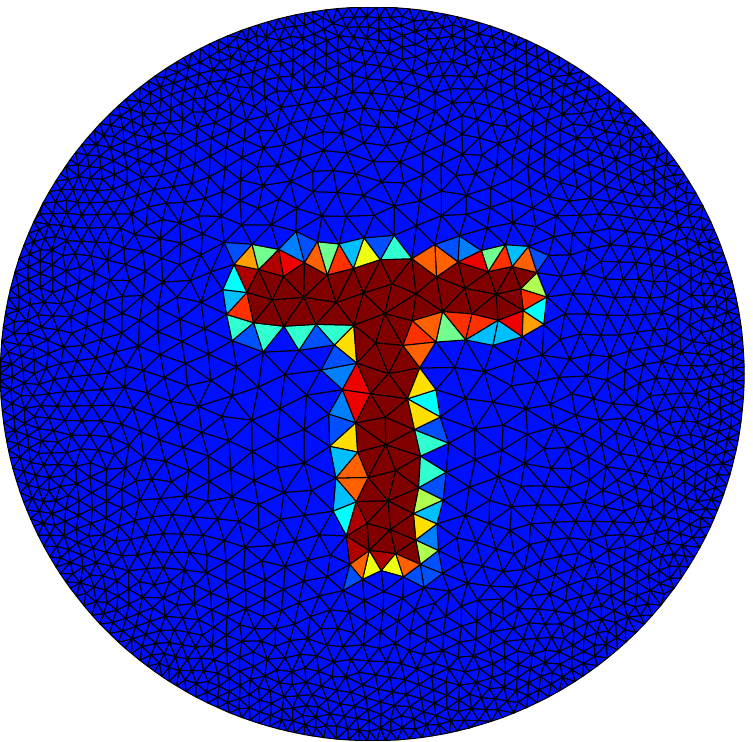}
  \hspace{\sizeAhlf cm}
  &
  \hspace{\sizeAhlf cm}
  \includegraphics[width=\sizeA cm]{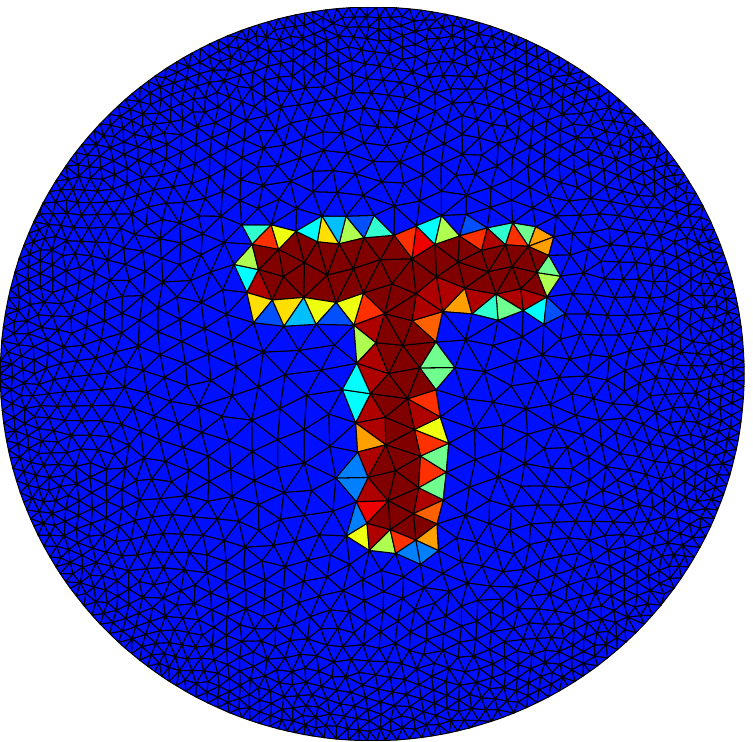}
  \\
  t=10 \hspace{\sizeAhlf cm} & \hspace{\sizeAhlf cm} t=15 \hspace{\sizeAhlf cm} & \hspace{\sizeAhlf cm} t=20
\end{tabular}
\begin{tabular}{cc||cc||cc}
  \includegraphics[width=\sizeA cm]{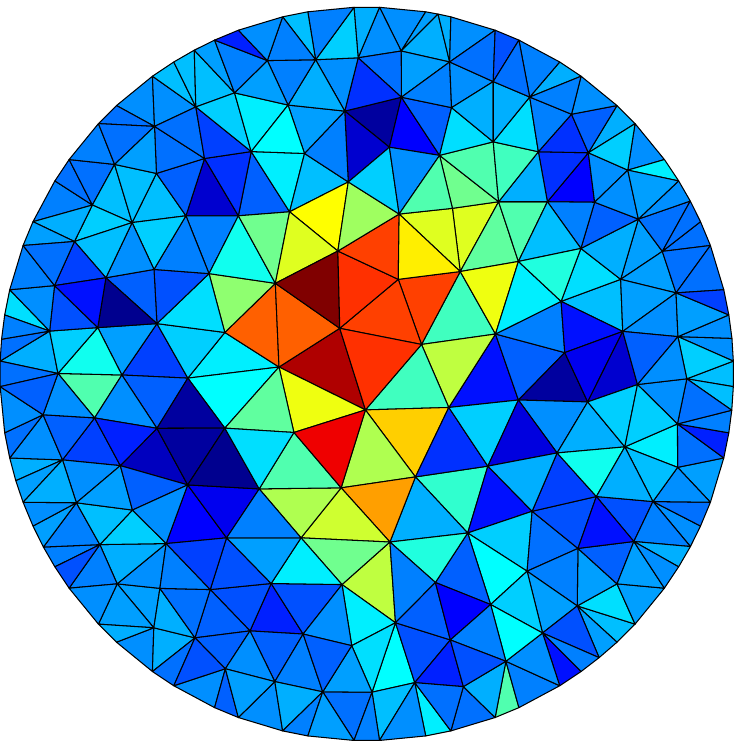}  &
  \includegraphics[width=\sizeA cm]{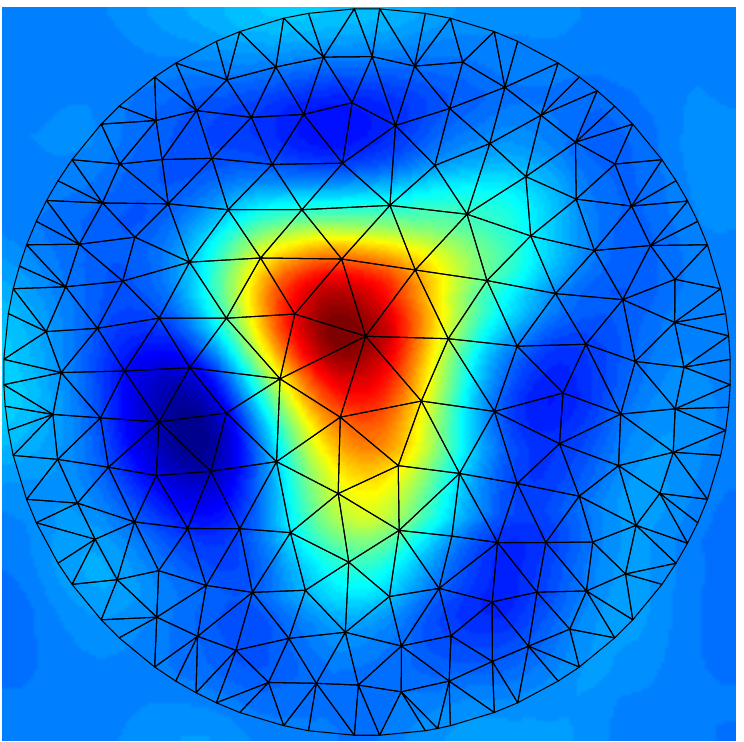}
  &
  \includegraphics[width=\sizeA cm]{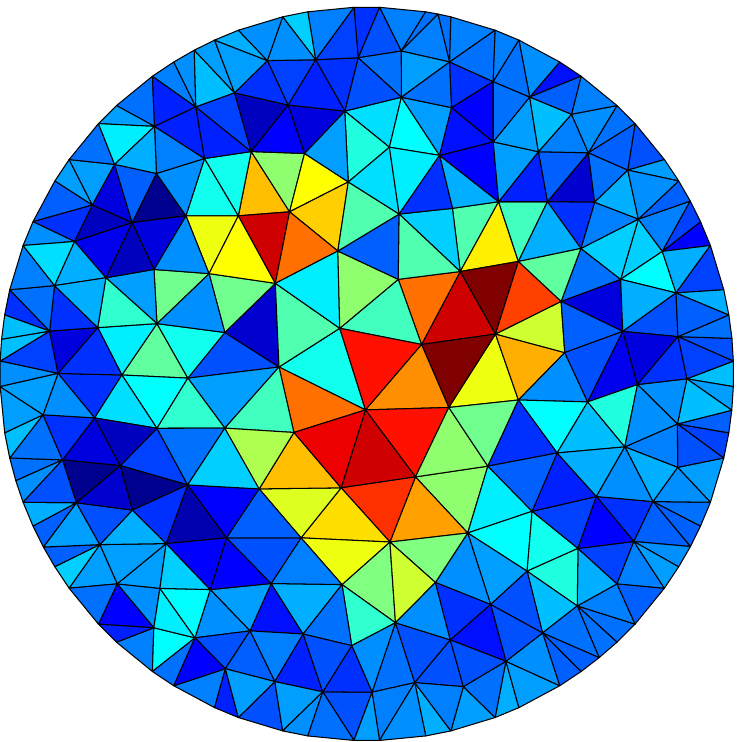}  &
  \includegraphics[width=\sizeA cm]{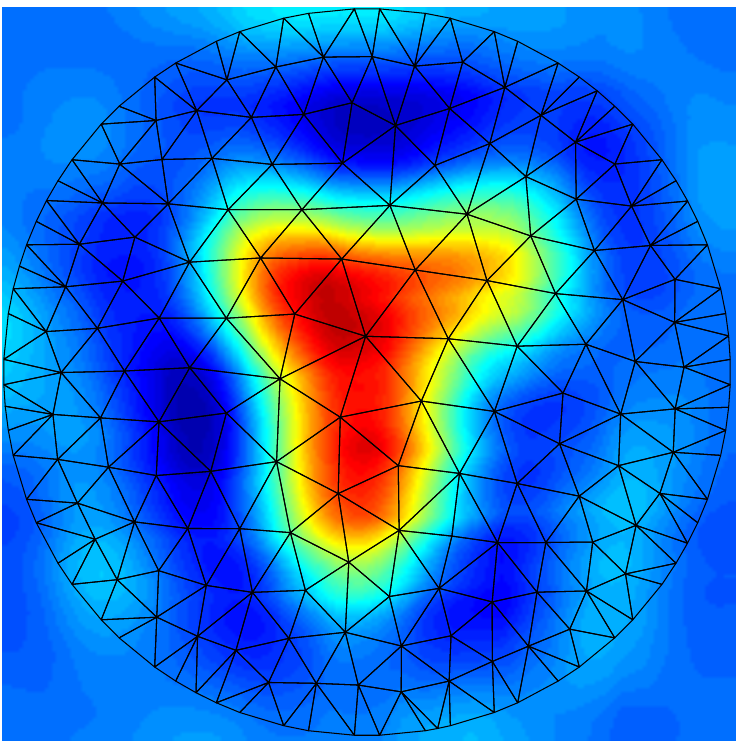}
  &
  \includegraphics[width=\sizeA cm]{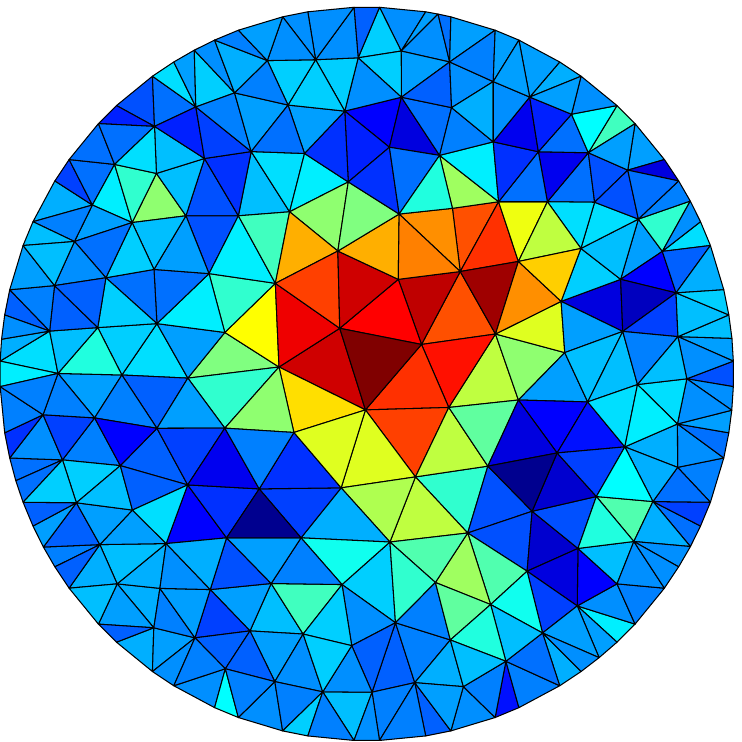}  &
  \includegraphics[width=\sizeA cm]{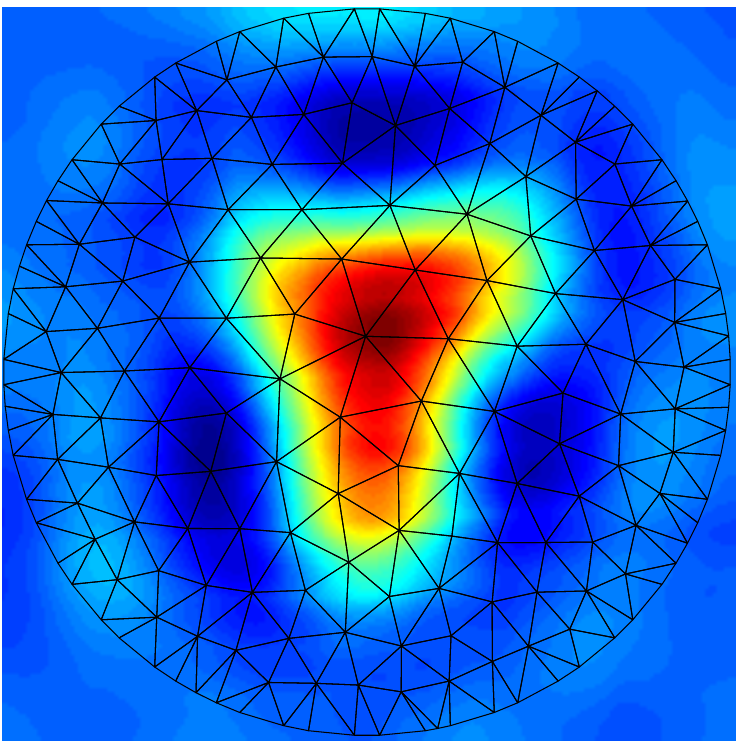} \\
  \includegraphics[width=\sizeA cm]{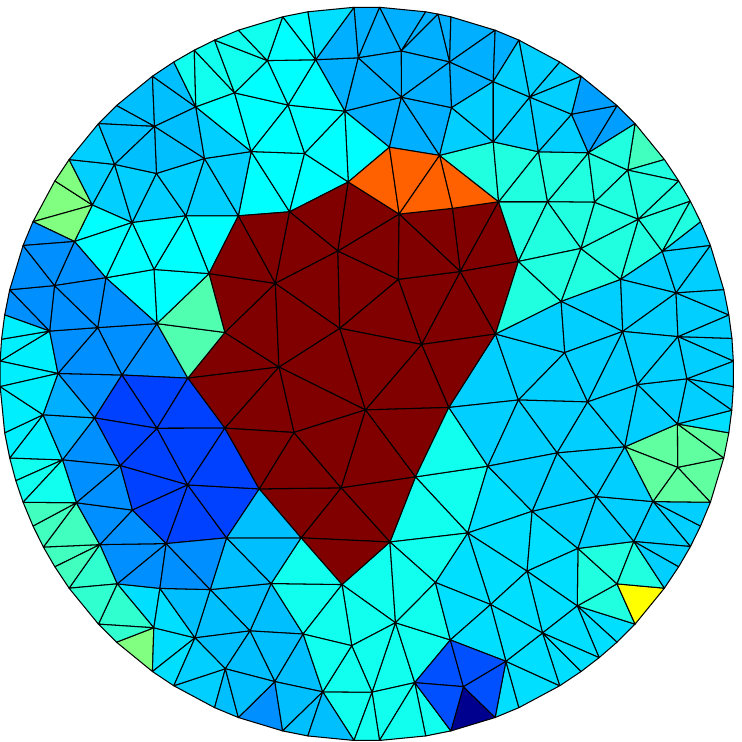}  &
  \includegraphics[width=\sizeA cm]{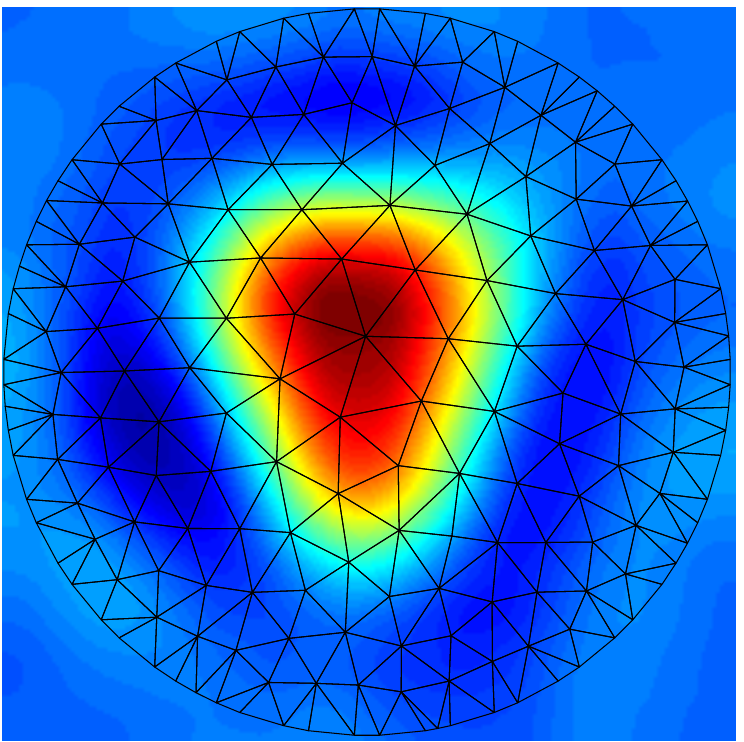}
  &
  \includegraphics[width=\sizeA cm]{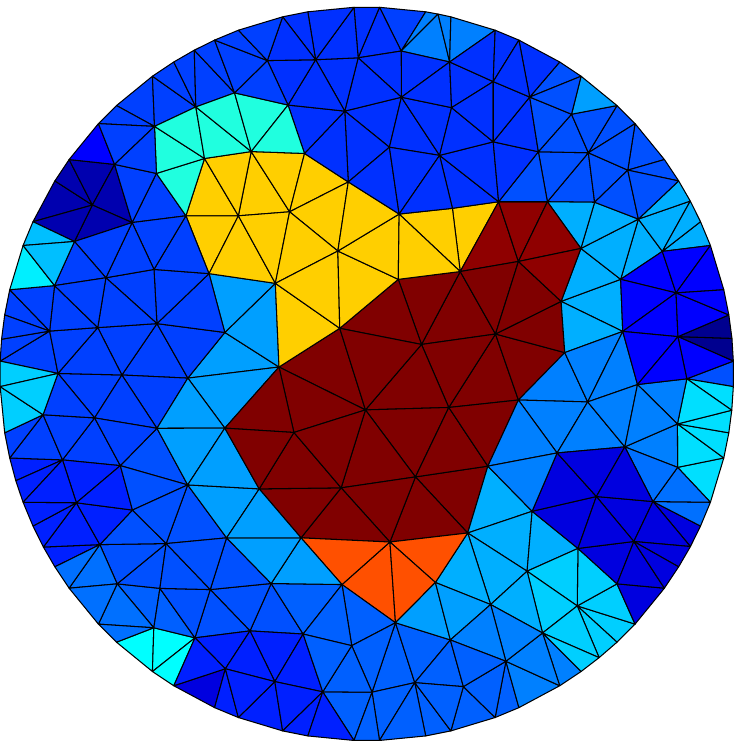}  &
  \includegraphics[width=\sizeA cm]{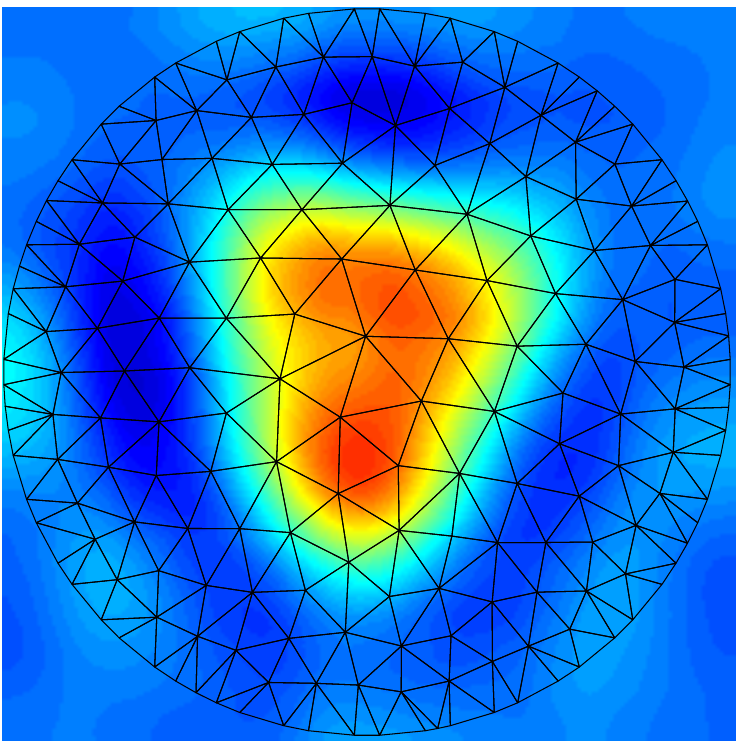}
  &
  \includegraphics[width=\sizeA cm]{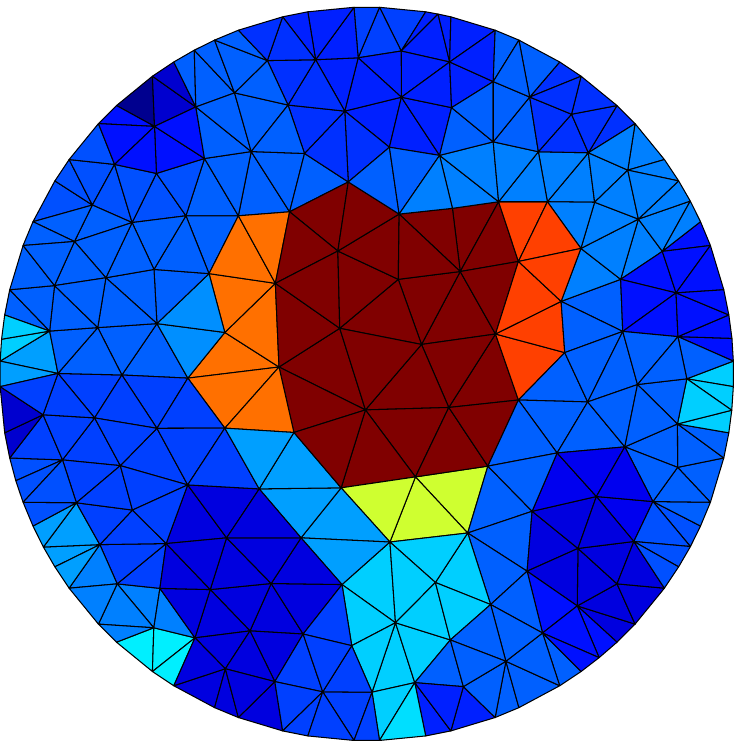}  &
  \includegraphics[width=\sizeA cm]{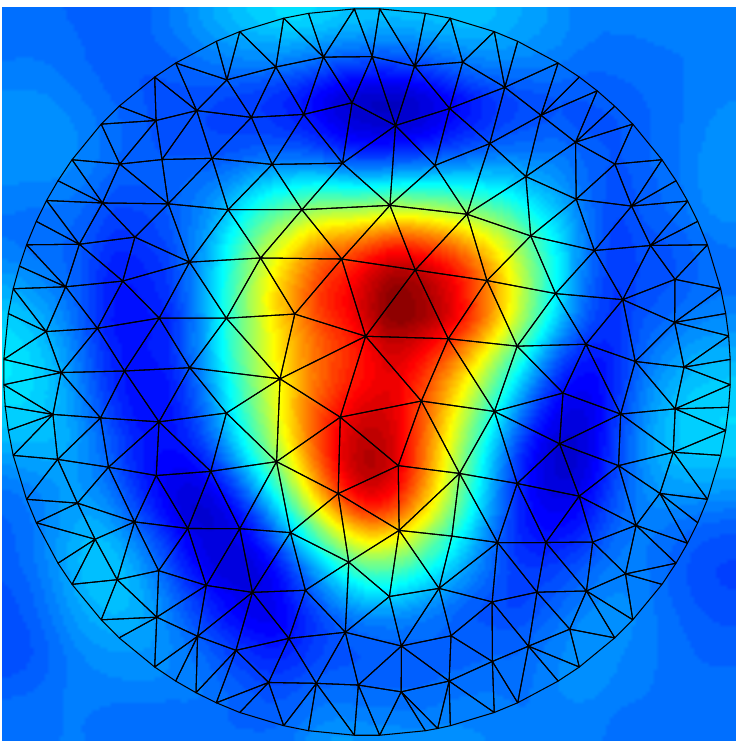} \\
  \includegraphics[width=\sizeA cm]{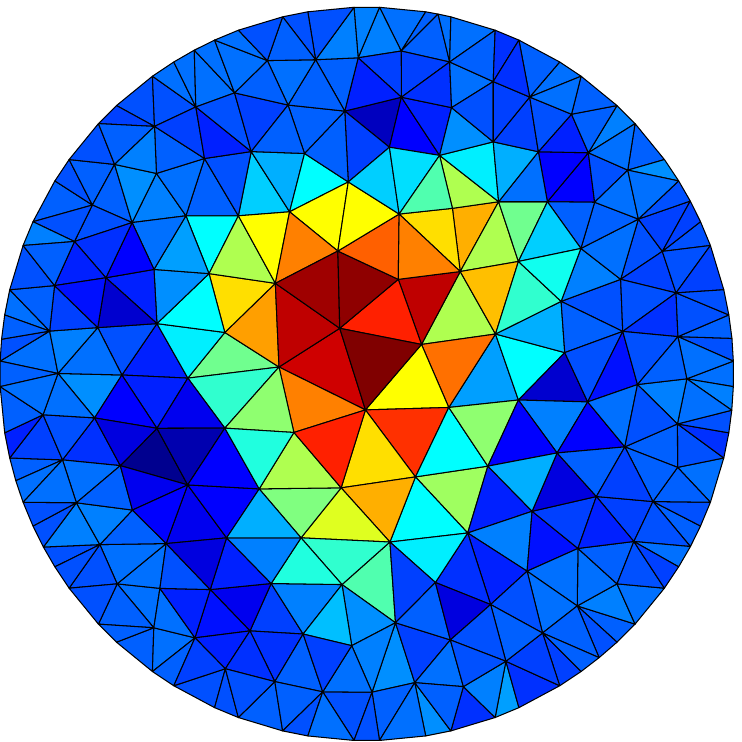}  &
  \includegraphics[width=\sizeA cm]{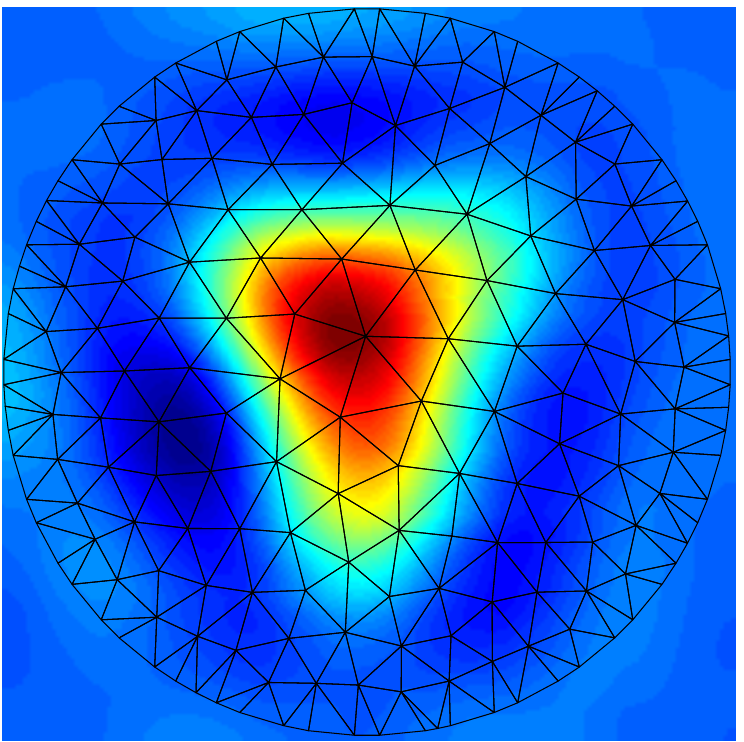}
  &
  \includegraphics[width=\sizeA cm]{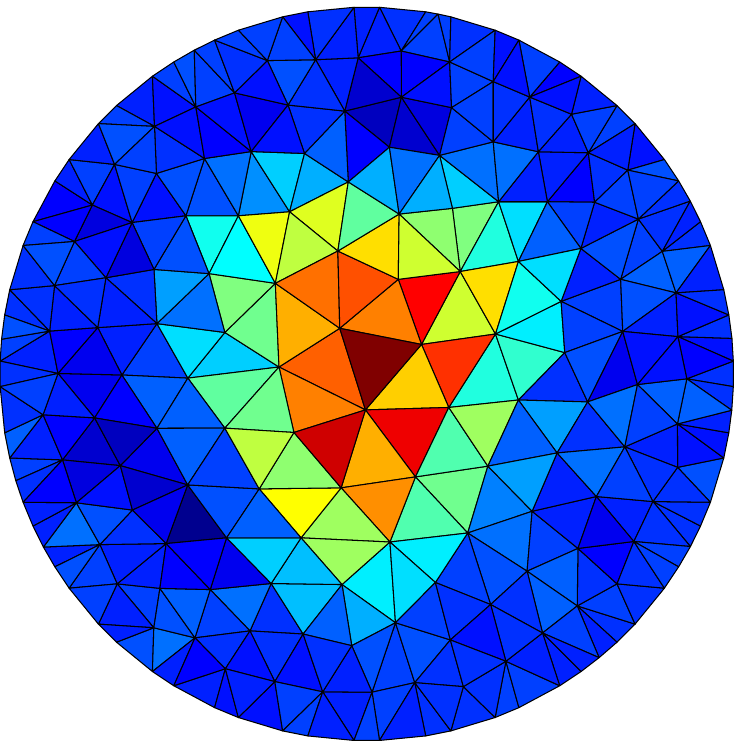}  &
  \includegraphics[width=\sizeA cm]{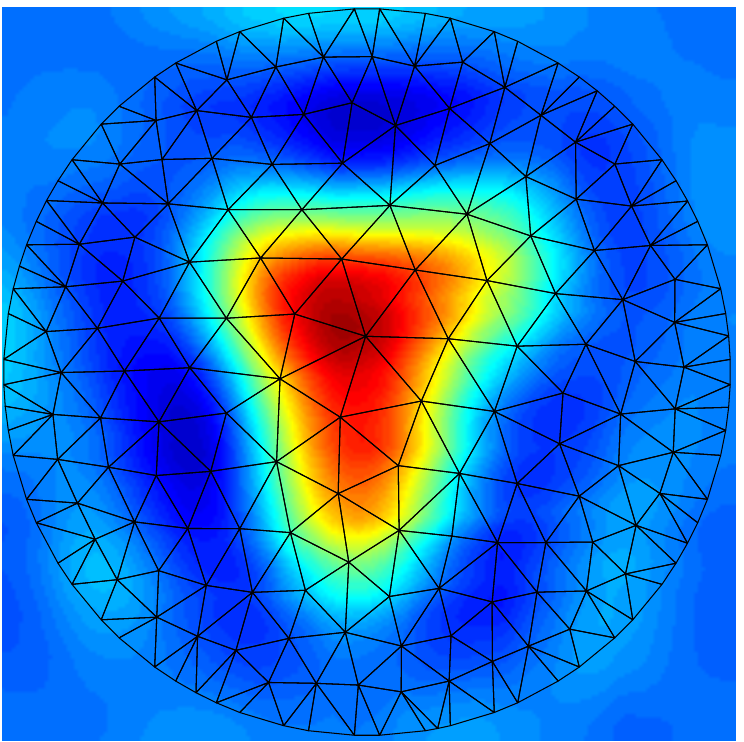}
  &
  \includegraphics[width=\sizeA cm]{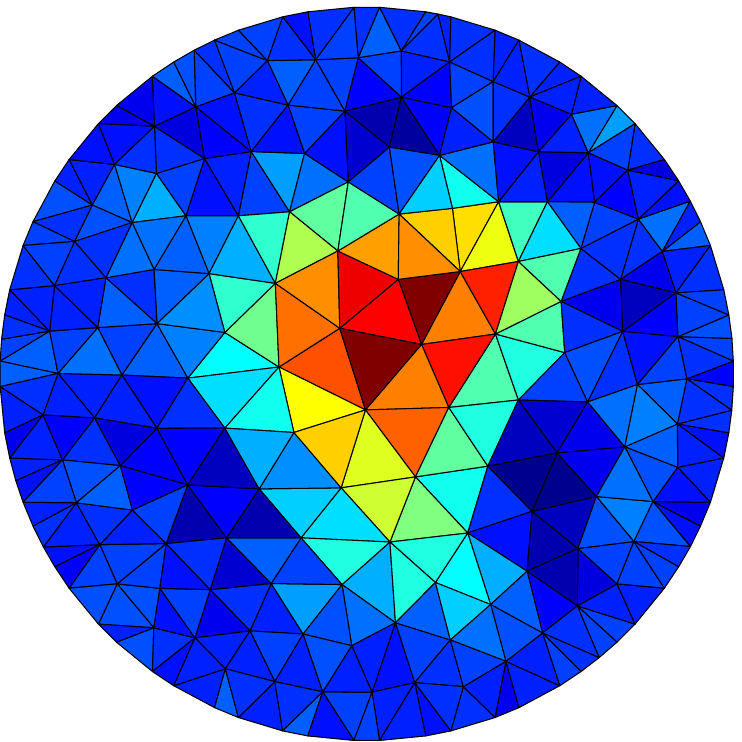}  &
  \includegraphics[width=\sizeA cm]{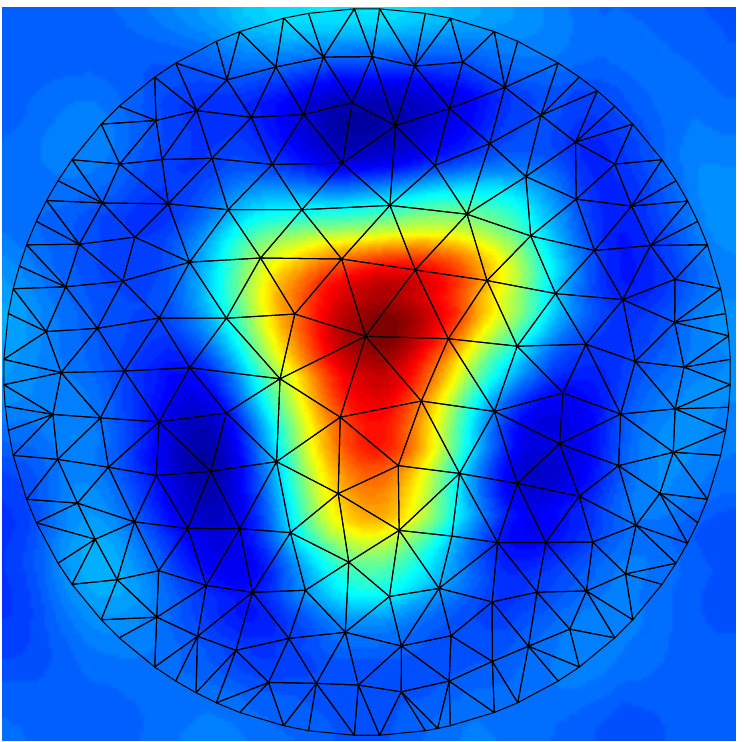} \\
  LR   &   SRR   &   LR   &   SRR   &   LR   &   SRR
\end{tabular}
\end{center}
\caption{Example 1-b: Translating T-shaped object with low measurement SNR and coarser FEM. Column groups indicate time instants (t=10, t=15, t=20). First row: Synthetic HR (desired) images (used in the EIT direct problem). Second, third and fourth row: LR EIT image and super resolved results (side by side) for the NOSER, TV and TS algorithms, respectively. Due to space limitations, only the super-resolved images considering motion estimated from the LR observations is displayed.}
\label{ex1b_results}
\end{figure*}

\renewcommand{\sizeA}{2.3}
\renewcommand{\sizeAhlf}{1.15}
\begin{figure*}[!htb] 
\begin{center}
\begin{tabular}{ccc}
  \includegraphics[width=\sizeA cm]{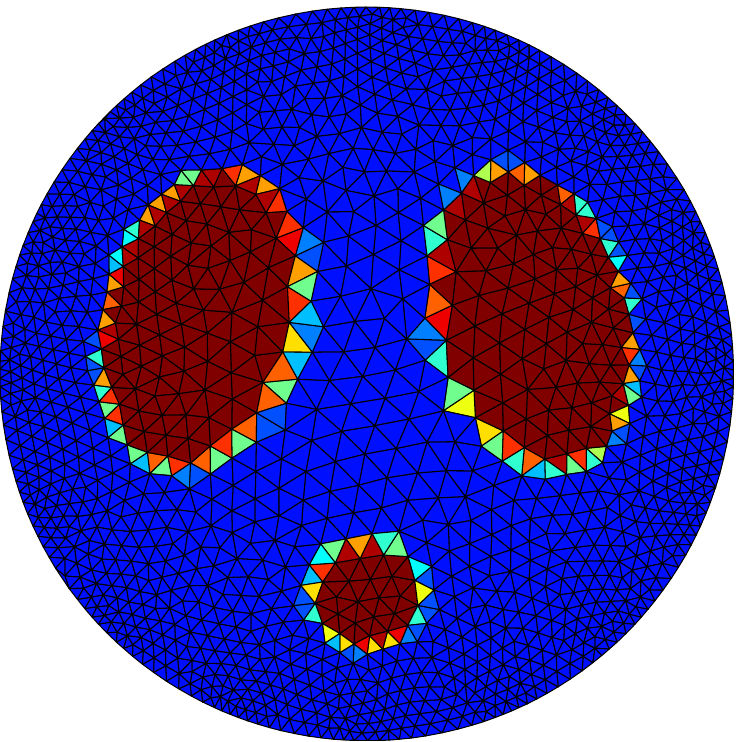}
  \hspace{\sizeAhlf cm}
  &
  \hspace{\sizeAhlf cm}
  \includegraphics[width=\sizeA cm]{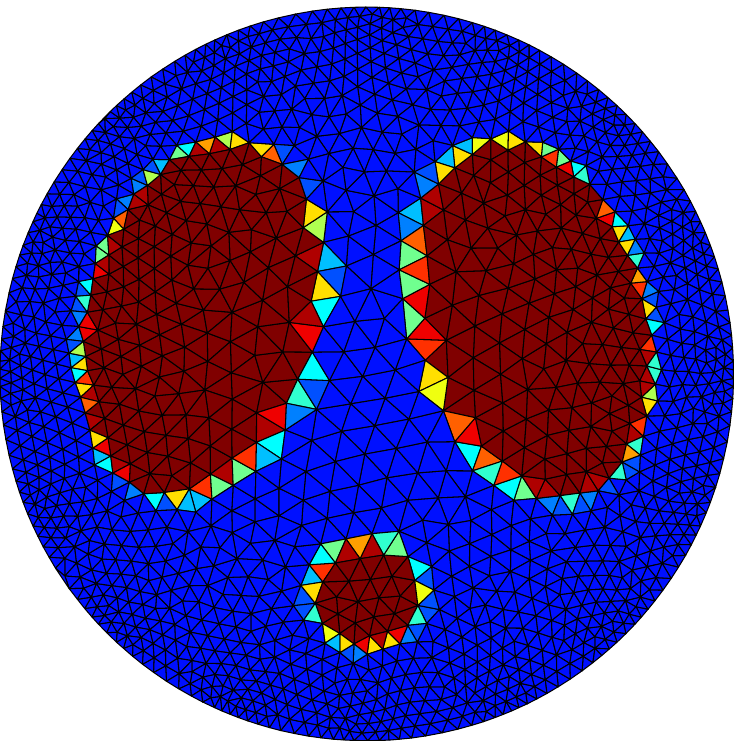}
  \hspace{\sizeAhlf cm}
  &
  \hspace{\sizeAhlf cm}
  \includegraphics[width=\sizeA cm]{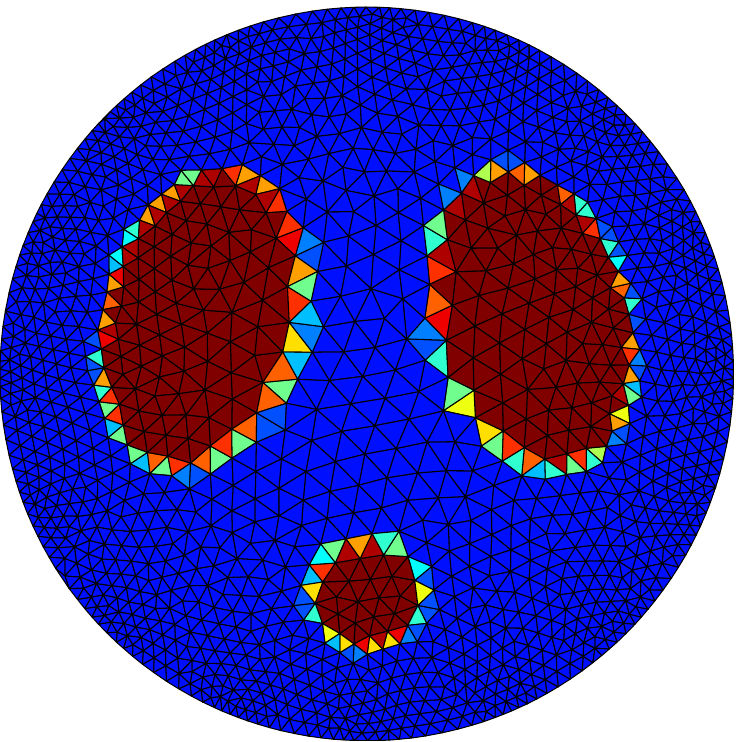}
  \\
  t=10 \hspace{\sizeAhlf cm} & \hspace{\sizeAhlf cm} t=15 \hspace{\sizeAhlf cm} & \hspace{\sizeAhlf cm} t=20
\end{tabular}
\begin{tabular}{cc||cc||cc}
  \includegraphics[width=\sizeA cm]{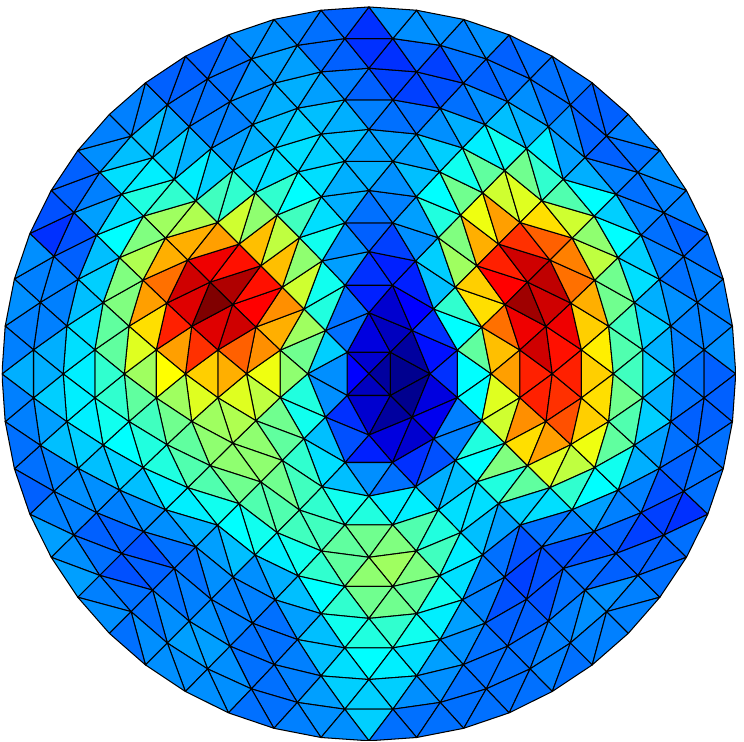}  &
  \includegraphics[width=\sizeA cm]{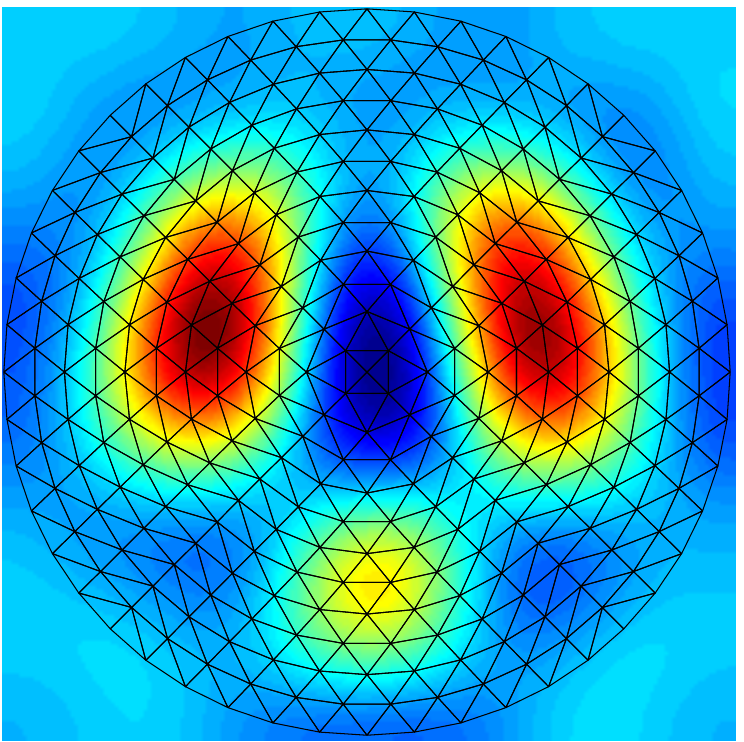}
  &
  \includegraphics[width=\sizeA cm]{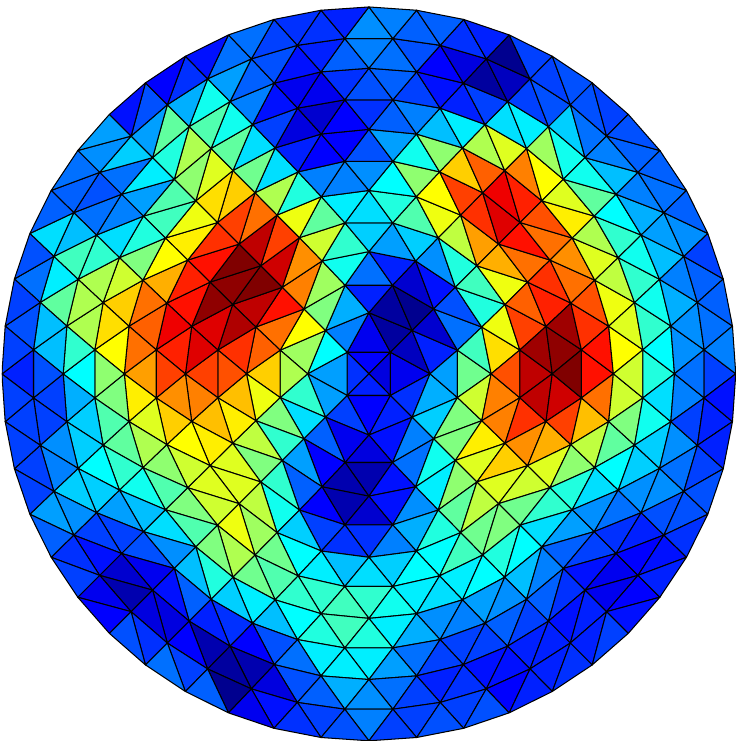}  &
  \includegraphics[width=\sizeA cm]{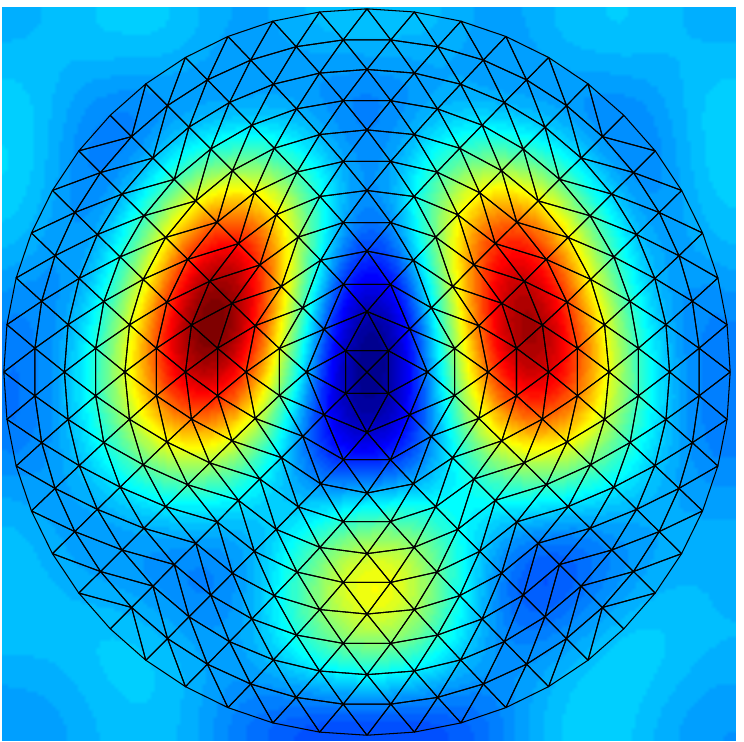}
  &
  \includegraphics[width=\sizeA cm]{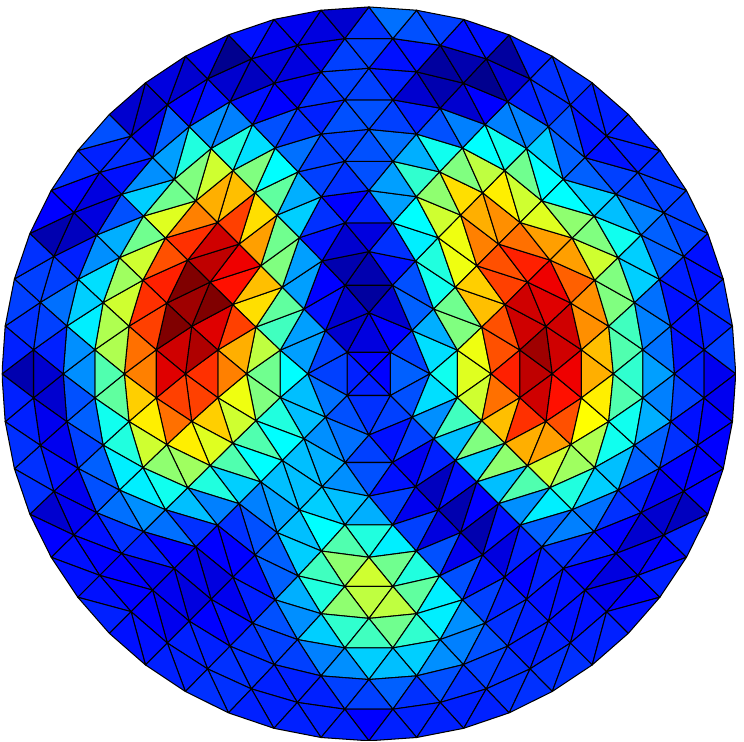}  &
  \includegraphics[width=\sizeA cm]{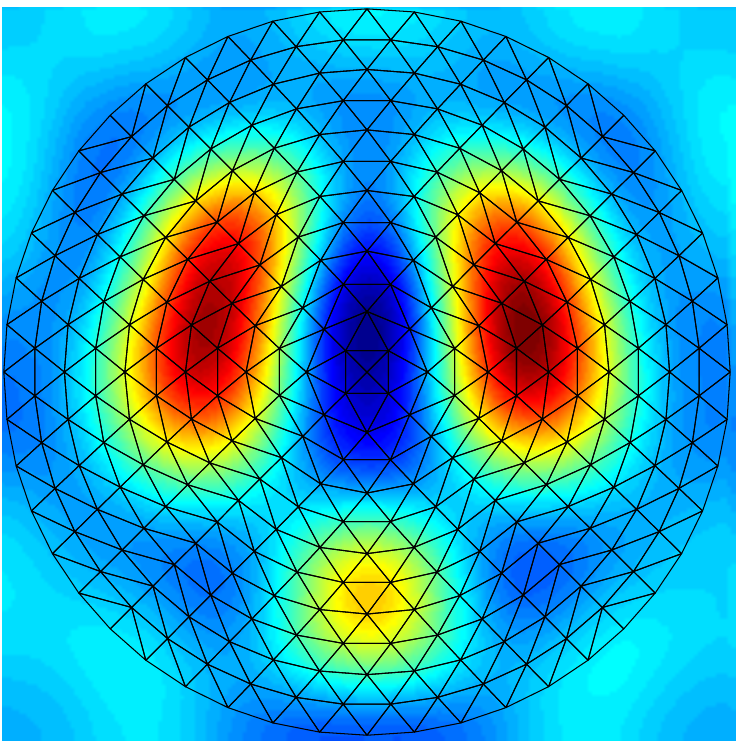} \\
  \includegraphics[width=\sizeA cm]{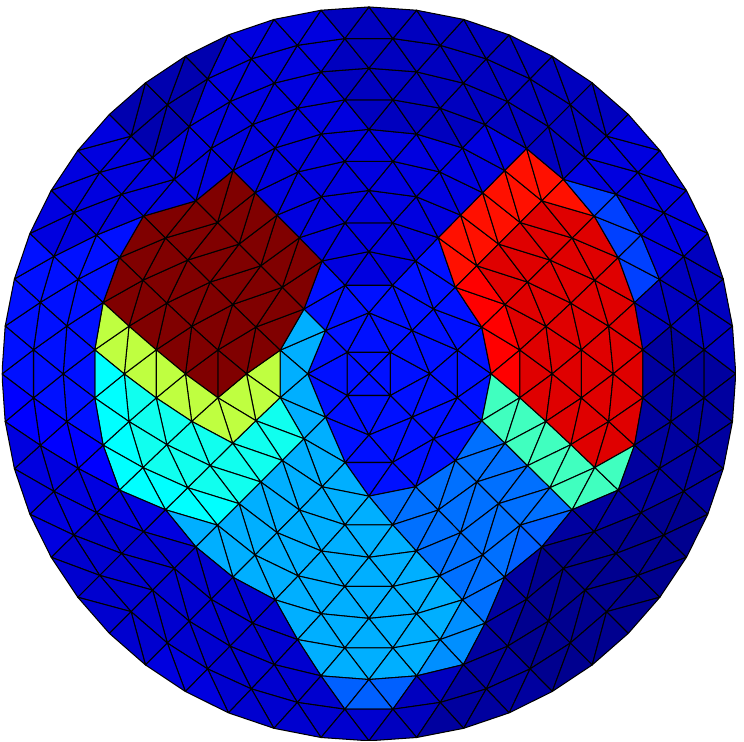}  &
  \includegraphics[width=\sizeA cm]{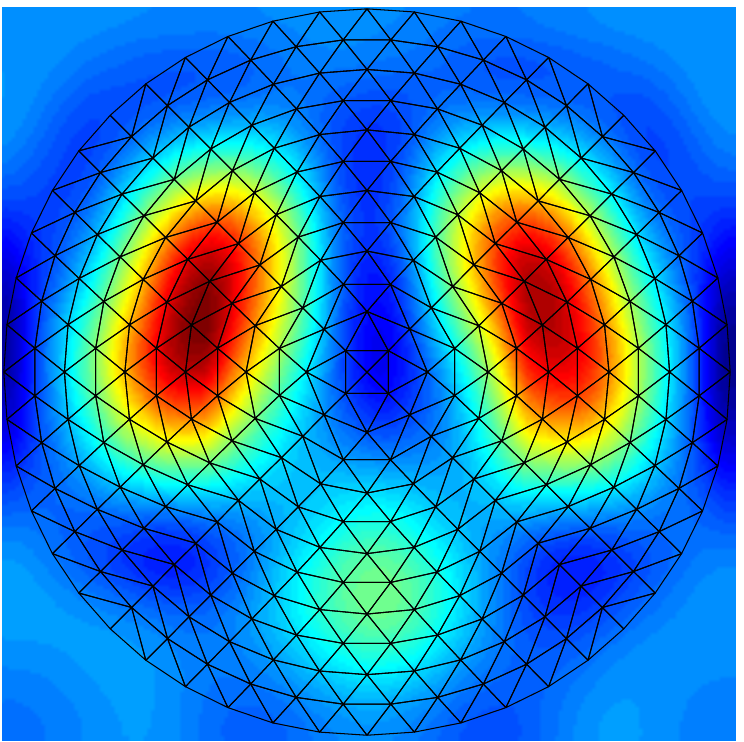}
  &
  \includegraphics[width=\sizeA cm]{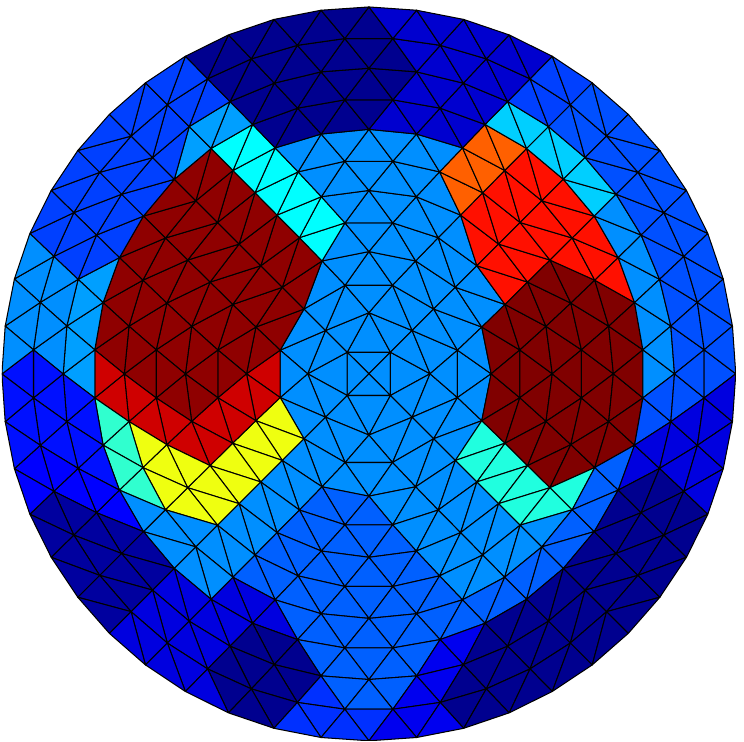}  &
  \includegraphics[width=\sizeA cm]{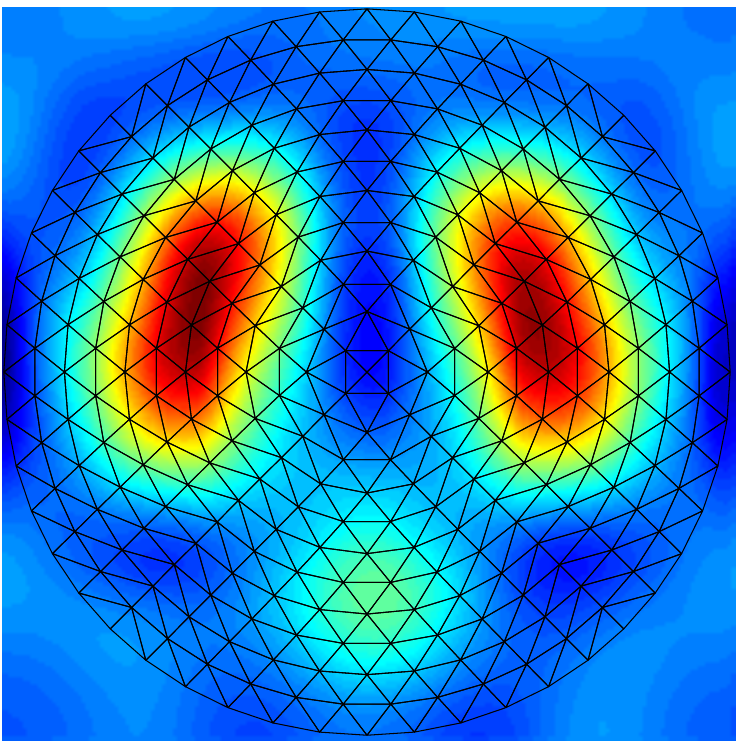}
  &
  \includegraphics[width=\sizeA cm]{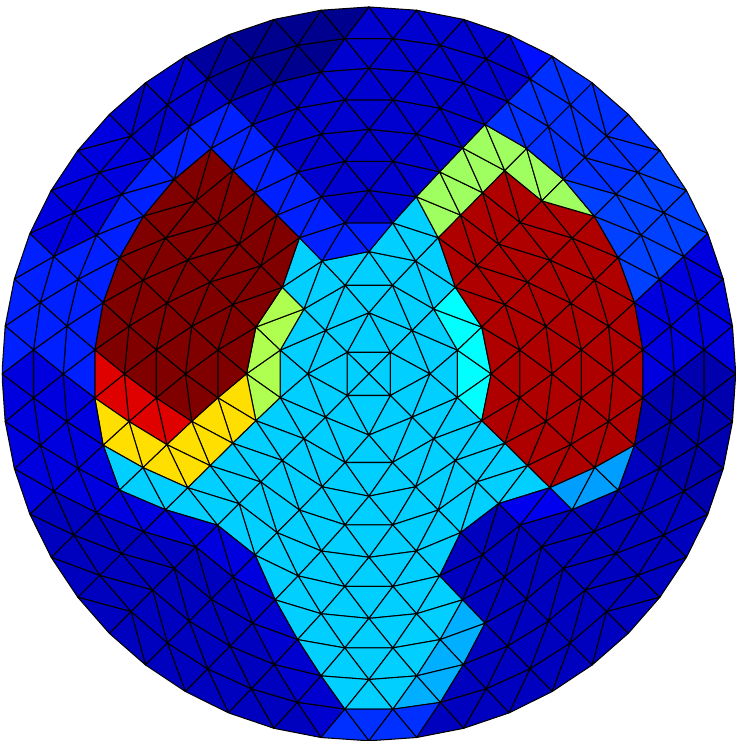}  &
  \includegraphics[width=\sizeA cm]{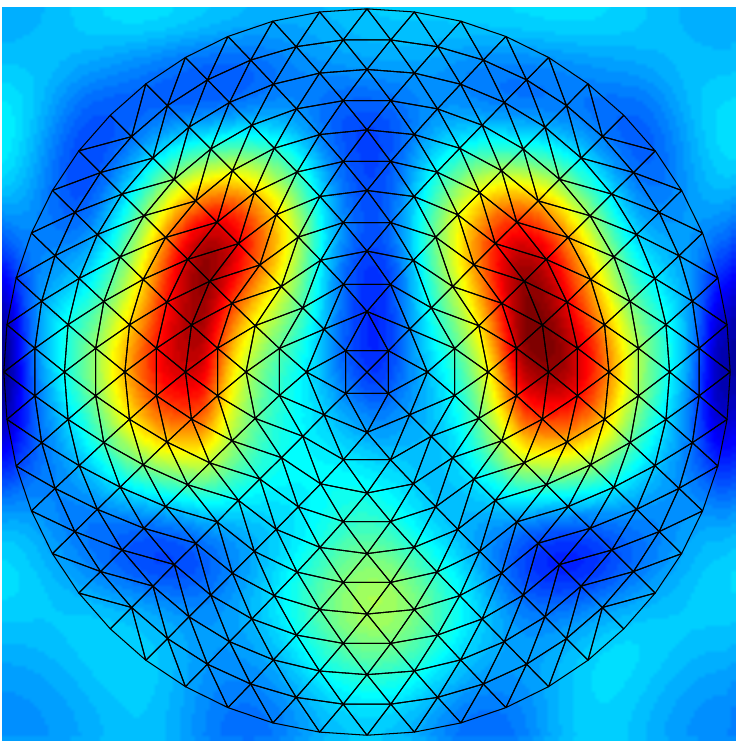} \\
  \includegraphics[width=\sizeA cm]{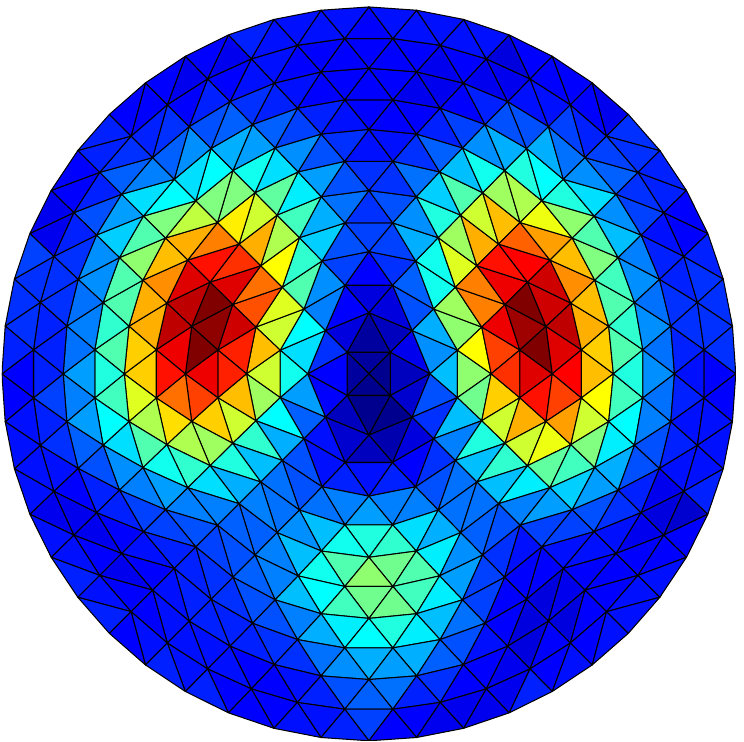}  &
  \includegraphics[width=\sizeA cm]{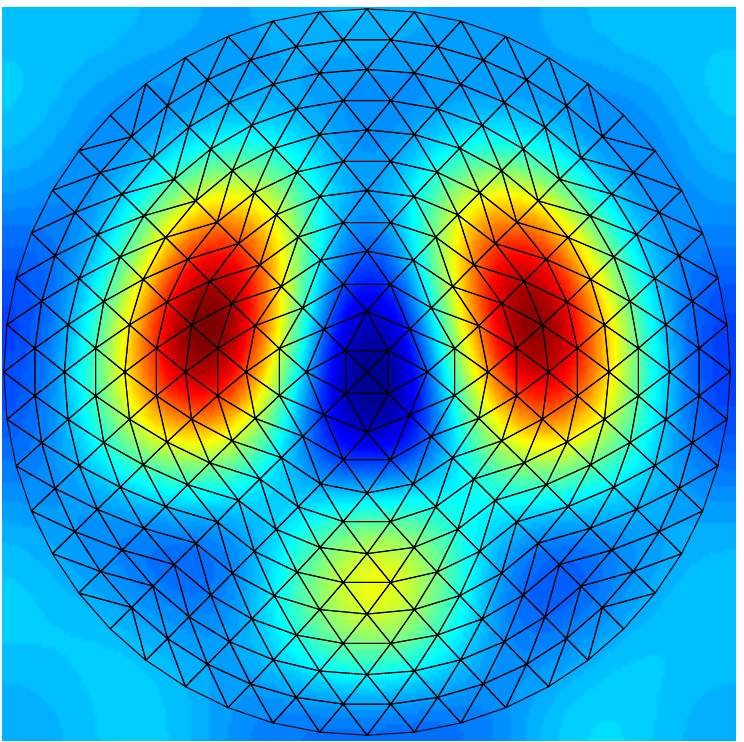}
  &
  \includegraphics[width=\sizeA cm]{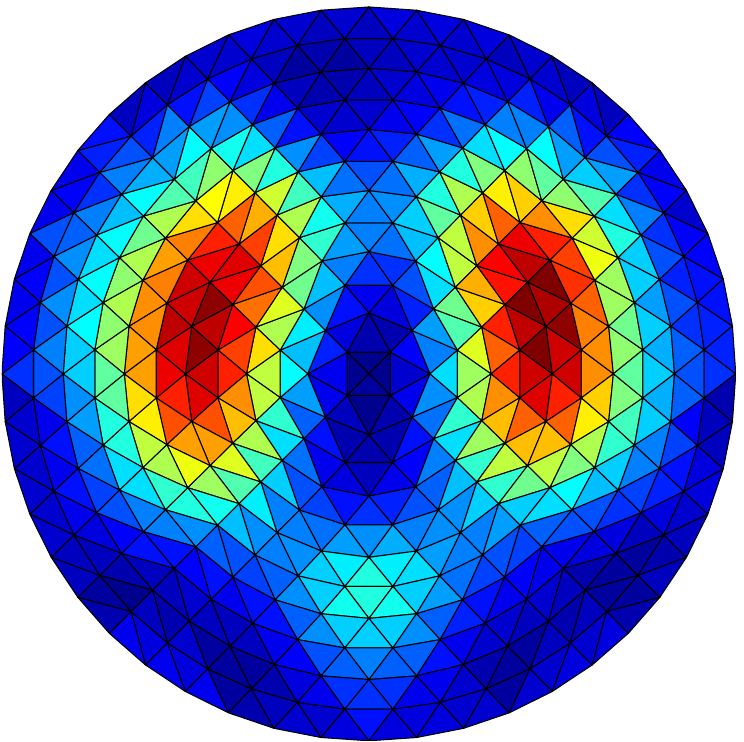}  &
  \includegraphics[width=\sizeA cm]{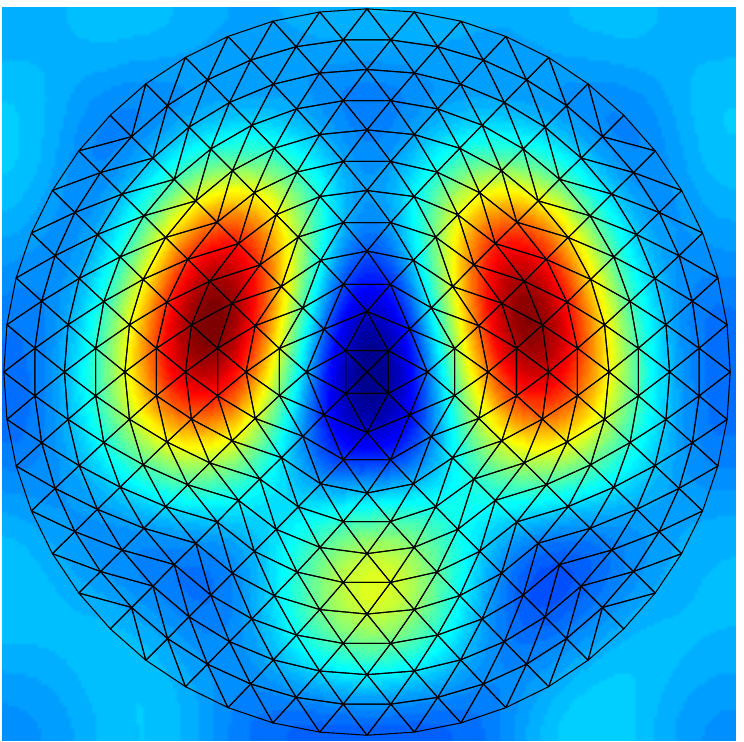}
  &
  \includegraphics[width=\sizeA cm]{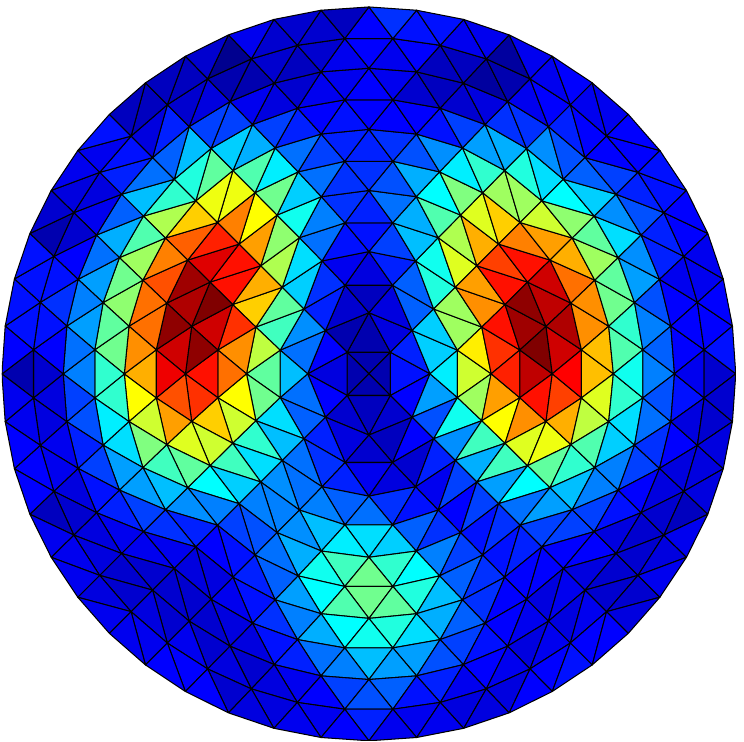}  &
  \includegraphics[width=\sizeA cm]{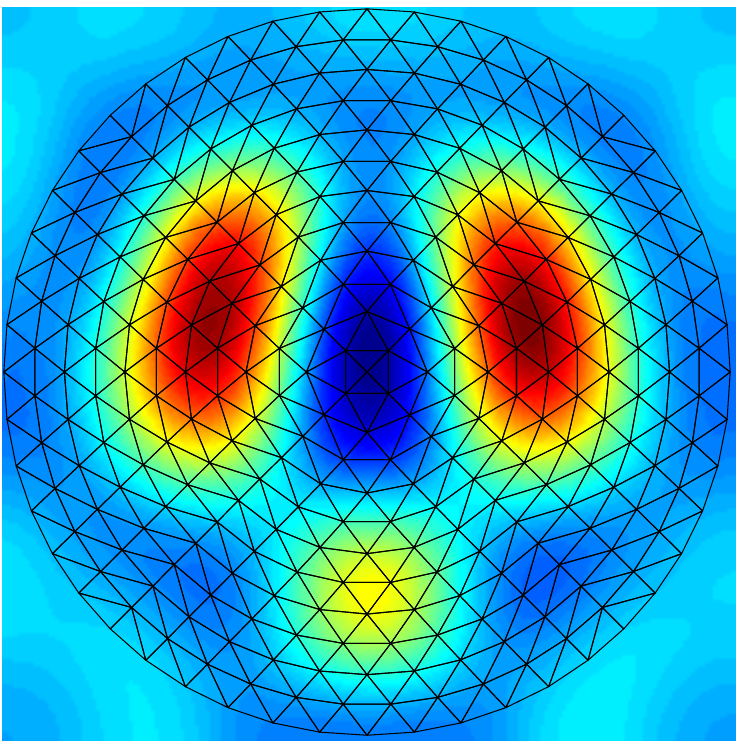} \\
  LR   &   SRR   &   LR   &   SRR   &   LR   &   SRR
\end{tabular}
\end{center}
\caption{Example 2-a: Synthetic lung object with high measurement SNR and finer FEM. Column groups indicate time instants (t=10, t=15, t=20). First row: Synthetic HR (desired) images (used in the EIT direct problem). Second, third and fourth row: LR EIT image and super resolved results (side by side) for the NOSER, TV and TS algorithms, respectively. Due to space limitations, only the super-resolved images considering motion estimated from the LR observations is displayed.}
\label{ex2a_results}
\end{figure*}

\renewcommand{\sizeA}{2.3}
\renewcommand{\sizeAhlf}{1.15}
\begin{figure*}[!htb] 
\begin{center}
\begin{tabular}{ccc}
  \includegraphics[width=\sizeA cm]{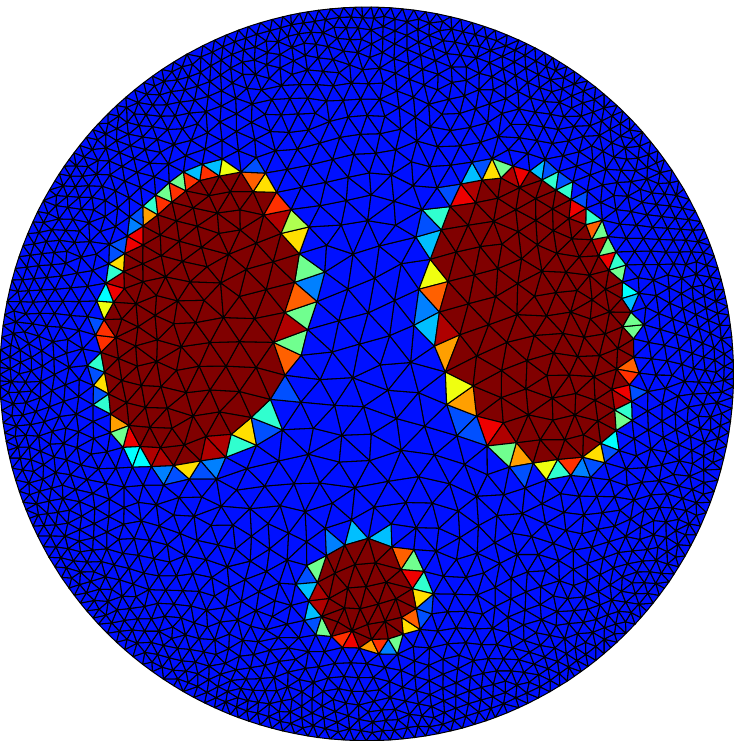}
  \hspace{\sizeAhlf cm}
  &
  \hspace{\sizeAhlf cm}
  \includegraphics[width=\sizeA cm]{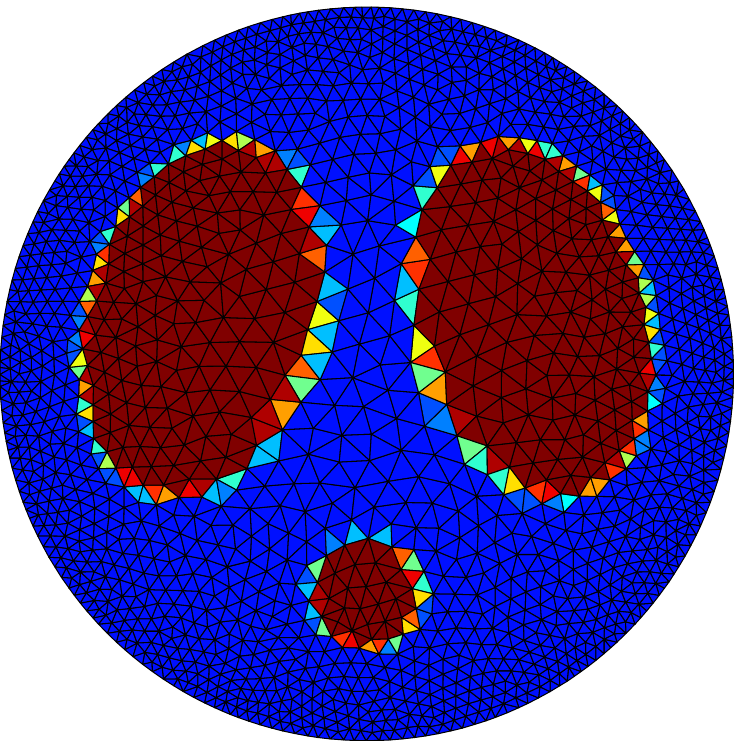}
  \hspace{\sizeAhlf cm}
  &
  \hspace{\sizeAhlf cm}
  \includegraphics[width=\sizeA cm]{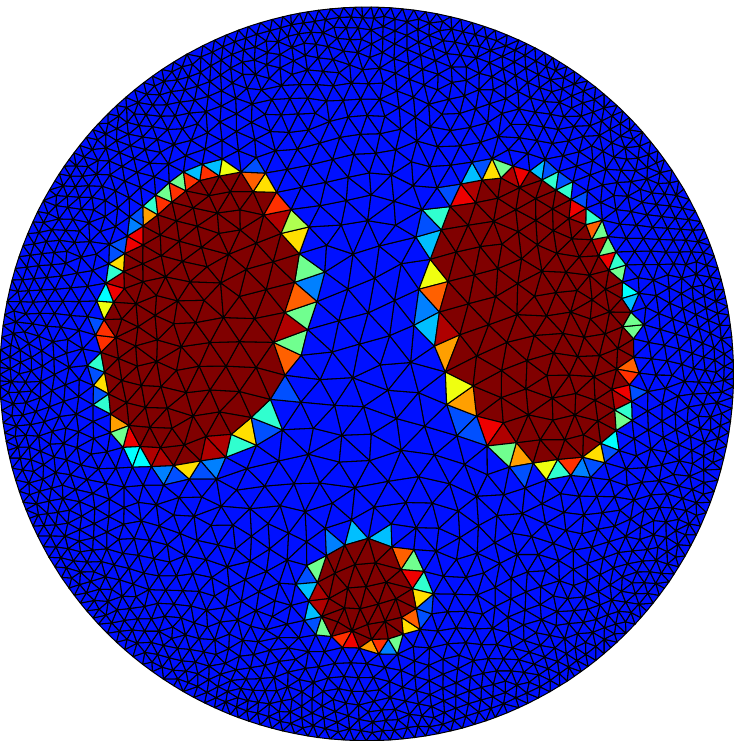}
  \\
  t=10 \hspace{\sizeAhlf cm} & \hspace{\sizeAhlf cm} t=15 \hspace{\sizeAhlf cm} & \hspace{\sizeAhlf cm} t=20
\end{tabular}
\begin{tabular}{cc||cc||cc}
  \includegraphics[width=\sizeA cm]{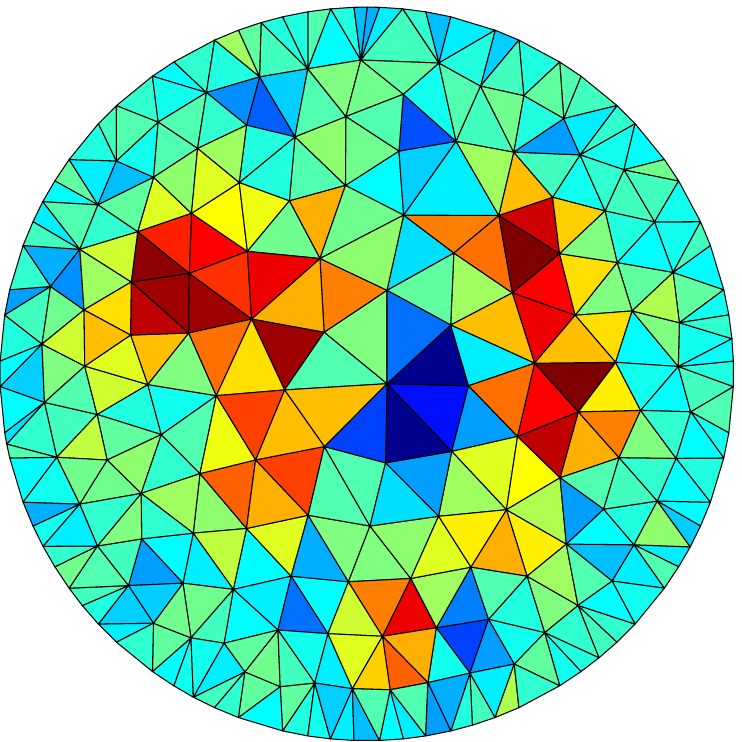}  &
  \includegraphics[width=\sizeA cm]{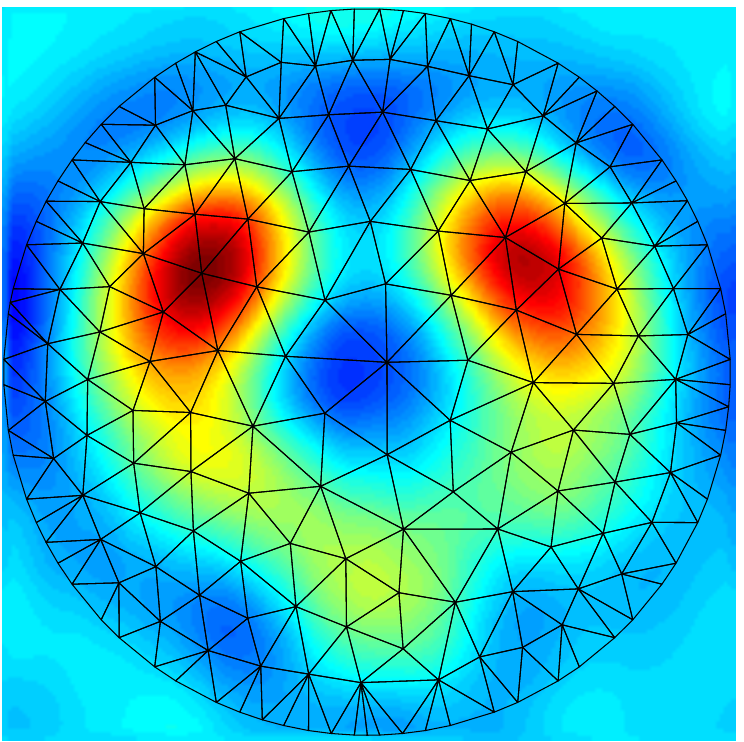}
  &
  \includegraphics[width=\sizeA cm]{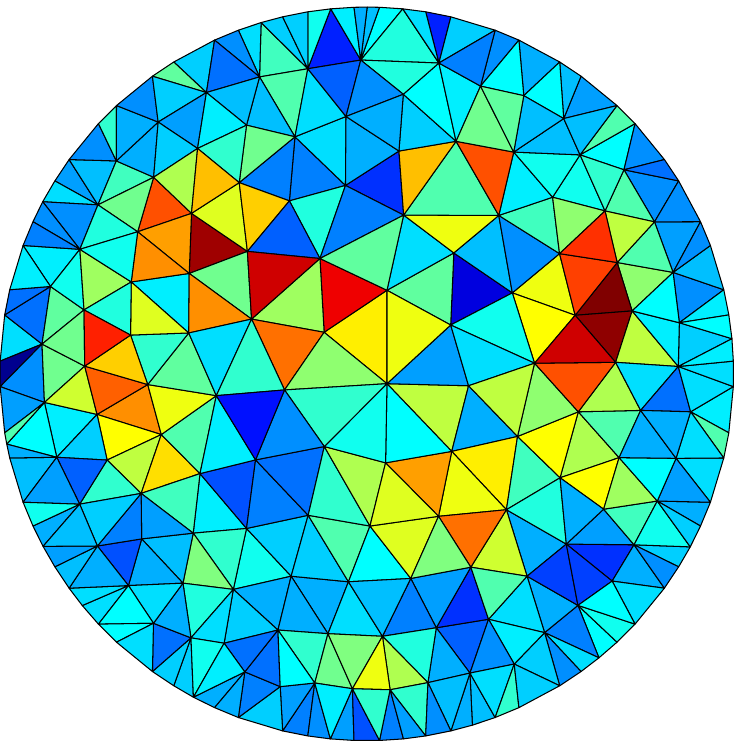}  &
  \includegraphics[width=\sizeA cm]{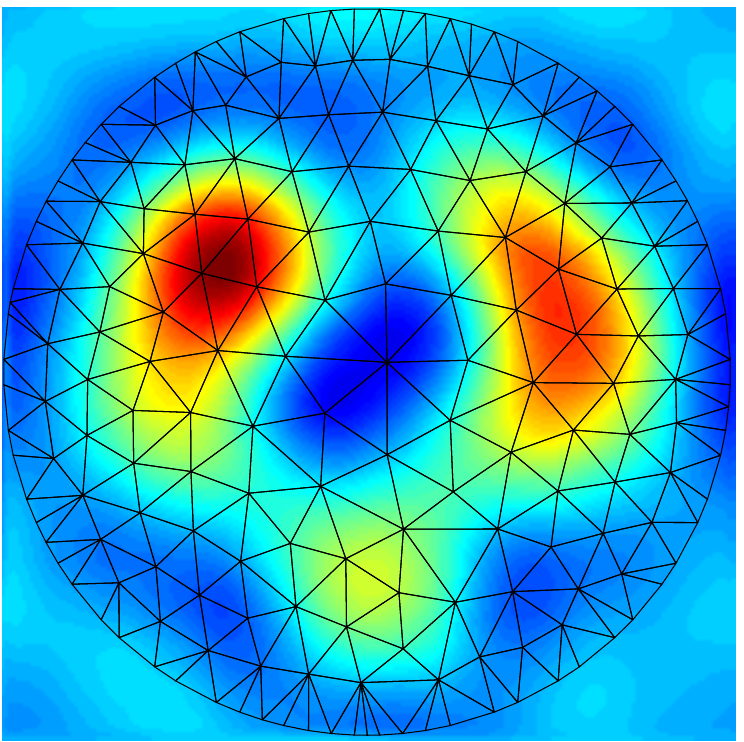}
  &
  \includegraphics[width=\sizeA cm]{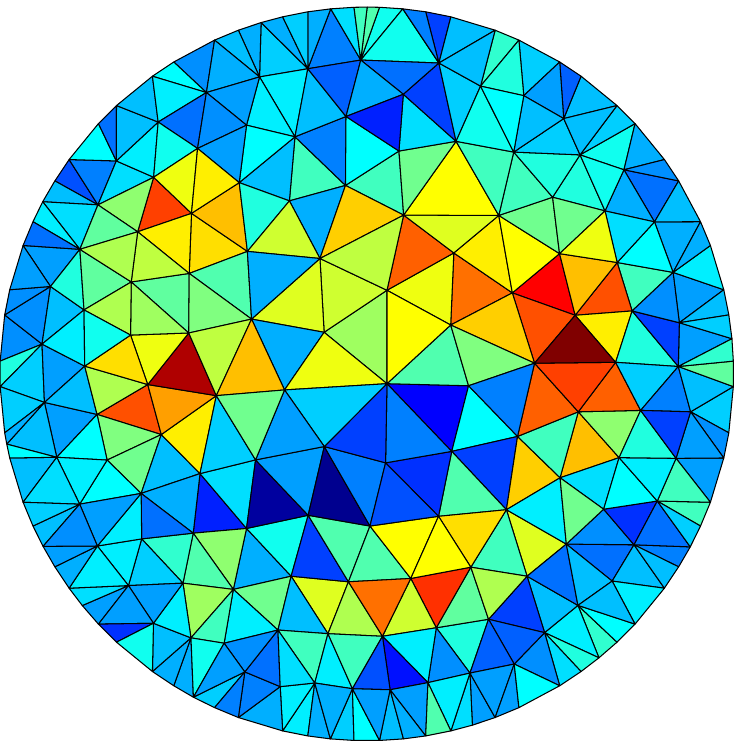}  &
  \includegraphics[width=\sizeA cm]{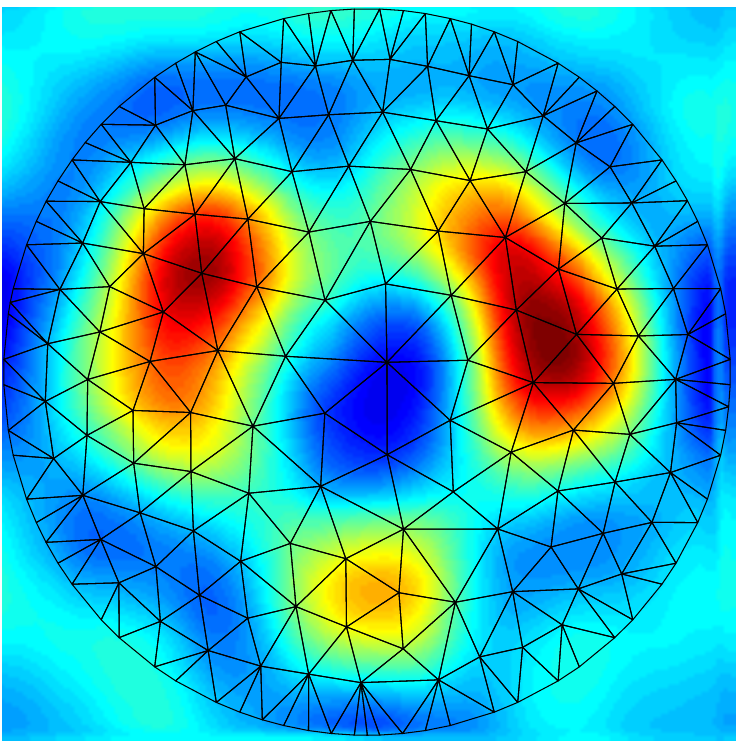} \\
  \includegraphics[width=\sizeA cm]{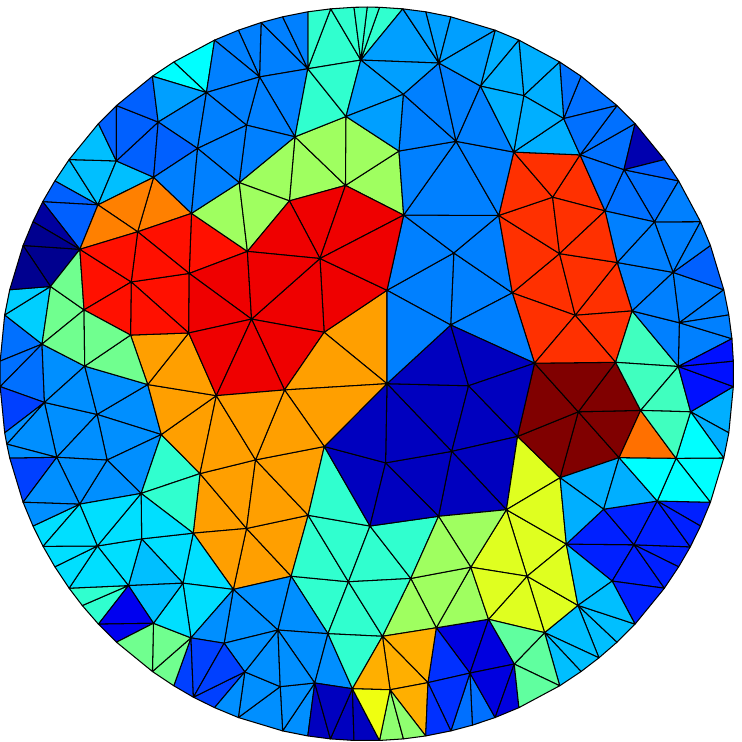}  &
  \includegraphics[width=\sizeA cm]{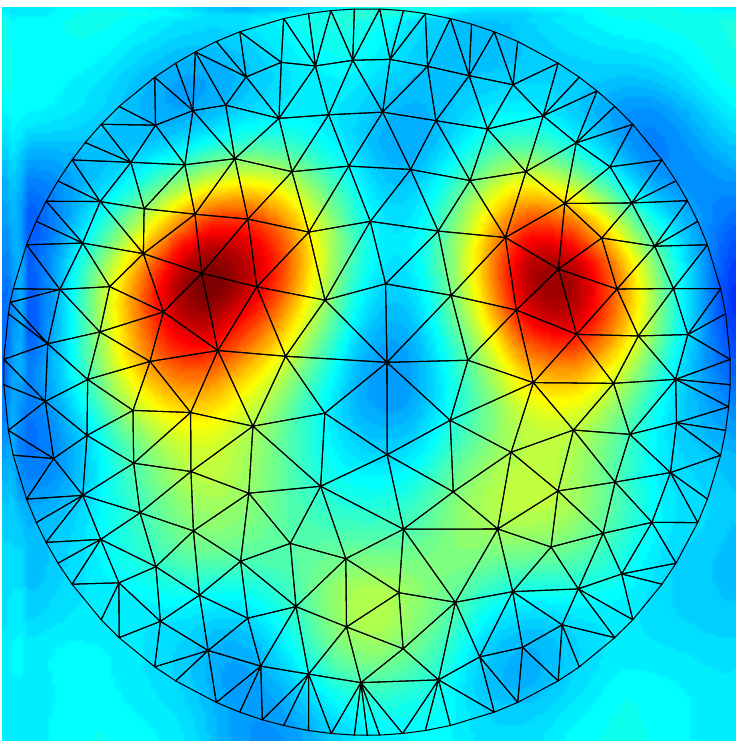}
  &
  \includegraphics[width=\sizeA cm]{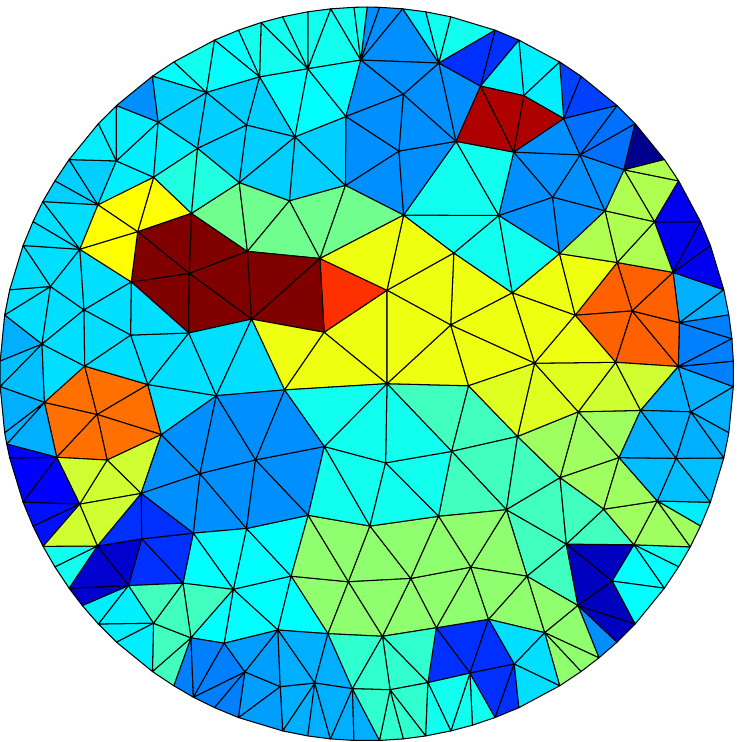}  &
  \includegraphics[width=\sizeA cm]{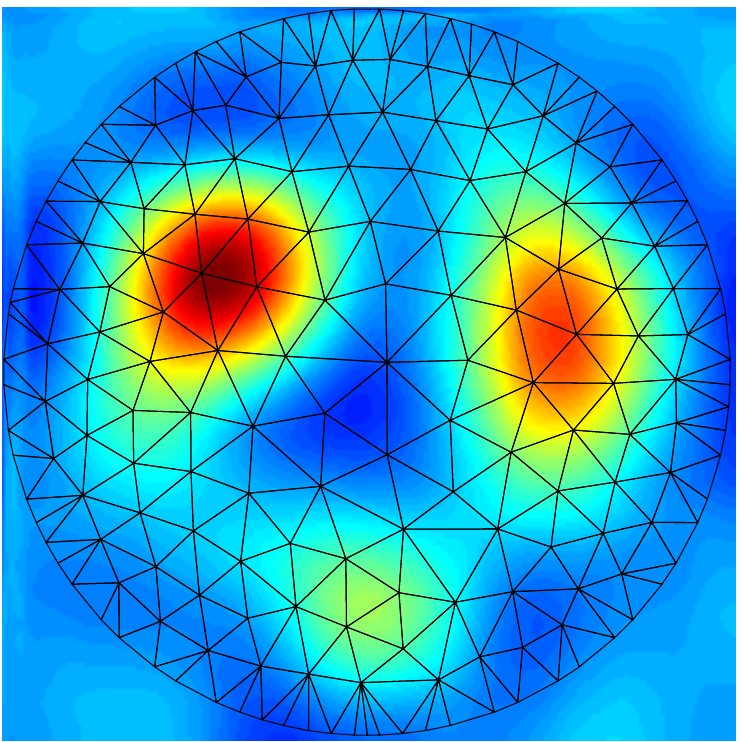}
  &
  \includegraphics[width=\sizeA cm]{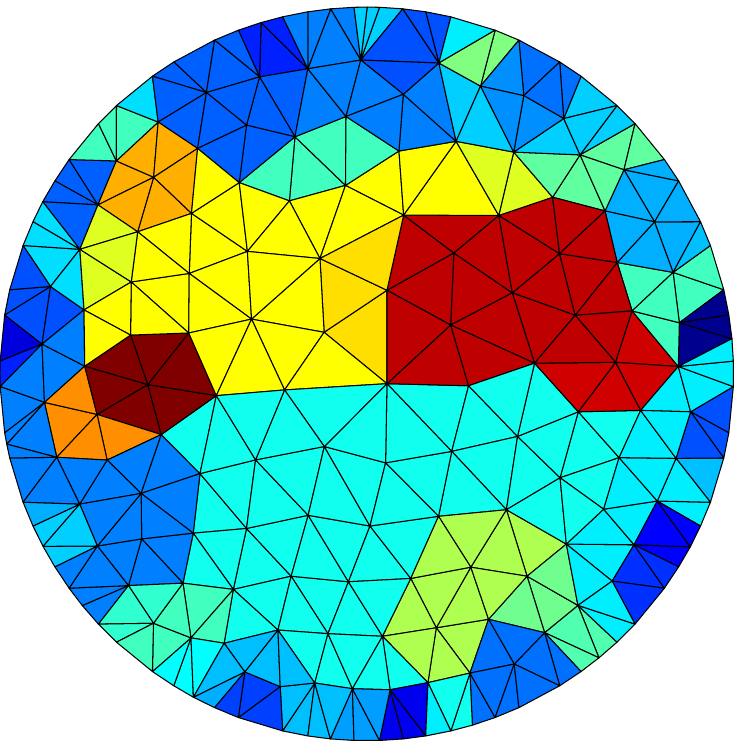}  &
  \includegraphics[width=\sizeA cm]{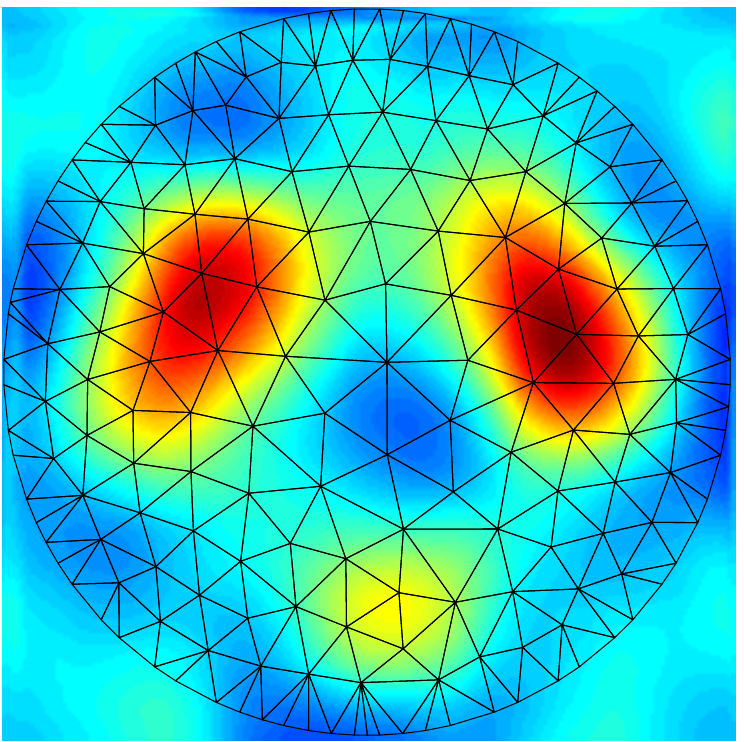} \\
  \includegraphics[width=\sizeA cm]{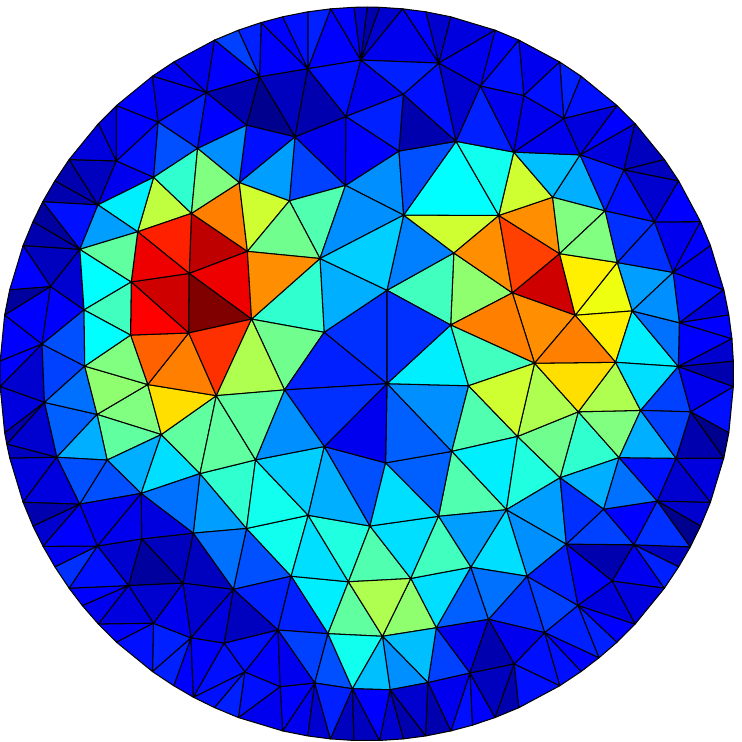}  &
  \includegraphics[width=\sizeA cm]{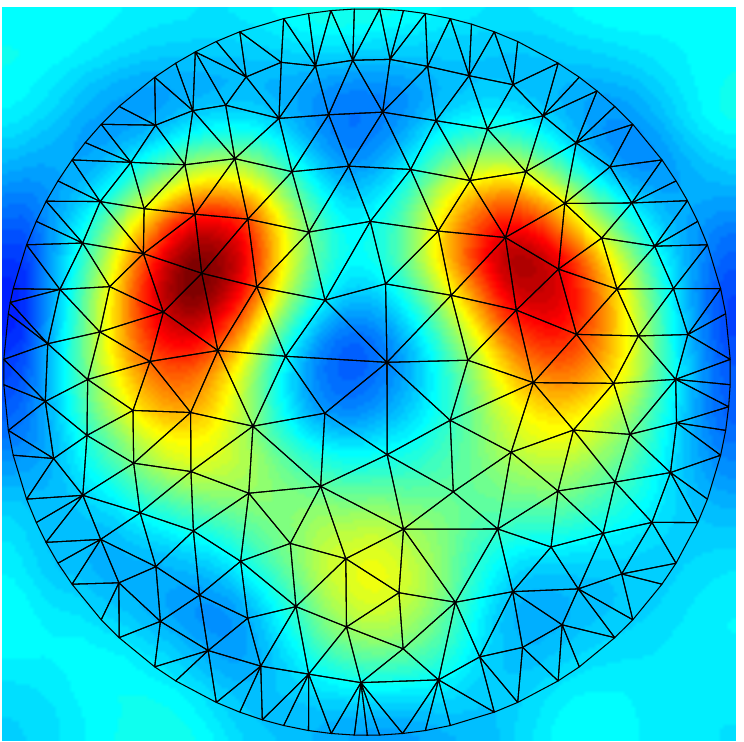}
  &
  \includegraphics[width=\sizeA cm]{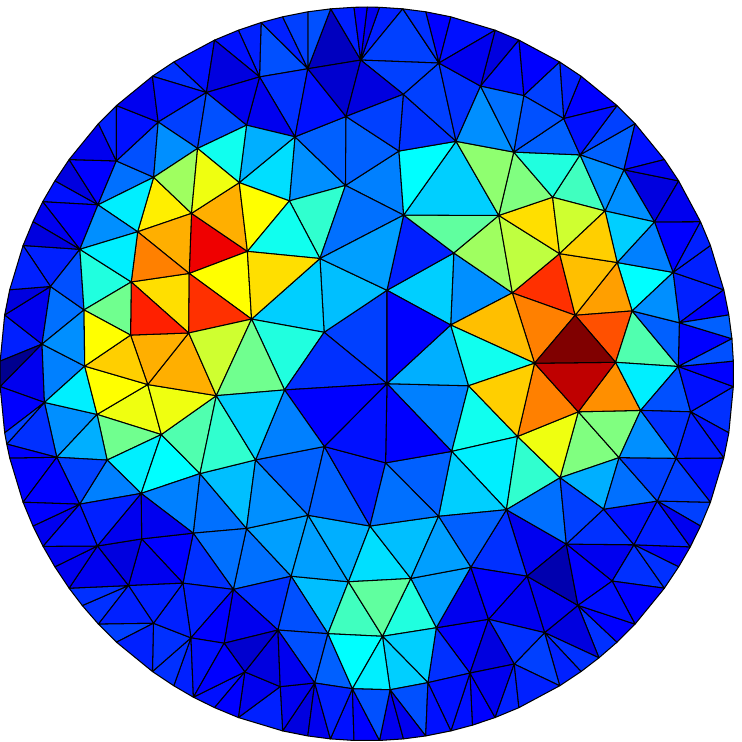}  &
  \includegraphics[width=\sizeA cm]{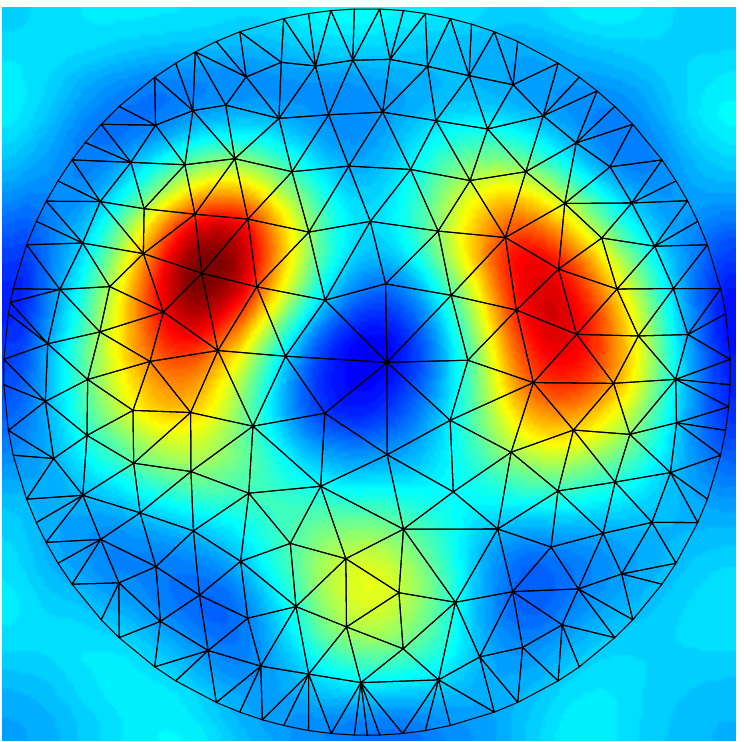}
  &
  \includegraphics[width=\sizeA cm]{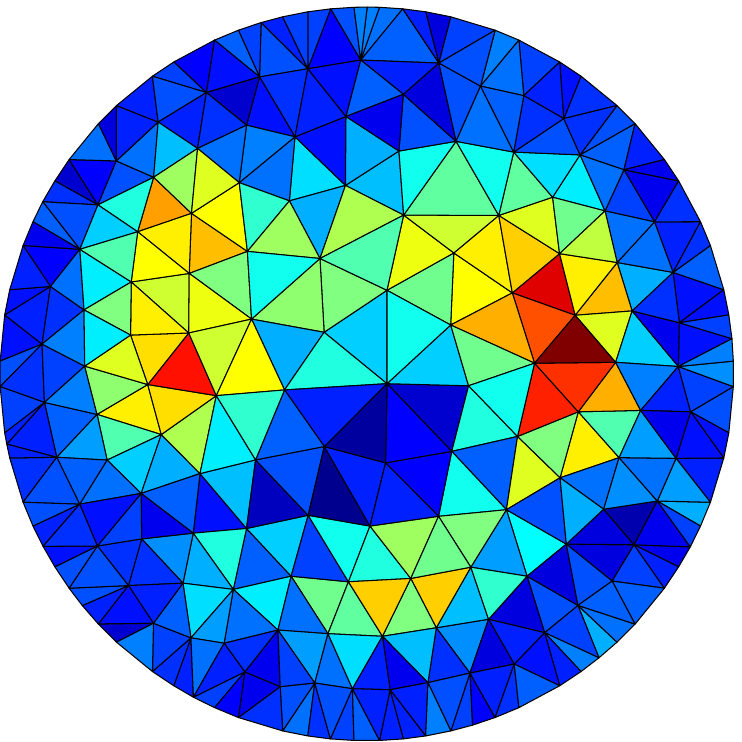}  &
  \includegraphics[width=\sizeA cm]{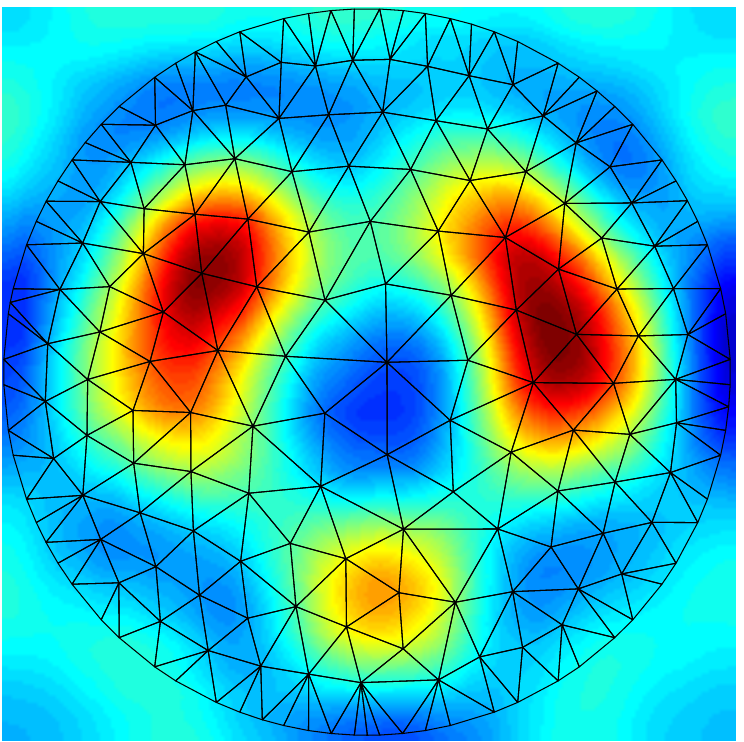} \\
  LR   &   SRR   &   LR   &   SRR   &   LR   &   SRR
\end{tabular}
\end{center}
\caption{Example 2-b: Synthetic lung object with low measurement SNR and coarser FEM. Column groups indicate time instants (t=10, t=15, t=20). First row: Synthetic HR (desired) images (used in the EIT direct problem). Second, third and fourth row: LR EIT image and super resolved results (side by side) for the NOSER, TV and TS algorithms, respectively. Due to space limitations, only the super-resolved images considering motion estimated from the LR observations is displayed.}
\label{ex2b_results}
\end{figure*}

\renewcommand{\sizeA}{3.5}
\begin{figure}[htb] 
\begin{center}
\begin{tabular}{cc}
  \includegraphics[width=\sizeA cm]{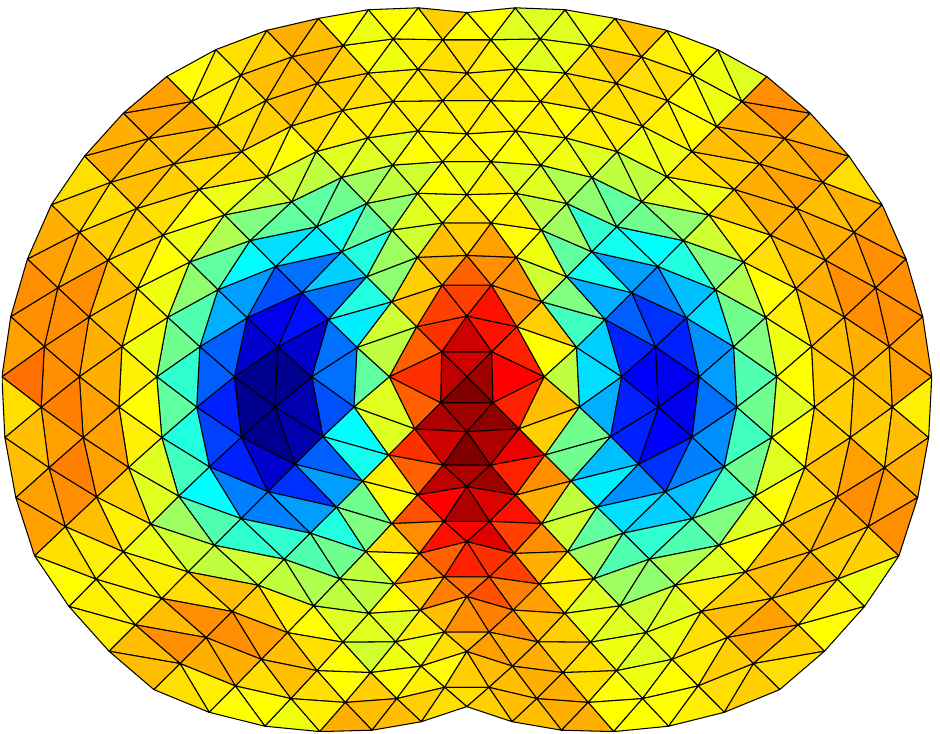}  &
  \includegraphics[width=\sizeA cm]{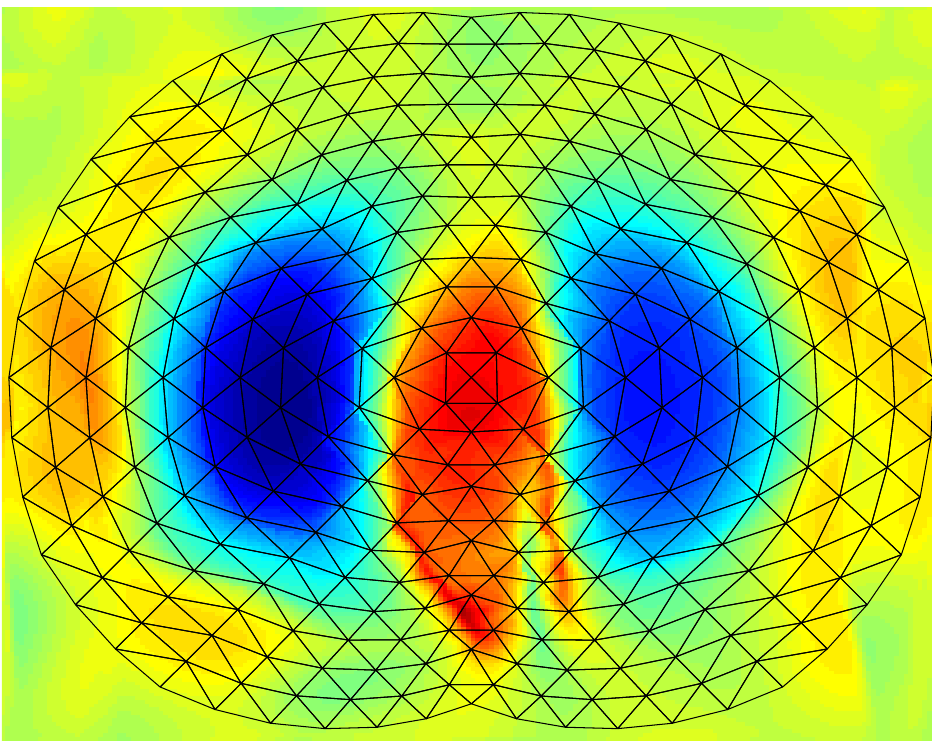}\\
  \includegraphics[width=\sizeA cm]{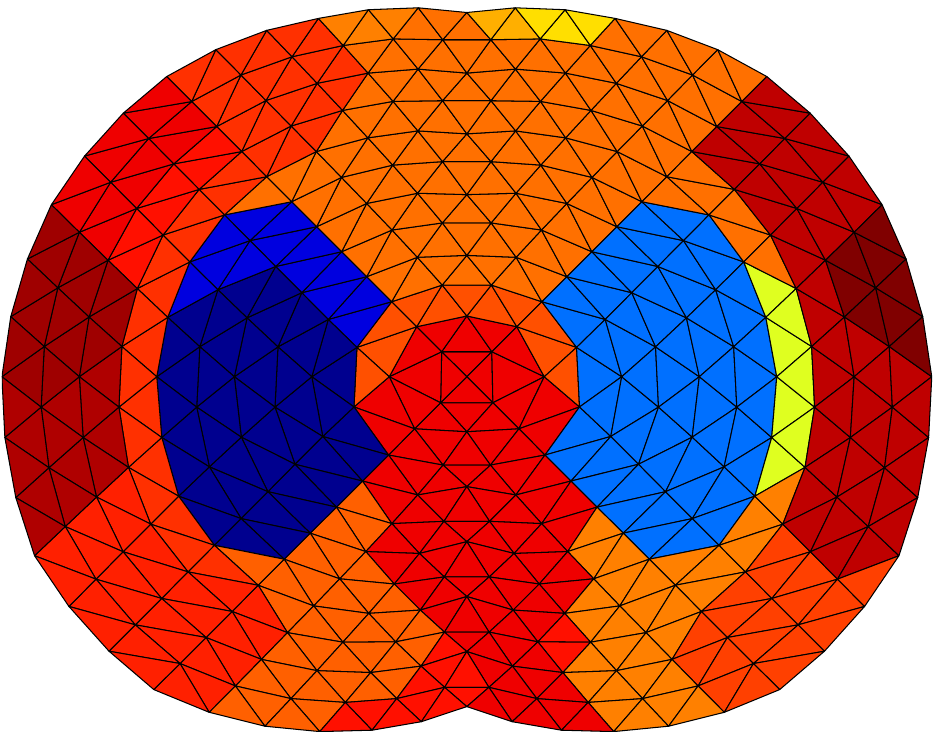}  &
  \includegraphics[width=\sizeA cm]{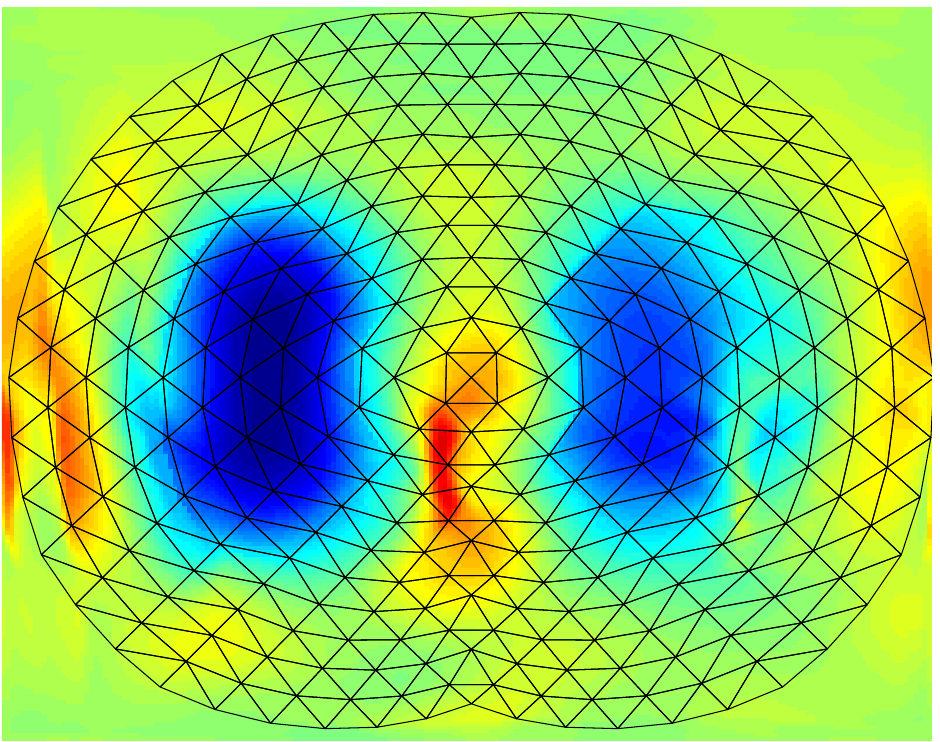}\\
  \includegraphics[width=\sizeA cm]{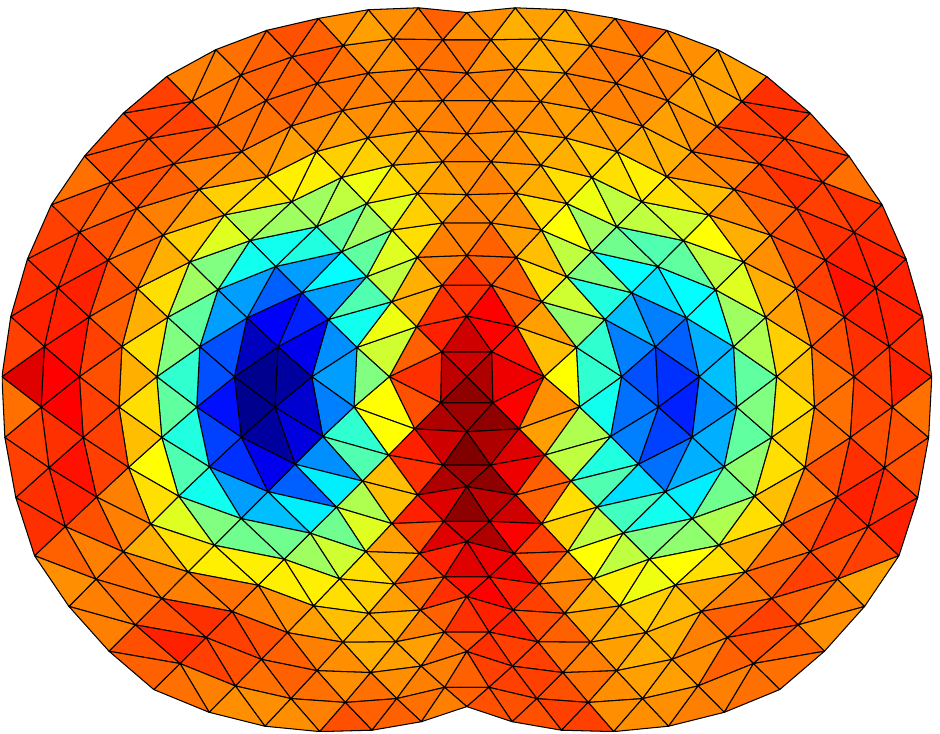}  &
  \includegraphics[width=\sizeA cm]{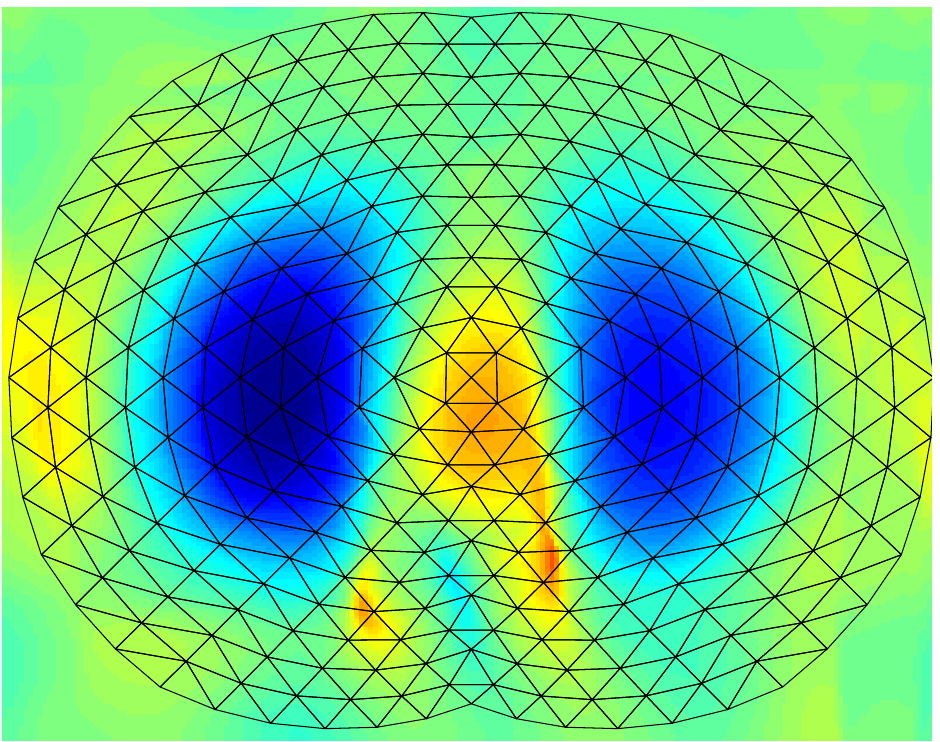} \\
  LR   &   SRR 
\end{tabular}
\end{center}
\caption{Example 3-a: Real lung image for adult shallow breathing  (data available through EIDORS), for the 19-th frame (end of an inspiratory cycle). First, second and third rows: LR EIT image and super resolved results (side by side) for the NOSER, TV and TS algorithms, respectively.}
\label{ex3a_results}
\end{figure}

\renewcommand{\sizeA}{3.5}
\begin{figure*}[!htb] 
\begin{center}
\begin{tabular}{cc}
  t=1s \hspace{\sizeA cm} & \hspace{\sizeA cm}  t=5s
\end{tabular}
\begin{tabular}{cc||cc}
  \includegraphics[width=\sizeA cm]{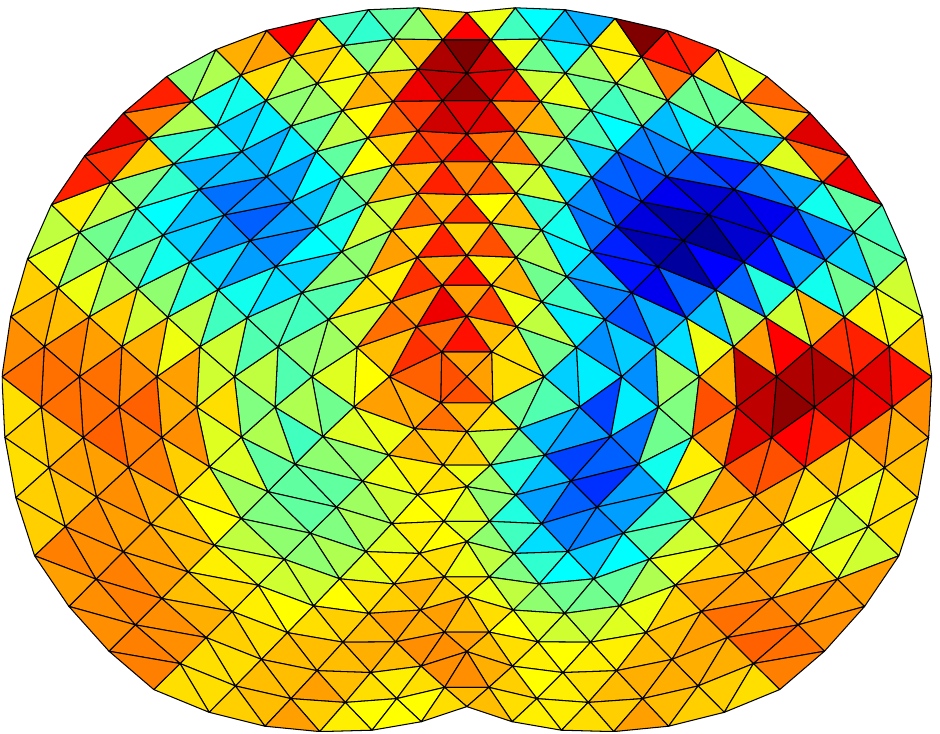}  &
  \includegraphics[width=\sizeA cm]{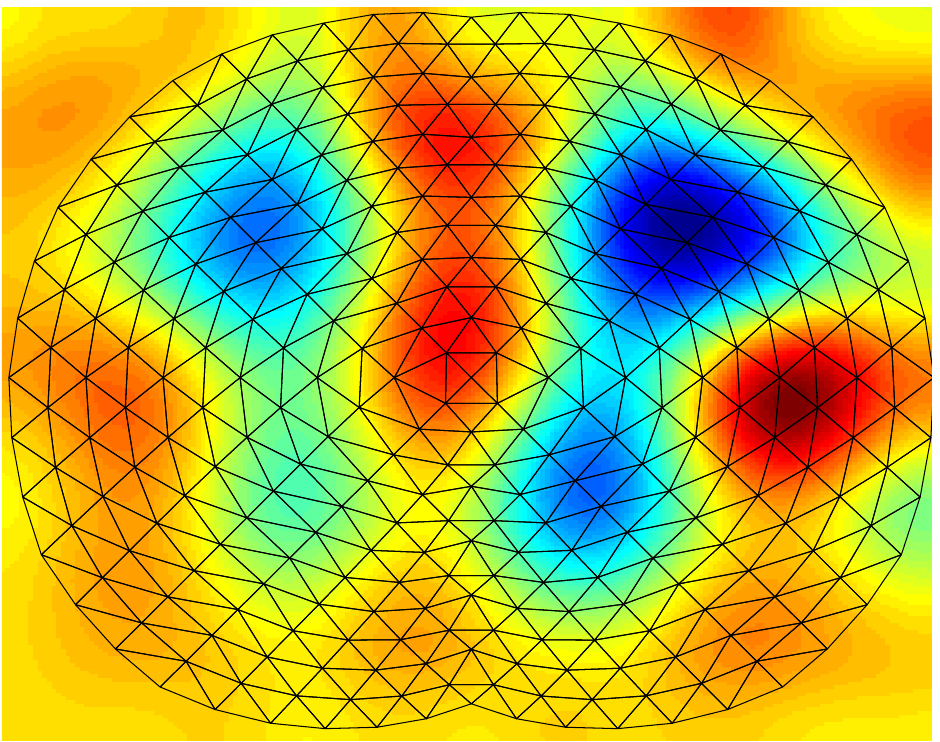} &
  \includegraphics[width=\sizeA cm]{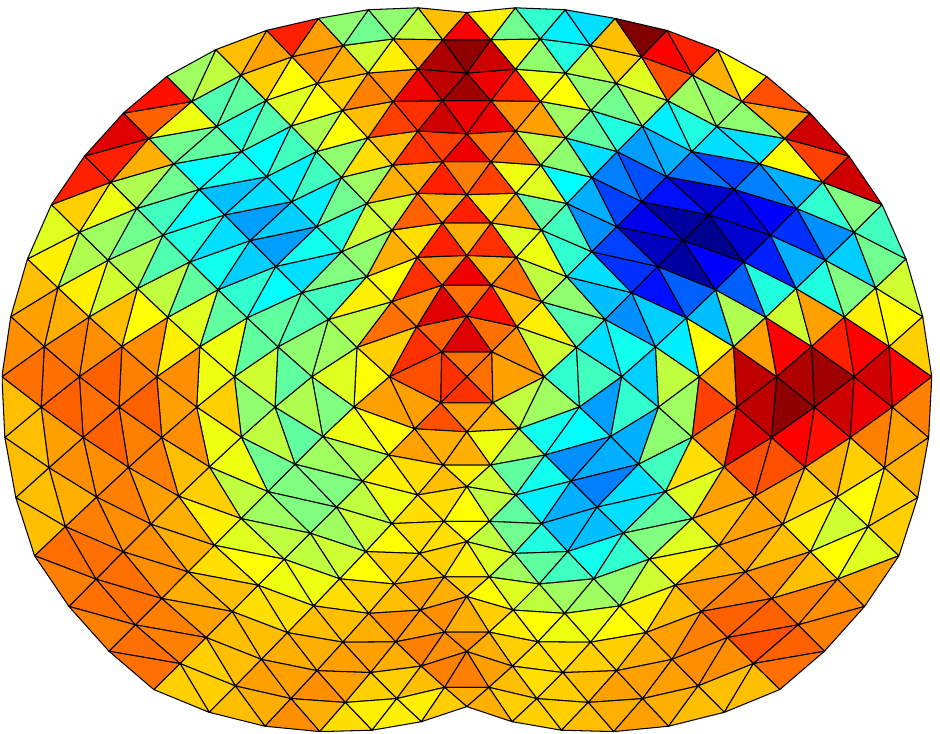}  &
  \includegraphics[width=\sizeA cm]{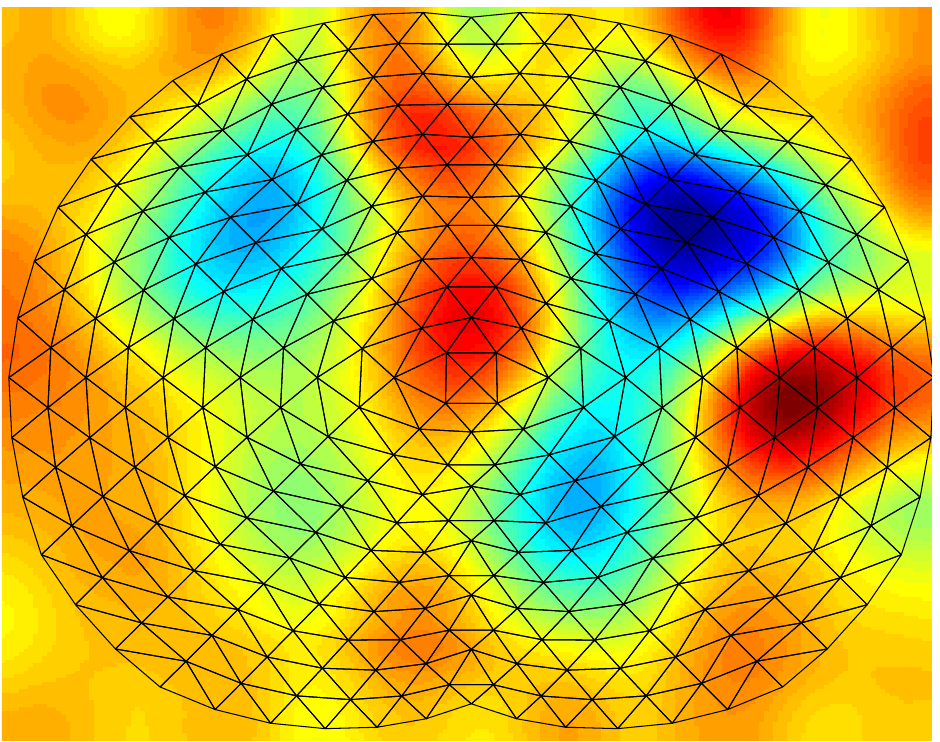}\\
  \includegraphics[width=\sizeA cm]{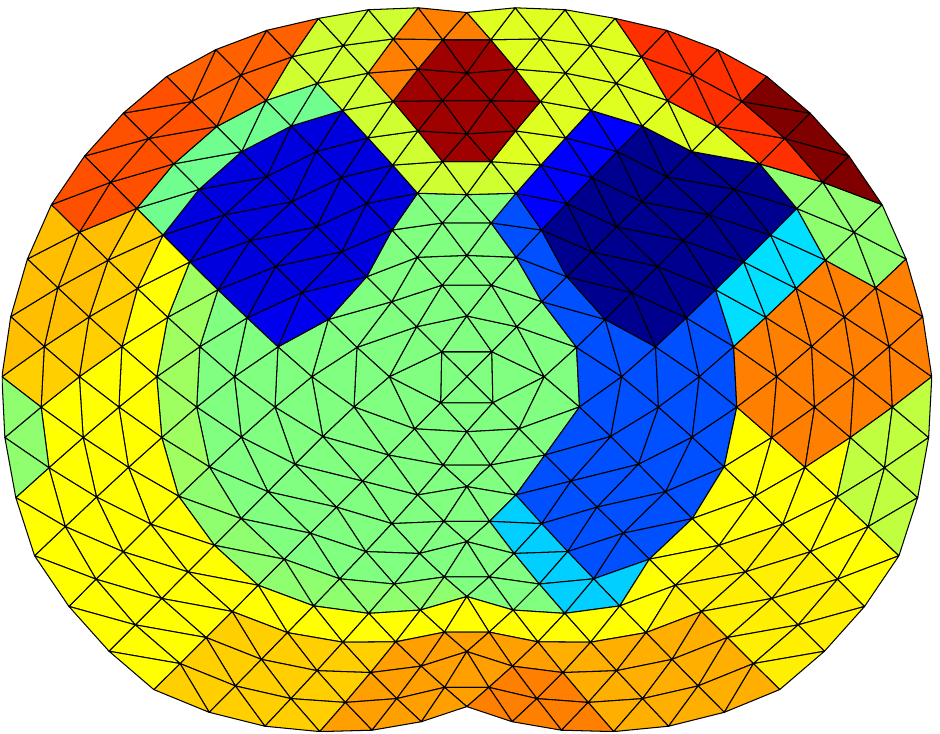}  &
  \includegraphics[width=\sizeA cm]{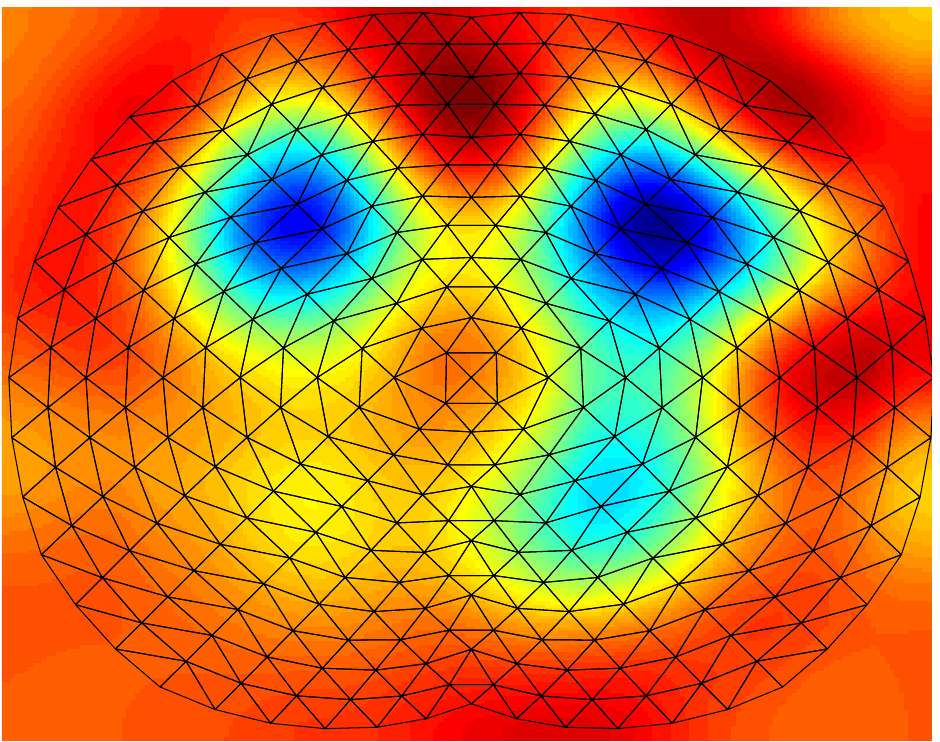} &
  \includegraphics[width=\sizeA cm]{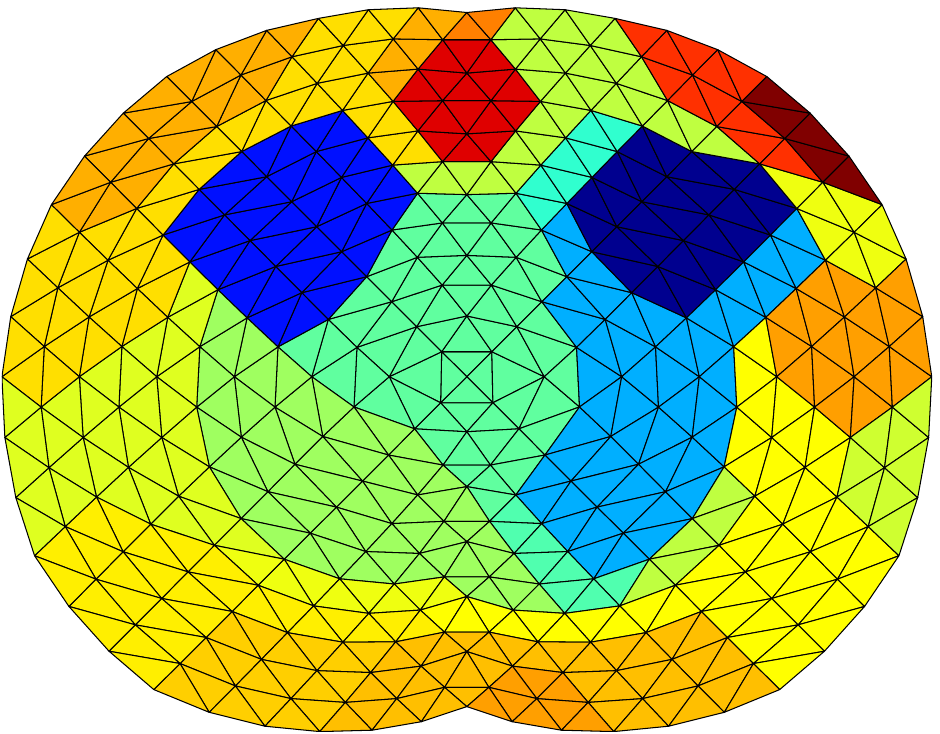}  &
  \includegraphics[width=\sizeA cm]{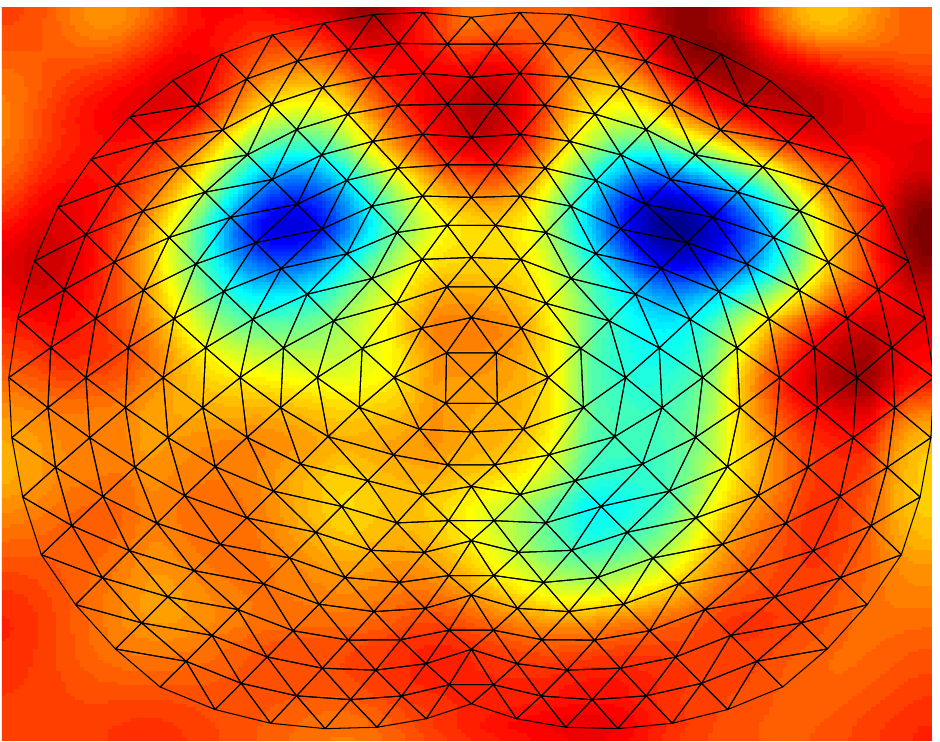}\\
  \includegraphics[width=\sizeA cm]{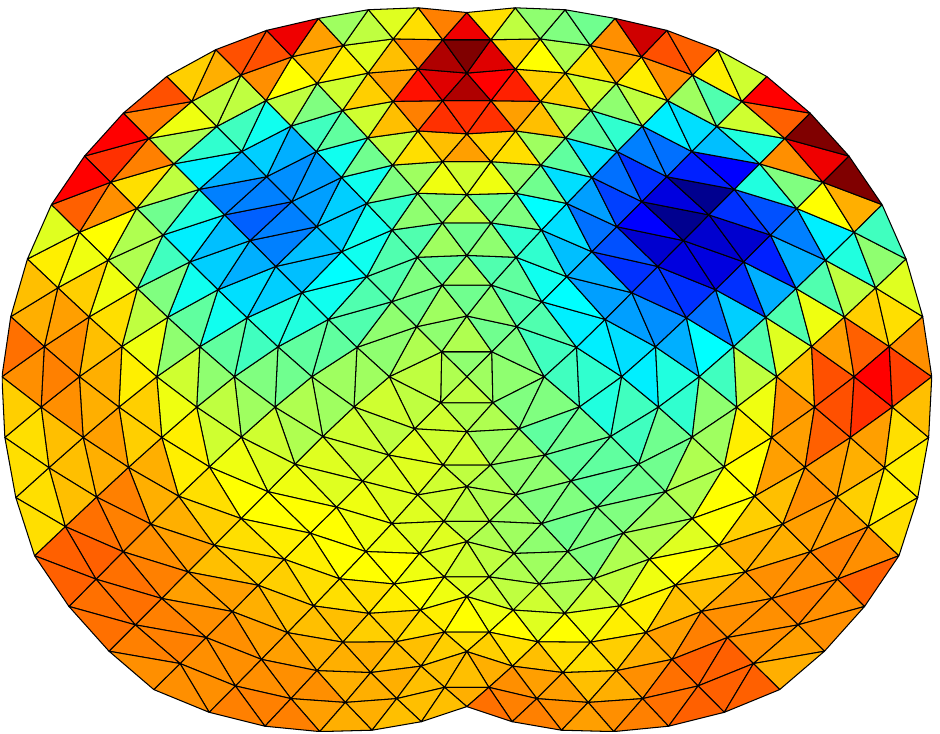}  &
  \includegraphics[width=\sizeA cm]{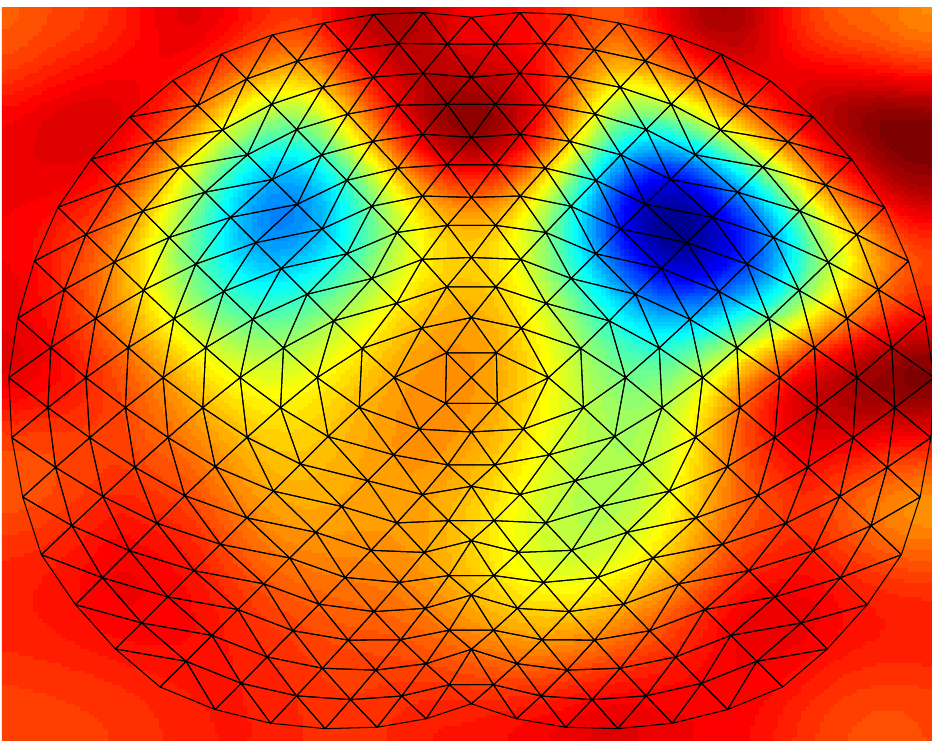} &
  \includegraphics[width=\sizeA cm]{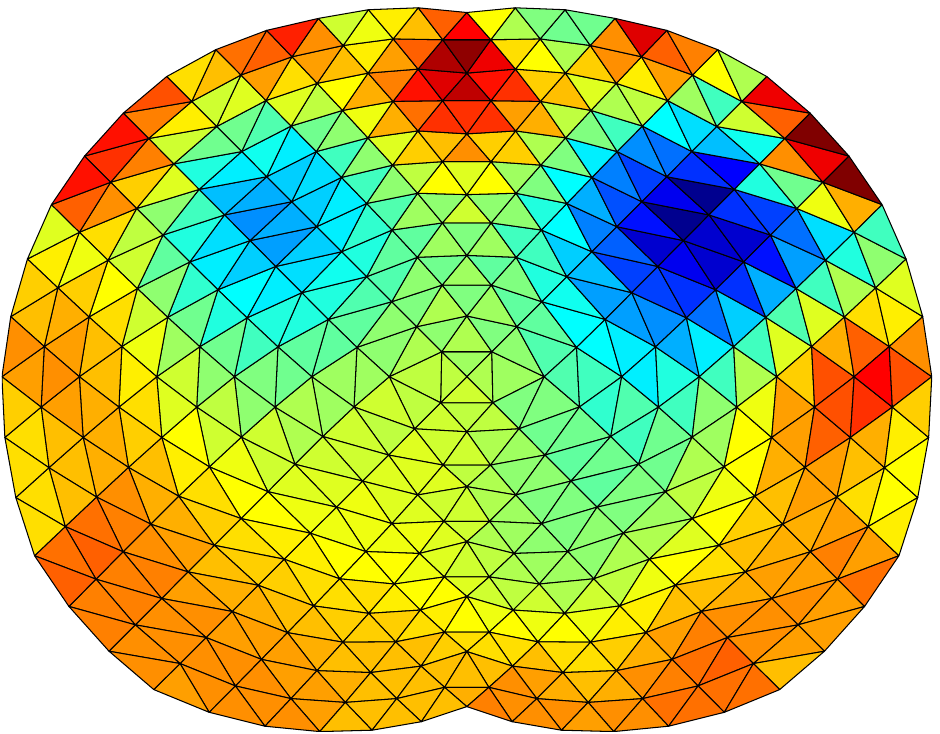}  &
  \includegraphics[width=\sizeA cm]{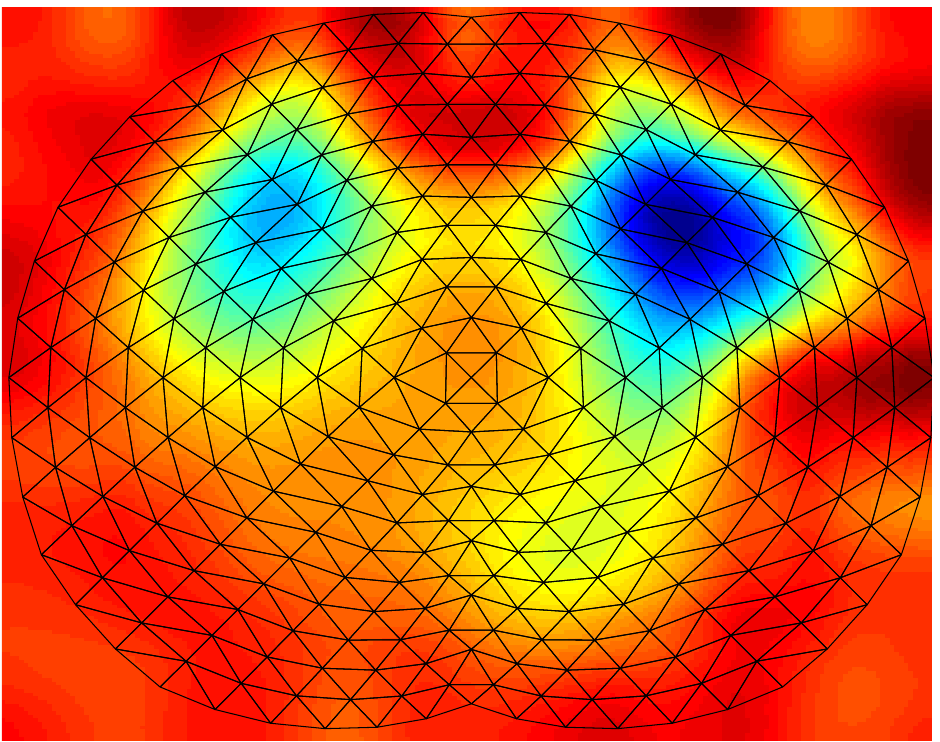} \\
  LR   &   SRR  &  LR  &  SRR
\end{tabular}
\end{center}
\caption{Example 3-b: Real lung images for tidal breathing (data available through EIDORS), for the end of two inspiratory cycles (t=1\,s and t=5\,s). First, second and third rows: EIT LR image and super-resolved results (side by side) for the NOSER, TV and TS algorithms, respectively.}
\label{ex3b_results}
\end{figure*}

\renewcommand{\arraystretch}{1.3}
\begin{table}
    \footnotesize
    \centering
    \begin{tabular}{|c|c|c|c|c|c|c|}
    \hline
    \multicolumn{7}{|c|}{Results for high SNR and finer mesh} \\ \hline
    &\multicolumn{2}{|c|}{Volume Ratio}& \multicolumn{2}{|c|}{Hausdorff Distance} & \multicolumn{2}{|c|}{MASD $(\times10^{-2})$} \\ \hline
       & LR & SRR & LR & SRR & LR & SRR \\ \hline
    NOSER & 0.46 & 0.52/0.48 & 0.12 & 0.09/0.11 & 0.99 & 0.73/0.96 \\ \hline
    TV    & 0.41 & 0.41/0.39 & 0.13 & 0.12/0.14 & 1.24 & 1.21/1.42 \\ \hline
    TS    & 0.43 & 0.48/0.46 & 0.14 & 0.11/0.12 & 1.20 & 0.90/1.10 \\ \hline
    \multicolumn{7}{|c|}{Results for low SNR and coarser mesh} \\ \hline
    &\multicolumn{2}{|c|}{Volume Ratio}& \multicolumn{2}{|c|}{Hausdorff Distance} & \multicolumn{2}{|c|}{MASD $(\times10^{-2})$} \\ \hline
       & LR & SRR & LR & SRR & LR & SRR \\ \hline
    NOSER & 0.38 & 0.50/0.46 & 0.21 & 0.10/0.12 & 1.70 & 0.83/1.10 \\ \hline
    TV    & 0.38 & 0.42/0.40 & 0.24 & 0.13/0.15 & 1.70 & 1.29/1.47 \\ \hline
    TS    & 0.39 & 0.46/0.44 & 0.16 & 0.12/0.13 & 1.52 & 1.03/1.22 \\ \hline
    \end{tabular}
    \vspace{0.25cm}
    \caption{Example 1: Average error metrics for translating T-Shaped object. SRR results are presented for the case considering known motion and using a registration algorithm (side by side).}
    \label{ref:tab_ex1_quant}
\end{table}

\renewcommand{\arraystretch}{1.3}
\begin{table}
    \footnotesize
    \centering
    \begin{tabular}{|c|c|c|c|c|c|c|}
    \hline
    \multicolumn{7}{|c|}{Results for high SNR and finer mesh} \\ \hline
    &\multicolumn{2}{|c|}{Volume Ratio}& \multicolumn{2}{|c|}{Hausdorff Distance} & \multicolumn{2}{|c|}{MASD $(\times10^{-2})$} \\ \hline
       & LR & SRR & LR & SRR & LR & SRR \\ \hline
    NOSER & 0.78 & 0.91/0.83 & 0.083 & 0.033/0.036 & 0.23 & 0.05/0.11 \\ \hline
    TV    & 0.70 & 0.87/0.80 & 0.174 & 0.086/0.081 & 0.81 & 0.30/0.31 \\ \hline
    TS    & 0.83 & 0.92/0.84 & 0.064 & 0.030/0.035 & 0.13 & 0.04/0.10 \\ \hline
    \multicolumn{7}{|c|}{Results for low SNR and coarser mesh} \\ \hline
    &\multicolumn{2}{|c|}{Volume Ratio}& \multicolumn{2}{|c|}{Hausdorff Distance} & \multicolumn{2}{|c|}{MASD $(\times10^{-2})$} \\ \hline
       & LR & SRR & LR & SRR & LR & SRR \\ \hline
    NOSER & 0.51 & 0.81/0.75 & 0.154 & 0.076/0.081 & 1.14 & 0.25/0.33 \\ \hline
    TV    & 0.46 & 0.79/0.73 & 0.216 & 0.085/0.087 & 1.85 & 0.31/0.39 \\ \hline
    TS    & 0.69 & 0.84/0.78 & 0.100 & 0.064/0.067 & 0.46 & 0.18/0.24 \\ \hline
    \end{tabular}
    \vspace{0.25cm}
    \caption{Example 2: Average error metrics for synthetic lung object. SRR results are presented for the motion estimated from the known HR images and from the LR observations (side by side).}
    \label{ref:tab_ex2_quant}
\end{table}




\section{Experimental Setup and Results}
\label{sec:results}

In this section we present experimental results illustrating the super resolution of EIT images using the imaging model proposed in Section~\ref{sec:model} for both synthetic and real data. Three examples were devised to illustrate the application of the proposed method under different circumstances. Afterwards, we present a discussion about the robustness and computational complexity of the SRR process.
The first example, depicted in Figures~\ref{ex1a_results} and~\ref{ex1b_results}, presents the reconstruction of images of a T-shaped object subject to translational displacements. The algorithm was tested both with motion known \textit{a priori} and with motion estimated by a registration algorithm.
The second example, depicted in Figures~\ref{ex2a_results} and~\ref{ex2b_results}, extends the first one to EIT images formed by ellipsoids of varying volume, which emulates the imaging of a breathing thorax.
The motion estimation was performed twice, using both the synthetic HR images and the observed IHR images as input, resulting in different levels of motion estimation errors.
%
In both Examples~1 and~2 the LR EIT images were generated considering two different cases: one with small and another with large measurement noise, using a finer and a coarser FE mesh, respectively.
This allows for a better assessment of the performance of the SRR methodology.
Finally, a third example, depicted in Figures~\ref{ex3a_results} and~\ref{ex3b_results}, present the super resolution of EIT images of real \textit{in vivo} data, considering two different experiments with voltage measurements acquired from healthy adult subjects breathing.

The EIT inverse problem was solved in EIDORS~\cite{Adler06eidors}, using three different algorithms, namely, the Gauss-Newton solver with the NOSER image prior~\cite{Cheney1990noser}, the Total Variation (TV) regularization~\cite{borsic2010invivoTV} and the temporal regularization (TS)~\cite{adler2007temporalEIT}, all considering difference imaging with the reference voltage set as the average of all measurements in the time window.
These algorithms include both a simple solution (NOSER) and sophisticated methodologies (TV and TS), which help to illustrate the performance of the SRR methodology for input images originating from a variety of algorithms, emphasizing its independence to the EIT inverse problem.
The FEMs referred to in all examples are standard meshes available in EIDORS~\cite{Adler06eidors}. The mesh was also displayed over the reconstructed results in the figures for ease of comparison.
The hyperparameters of the algorithms were manually selected in order to achieve a good performance.
For the examples with synthetic images, these were generated with a background conductivity of $1\,$S/m and inclusions of $2\,$S/m.

We have applied the Least Mean Squares SRR~\cite{Elad99} algorithm to super resolve the images in all the examples, with step size $\mu=0.01$ and $K=100$ iterations for each input image.
The motion estimation was performed using the \textit{Horn \& Schunk} Optical Flow algorithm of \cite{Sun10}\footnote{With parameters { \texttt{lambda=10$^6$}, \texttt{pyramid\_levels=4}, \texttt{pyramid\_spacing=2}}.} using IHR image pairs as input.
In all examples, matrix $\mH_{\tiny{\text{D}}}(t)$ corresponded to the space-variant kernel associated with the nonuniform mesh of each LR EIT image, which was assumed to be time-invariant. The distortion matrix $\mH_{\tiny{\text{B}}}(t)$ was modeled as a space- and time-invariant $60\times60$ Gaussian convolution mask with standard deviation $\sigma=20$ and Neumann (symmetric) boundary conditions. The nonuniform LR EIT images were upsampled to a $200\times200$ IHR uniform grid before processing, where the region outside the nonuniform mesh domain was padded with zeros.


Since the image sequences of Examples 1 and 2 were synthetically generated, the desired HR images were available, allowing a quantitative assessment of the SRR performance.
To evaluate the resolution of the LR and reconstructed images with respect to the HR images, we resort to quantitative metrics that evaluate how close the shapes of the objects in these images are to one another, using metrics employed in the evaluation of medical image segmentation algorithms \cite{sluimer2005lungCTsegmentation,gerig2001validationSegmentation}, described in detail below.

Denote the HR and the upsampled or super-resolved images as $\vx$ and $\vxh$, respectively.
We create a binary map for the shapes by thresholding the upsampled and super-resolved EIT images at 25\% of their maximum image value, i.e.
\[[\vxh_{\text{bin}}]_{ij} = 
\left\{\begin{array}{cc}
    1\,, & [\vxh]_{ij} \geq 0.25 \max(\vxh) \\
    0 & \text{otherwise.}
\end{array}\right.\]

This binary image representing the shape of the reconstructed object is then compared with the binary image $\vx_{\text{bin}}$ representing the shape of the true object in the HR image, using the following metrics~\cite{sluimer2005lungCTsegmentation,gerig2001validationSegmentation}:

\textit{1) Volume overlap fraction}: Measures the relative overlap of the true and estimated volumes, scoring 1 for perfect agreement and 0 for complete disagreement
\begin{align}
    O(\vxh_{\text{bin}},\vx_{\text{bin}}) = \frac{\|\vxh_{\text{bin}}\cap\vx_{\text{bin}}\|}{\|\vxh_{\text{bin}}\cup\vx_{\text{bin}}\|} \,.
    \nonumber
\end{align}
Note that the larger $O(\vxh_{\text{bin}},\vx_{\text{bin}})$ is, the closer the estimated binary image $\vxh_{\text{bin}}$ is to the desired (true) binary image $\vx_{\text{bin}}$.

\textit{2) Hausdorff distance}: Measures the largest distance that occurs between two surfaces. Denote by $d_{\min}(x,B)=\inf\{d(x,b), \,b\in B\}$ the minimal distance between a point $x$ and a surface $B$. The Hausdorff distance is the symmetric extension of the largest distance from all points $x\in\vx_{\text{bin}}$ and the surface $\vxh_{\text{bin}}$, given by
\begin{align}
    H(\vxh_{\text{bin}},\vx_{\text{bin}}) = \max\{
    & \sup\{d_{\min}(x,\vxh_{\text{bin}}),x\in\vx_{\text{bin}}\},
    \nonumber\\ & \nonumber
    \sup\{d_{\min}(x,\vx_{\text{bin}}),x\in\vxh_{\text{bin}}\}
    \} \,.
\end{align}

\textit{3) Mean absolute surface distance (MASD)}: Measures how much the two surfaces differ on average, consisting of a border positioning measure integrated along the entire surface. Denoting the average of the minimal distance between two surfaces $A$ and $B$ by $\bar{d}_{\min}(A,B)=\frac{1}{|A|}\sum_{a\in A} d_{\min}(a,B)$, the MASD is defined as 
\[MASD(\vx_{\text{bin}},\vxh_{\text{bin}}) = \frac{1}{2}[\,\bar{d}_{\min}(\vx_{\text{bin}},\vxh_{\text{bin}})+\bar{d}_{\min}(\vxh_{\text{bin}},\vx_{\text{bin}})] \,.\]

\subsection{Example 1}

In this example, a sequence of 20 images of a T-shaped object subject to random translational motion inside a circular body was generated.
The amplitude of the motion was determined by samples of a zero-mean white Gaussian variate with variance $0.3$, truncated so that  the resulting position was limited to the interval $[-0.15,0.15]$ to assure that the object lied entirely inside the body.

The synthetic (desired) EIT images were generated using the mesh ``d2d1c" with 32 electrodes for the direct problem, and random noise was added to the voltage measurements.
The inverse problem was then solved for two situations, one considering a measurement signal to noise ratio (SNR) of $10$dB and using finer mesh (``c2c"), displayed in Figure~\ref{ex1a_results}, and another considering a SNR of $-5$dB and using a coarser FEM (``a2d3c"),  displayed in Figure~\ref{ex1b_results}.

The super resolution reconstruction was performed twice, first considering the motion to be known \textit{a priori} and afterwards estimating it using a registration algorithm~\cite{Sun10}.
The synthetic desired HR images, their corresponding EIT observations (LR images $\vy_{\triangle}(t)$), and the super resolution results for the case of estimated motion are depicted in Figures~\ref{ex1a_results} and~\ref{ex1b_results}.
The corresponding quantitative results are shown in Table~\ref{ref:tab_ex1_quant}.

Although the object is already well identifiable in the LR images in the case of a high measurement SNR, a slight improvement in the representation of the object's shape can be  generally noticed in the reconstructed images, specially for the NOSER and TS algorithms.
This is also reflected in the quantitative results with a consistent improvements across all evaluated metrics, except for the case of the TV algorithm, where the results considering estimated motion did not provide a good quantitative performance.

The improvements in the reconstructed results are much more noticeable when a lower measurement SNR is employed, in which case the shape of the object in the reconstructed images is much closer to that of the desired one (in the original HR EIT image) when compared to the LR counterparts.
This observation is consistently reflected in the quantitative performance metrics.

Note that, in both cases a slight degradation in the performance of the quantitative metrics is observed when a registration algorithm is used, which is a well known fact in SRR literature. Nevertheless, a significant improvement in resolution can still be generally verified in the super resolved images when compared with their LR counterparts, which indicates that the proposed approach is effective in practical scenarios.

\subsection{Example 2}

In this example, we performed the reconstruction of a sequence of 20 images of a synthetic lung phantom. A thorax was emulated by a circular body, with ellipses of varying area acting as the lungs, and an additional static circle denoting the spine.
Like in Example 1, synthetic (desired) EIT images were generated using the mesh ``d2d1c" with 32 electrodes for the direct problem, and random noise was added to the voltage measurements.
The inverse problem was again solved for two situations, one considering a measurements SNR of $10$dB and using finer mesh (``c2c"), displayed in Figure~\ref{ex2a_results}, and another considering a SNR of $-5$dB and using a coarser FEM (``a2d3c"),  displayed in Figure~\ref{ex2b_results}.

In order to evaluate the effects of the registration accuracy in the SRR algorithm, we preformed the super resolution reconstruction twice, once estimating the motion from the HR images, and another estimating the motion from the LR observations~\cite{Sun10}.
The synthetic desired HR images, their corresponding EIT observations (LR images $\vy_{\triangle}(t)$), and the super resolution results for the case of motion estimated from the LR observations are depicted in Figures~\ref{ex2a_results} and~\ref{ex2b_results}.
The corresponding quantitative results are shown in Table~\ref{ref:tab_ex2_quant}.

The performance improvement in the reconstruction results for this example is more significant than in the previous one. Both in the case with higher and lower SNR, the shape of the objects in the reconstructed images is generally much closer to the desired one in the reconstructed images, especially when the SNR is lower.
The quantitative results are consistent and also corroborate with the perceptual results, and a degradation in performance is again noticed when a less accurate registration information is used.

This example also points to the better ability of the temporal solver~\cite{adler2007temporalEIT} to operate in low SNR conditions, since it exploits the correlation between the images in adjacent time instants. 
Nevertheless, a clear improvement is noted for the SRR results in the case of a lower measurement SNR.

\subsection{Example 3}

In this example the proposed technique is employed in super-resolving sequences of lung EIT images of adult subjects from two different data sets, both available through EIDORS~\cite{Adler06eidors}.
In both cases, we use the ``c2t2" mesh with 16 electrodes in the inverse problem, which represents the shape of the adult human thorax more accurately.
The motion between the frames was estimated using a registration algorithm~\cite{Sun10}.
Since the HR (desired) images were not available, the quantitative performance metrics could not be evaluated for this example.
Figures~\ref{ex3a_results} and~\ref{ex3b_results} show the LR observations (EIT image $\vy_{\triangle}(t)$) and the corresponding reconstruction results for frames selected at the end of expiratory cycles.

For the results of the first data set depicted in Figure~\ref{ex3a_results}, it can be noted that the lungs are already well represented in the LR EIT images.
Nevertheless, the boundary of the lungs appears to be slightly clearer in the reconstructed images. Besides, the size of the lungs seems more consistent across the super resolved images generated using the different EIT algorithms, as opposed to their LR counterparts.

For the results of the second data set depicted in Figure~\ref{ex3b_results}, more pronounced improvements are obtained in the super resolved images.
The lungs (objects of interest) are more well defined and more easily identifiable from the image background, with a reduced amount of artifacts when compared to the observed LR EIT images.

\begin{figure*}[thb] 
\begin{center}
\begin{tabular}{cccc}
  \scalebox{.7}{\includegraphics{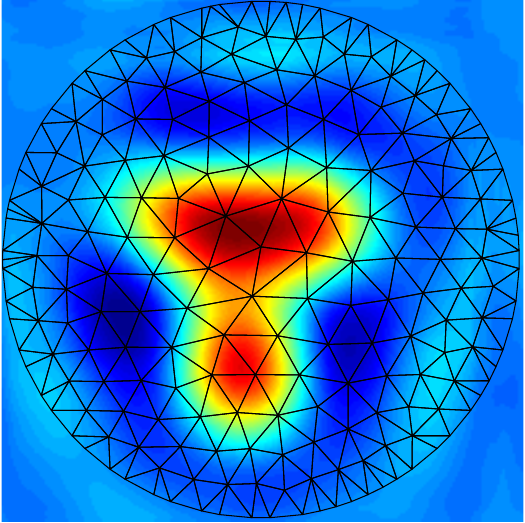}} 
  \hspace{0.5cm} & \hspace{0.75cm} 
  \scalebox{.7}{\includegraphics{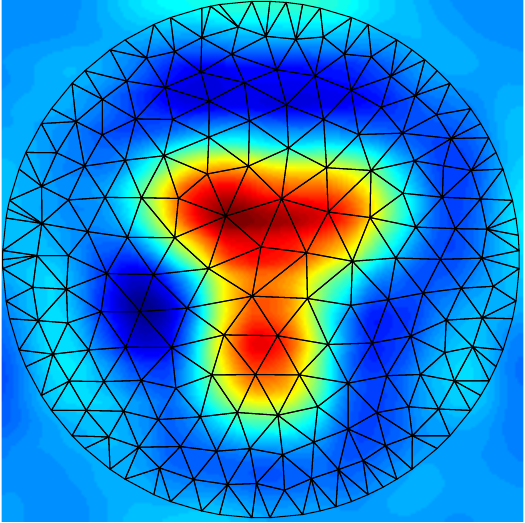}}
  \\
  (a) 
  \hspace{0.5cm} & \hspace{0.75cm} 
  (b)  \\
  %
\end{tabular}
\end{center}
\caption{\cred{Robustness of the reconstruction results. (a) Super-resolved EIT image considering known motion. (b) Super-resolved EIT image considering estimated motion.}}
\label{added_robustness_misreg}
\end{figure*}

\bigskip
\subsection{Robustness and Computational Complexity}

A critical step of the super-resolution process consists of estimating the relative motion between the different LR images. This estimate is often imprecise, which makes robustness to registration errors an important property of super-resolution algorithms~\cite{Nasrollahi14}.
To assess the robustness of the proposed method to misregistration, we first repeat the simulation of Example 1-b, consisting of a translating T-shaped object with low measurement SNR and a coarser FEM, using the NOSER algorithm~\cite{Cheney1990noser} and perform a visual comparison of the SRR results with both known and estimated motion.
The results, depicted in Figure~\ref{added_robustness_misreg}, indicate that the use of a registration algorithm does not introduce significant distortions in the reconstructed image, with only a small distortion in the upper part of the "T" shape and a slightly more asymmetric amplitude distribution inside the object being noticeable.
The quantitative results for both Example~1 and Example~2 also show that, although there is a slight reduction in performance when the motion is estimated from the LR images, the overall image quality is still considerably better than that of the LR observations, which illustrates the feasibility of the super-resolution of EIT images.

%
%
Another important issue regarding the application of SRR techniques to EIT images is the computational complexity of the solution. 
Since one of the main advantages of electrical impedance tomography lies on its suitability for real-time operation, as evidenced by the great interest in its use in bedside and real-time monitoring \cite{adler2012whitherLungEIT}, EIT systems generate a large amount of data which must be processed in a timely fashion.
This places the SRR of EIT images on the same category of a multitude of emerging applications for which efficient solutions are required to process vast volumes of data, including data mining~\cite{wu2014dataMining}, hyperspectral image processing~\cite{borsoi2017tech,imbiriba2018low}, and health care applications~\cite{belle2015big}.

Fortunately, since the proposed EIT imaging model has the same form of the acquisition models for traditional SRR applications, there is no additinal complexity introduced due to the different underlying imaging modality.
For instance, the solution of the problem in~\eqref{eq:cost_function_illustrative_vid} can be performed through a matrix inversion approach with complexity of the order of $\mathcal{O}(M^3)$ (i.e. cubic in the number of HR pixels).
Nevertheless, more efficient SRR algorithms exist, exploiting the characteristics of the blurring and decimation matrices~\cite{elad1997srr_simple} or using iterative methods~\cite{Elad99}, which can also benefit from preconditioning techniques~\cite{Nasrollahi14} and can be made to operate in real time~\cite{borsoi2017srr_jrnl}.

\section{Conclusion}
\label{sec:conclusions}

In this paper we have proposed a new method for super-resolution reconstruction of EIT images. The proposed method uses a new model for the EIT image formation system, which is based on the Penrose Pixels approach developed for optical image acquisition systems. Although the Penrose Pixels Super-Resolution methodology still finds no practical application due to the unavailability of appropriate sensors, we have shown that its essence can be adapted to handle the non-conventional imaging sensors typical of EIT images.
Simulation results employing the LMS-SRR algorithm show that a significant increase in the visual quality and resolution of EIT images obtained using different finite element meshes and EIT algorithms can be achieved with the proposed method. This indicates the possibility of mitigating the severe deficiency in image quality that currently prevents EIT imaging systems to be more widely used.
The visual quality of the resulting image was significantly improved, and quantitative performance metrics show that the shapes of the objects in the reconstructed images are substantially closer to those in the HR images when compared to the LR observations, indicating that super-resolution effectively has been obtained. Significant improvements were verified for both idealized and practical scenarios.

\section*{Acknowledgments}
This work has been supported by the the National Council for Scientific and Technological Development (CNPq).

\bibliography{references}

\end{document}